\documentclass[sigconf,authorversion,nonacm]{acmart}

\usepackage{hyperref}
\usepackage{url}
\usepackage{graphicx}
\usepackage{amsmath}
\usepackage{bbm}
\usepackage{multirow}
\usepackage{xcolor, soul}
\usepackage{makecell}
\usepackage{enumitem}
\usepackage{pifont}
\usepackage{listings}

\definecolor{codegreen}{rgb}{0,0.6,0}
\definecolor{codegray}{rgb}{0.5,0.5,0.5}
\definecolor{codepurple}{rgb}{0.58,0,0.82}
\definecolor{backcolour}{rgb}{0.95,0.95,0.92}

\lstdefinestyle{mystyle}{
    backgroundcolor=\color{backcolour},   
    commentstyle=\color{codegreen},
    keywordstyle=\color{magenta},
    numberstyle=\tiny\color{codegray},
    stringstyle=\color{codepurple},
    basicstyle=\ttfamily\footnotesize,
    breakatwhitespace=false,         
    breaklines=true,                 
    captionpos=b,                    
    keepspaces=true,                 
    numbers=left,                    
    numbersep=5pt,                  
    showspaces=false,                
    showstringspaces=false,
    showtabs=false,                  
    tabsize=2
}

\usepackage{tikz}
\newcommand{\tikzxmark}{%
\tikz[scale=0.23] {
    \draw[line width=0.7,line cap=round] (0,0) to [bend left=6] (1,1);
    \draw[line width=0.7,line cap=round] (0.2,0.95) to [bend right=3] (0.8,0.05);
}}
\newcommand{\tikzcmark}{%
\tikz[scale=0.23] {
    \draw[line width=0.7,line cap=round] (0.25,0) to [bend left=10] (1,1);
    \draw[line width=0.8,line cap=round] (0,0.35) to [bend right=1] (0.23,0);
}}

\newcommand{\mb}{\mathbf}

\newcommand{\mc}{\mathcal}

\newcommand{\ie}{\textit{i.e.,}}
\newcommand{\eg}{\textit{e.g.,}}

\newtheorem{definition}{\textsc{Definition}}

\newcommand{\our}{\textsc{Graph-ToolFormer}}
\newcommand{\toolformer}{\textsc{Toolformer}}

\newcommand{\bapi}{\textsc{<API>}}
\newcommand{\eapi}{\textsc{</API>}}

\begin{document}
\title{Graph-ToolFormer: To Empower LLMs with Graph Reasoning Ability via Prompt Augmented by ChatGPT}

\author{Jiawei Zhang}
\email{jiawei@ifmlab.org}
\affiliation{%
  \institution{IFM Lab \\Department of Computer Science, \\University of California, Davis}
  \city{Davis}
  \state{California}
  \country{USA}
  \postcode{95616}\\
  {\textcolor{blue}{https://github.com/jwzhanggy/Graph\_Toolformer}}
}

\begin{abstract}

In this paper, we aim to develop a large language model (LLM) with the reasoning ability on complex graph data. Currently, LLMs have achieved very impressive performance on various natural language learning tasks, extensions of which have also been applied to study the vision tasks with data in multiple modalities. However, when it comes to the graph learning tasks, existing LLMs present very serious flaws due to their inherited weaknesses in performing \textit{precise mathematical calculation}, \textit{multi-step logic reasoning}, \textit{perception about the spatial and topological factors}, and \textit{handling the temporal progression}.

To address such challenges, in this paper, we will investigate the principles, methodologies and algorithms to empower existing LLMs with the graph reasoning ability, which will have tremendous impacts on the current research of both LLMs and graph learning. Inspired by the latest ChatGPT and Toolformer models, we propose the {\our} (Graph Reasoning oriented Toolformer) framework to teach LLMs themselves with prompts augmented by ChatGPT to use external graph reasoning API tools. Specifically, we will investigate to teach {\our} to handle various graph data reasoning tasks in this paper, including both (1) \textit{very basic graph data loading and graph property reasoning tasks}, ranging from simple graph order and size to the graph diameter and periphery, and (2) \textit{more advanced reasoning tasks on real-world graph data}, such as bibliographic paper citation networks, protein molecular graphs, sequential recommender systems, online social networks and knowledge graphs. 

Technically, to build {\our}, we propose to hand-craft both the instruction and a small amount of prompt templates for each of the graph reasoning tasks, respectively. Via in-context learning, based on such instructions and prompt template examples, we adopt ChatGPT to annotate and augment a larger graph reasoning statement dataset with the most appropriate calls of external API functions. Such augmented prompt datasets will be post-processed with selective filtering and used for fine-tuning existing pre-trained causal LLMs, such as the GPT-J and LLaMA, to teach them how to use graph reasoning tools in the output generation. To demonstrate the effectiveness of {\our}, we conduct extensive experimental studies on various graph reasoning datasets and tasks, and have also launched a LLM demo with various graph reasoning abilities. All the source code of {\our} framework, the demo for graph reasoning, and the graph and prompt datasets have been released online at the project github page.

\end{abstract}


\maketitle


%
%
%
%

\section{Introduction}\label{sec:introduction}

\begin{figure*}
    \centering
    \begin{minipage}{\textwidth}
    	\includegraphics[width=\linewidth]{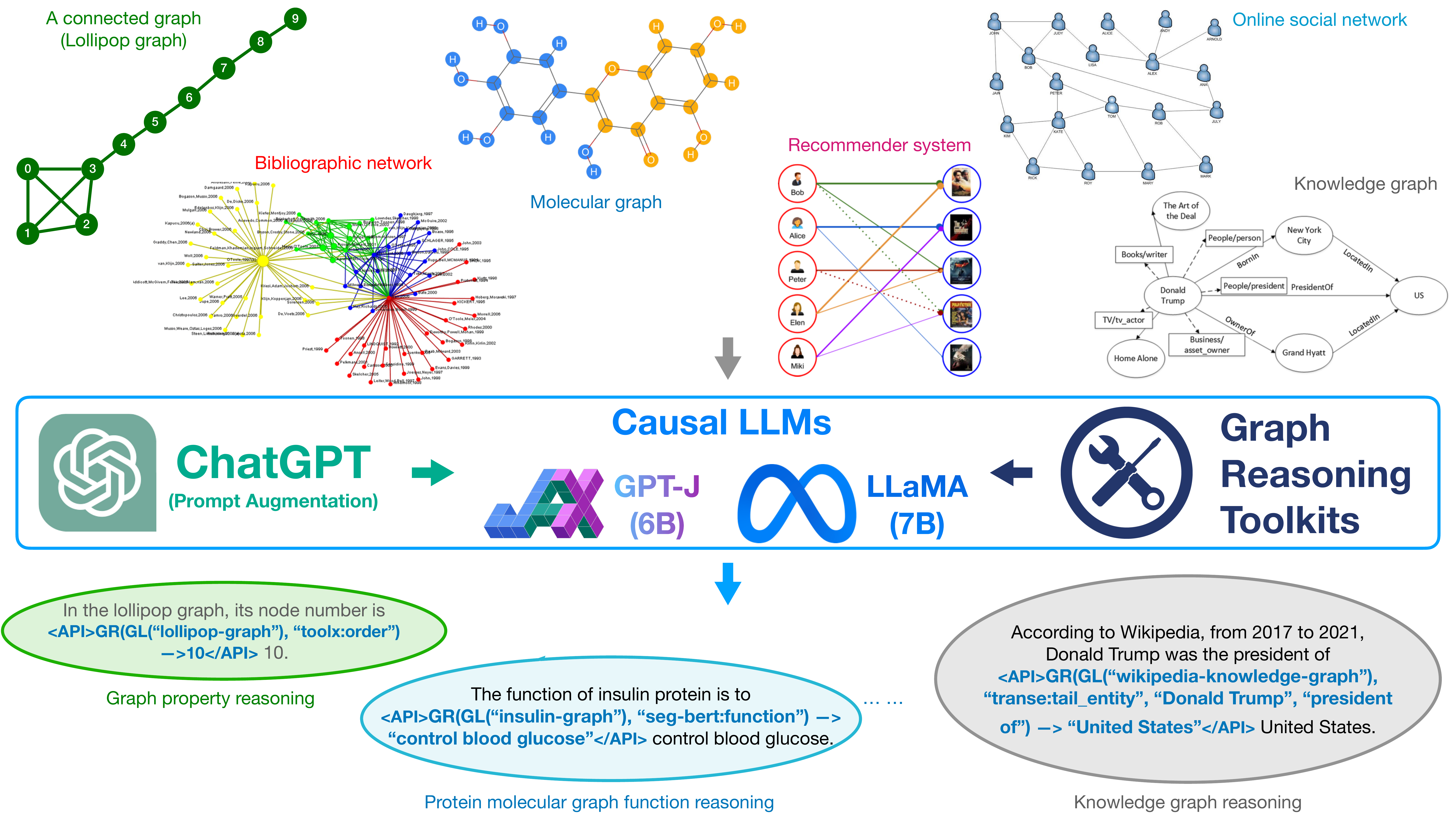}
        \caption{An Illustration of LLMs based Graph Reasoning Tasks. Based on the input graph data from various domains and a handful number of prompt examples with brief instructions, we propose to use ChatGPT to annotate and augment a large prompt dataset that contains graph reasoning API calls of external graph reasoning tools. The generated prompt dataset will be used to fine-tune the existing pre-trained LLMs, like GPT-J or LLaMA, to teach them to automatically use the most appropriate external API tools for accomplishing the input graph reasoning tasks.}
    	\label{fig:problem_illustration}
	\end{minipage}
\end{figure*}

In recent years, large language models (LLMs) \cite{Vaswani2017AttentionIA, Devlin2019BERTPO, Ouyang2022TrainingLM} have achieved very impressive performance on a variety of natural language processing tasks \cite{OpenAI2023GPT4TR, Touvron2023LLaMAOA, Ouyang2022TrainingLM}, extensions of which have also been extensively applied to solve many other problems with data in different modalities as well \cite{OpenAI2023GPT4TR, Dosovitskiy2020AnII, Ramesh2022HierarchicalTI, Ramesh2021ZeroShotTG}. With the launch of ChatGPT and new Microsoft Bing Chat based on both GPT-3.5 and GPT-4, LLMs have also been widely used in people's daily production and life. At the same time, due to their inherent limitations, these LLMs have also received lots of criticisms in their usages due to their inherited weaknesses, like \textit{inability in performing precise calculations} \cite{patel-etal-2021-nlp}, \textit{difficulty in addressing multi-step logic reasoning problems} \cite{Creswell2022SelectionInferenceEL}, \textit{incapable to conduct spatial and topological reasoning} \cite{Bang2023AMM}, and \textit{unawareness of progression of temporal factors} \cite{10.1162/tacl_a_00459}.

With the parallel development of natural language processing and computer vision, transformer based deep learning models on graph structured data has also received lots of attention from the community in recent years \cite{Zhang2020GraphBertOA, Yun2019GraphTN, Hu2020GPTGNNGP}. Graph provides a unified representation for many inter-connected data in the real-world, which models both the diverse attributes of the nodes and the extensive links connecting the nodes with each other. Besides the classic graph structures we learn from the \textit{discrete math} and \textit{algorithm} courses, as shown in Figure~\ref{fig:problem_illustration}, lots of real-world data can also be modeled as graphs \cite{Shi2015ASO}, like \textit{bibliographic networks} \cite{10.14778/3402707.3402736}, \textit{protein molecular graphs} \cite{doi:10.1142/S0219633602000117}, \textit{recommender systems} \cite{PremRec}, \textit{online social networks} \cite{10.1145/1298306.1298311}, and \textit{knowledge graphs} \cite{9416312}. 

Meanwhile, compared with the prosperous research explorations on incorporating vision and language data into LLMs for designing the ambitious AGI development plan \cite{agi}, it seems researchers have either ``\textit{unintentionally}'' or ``\textit{intentionally}'' ignored the widely existed graph data and don't seem to have any plans to include them into the LLMs building for achieving the AGI. 

Here, we say researchers have ``\textit{unintentionally}'' ignored graphs, since compared with texts and images that we deal with everyday, graph has long-time been merely used as an intermediate modeling data structure for real-world data and we normally have no direct interactions with graph actually. It is natural that people may mistakenly think graph should not be the focus at the current stage for creating AIGC and building the AGI systems. At the same time, we say researchers may have ``\textit{intentionally}'' ignored graphs, since graph learning may involve (1) lots of precise mathematical calculations of graph properties, (2) multi-hop logical reasoning through the links, (3) capturing the extensively connected graph spatial and topological structures, and (4) sometimes we also need to handle the dynamics of graphs that are changing with time. Careful readers may have noticed that these requirements mentioned for graph learning actually hit the nail on the head, which exactly correspond to the weaknesses of the current LLMs we mentioned at the very beginning.

Regardless of the potential challenges ahead of us, ``\textit{an AGI without graph reasoning ability will never be the AGI we may desire}''. Based on such motivations, we write this paper trying to incorporate graph data into LLMs for various graph reasoning tasks. On the one hand, we really hope the currently AI-leading companies like OpenAI, Microsoft, Google and Meta can take graph structured data reasoning into consideration when they develop their missions and plans for achieving the AGI, so that the graph learning community will be able to contribute our efforts to building the AGI system together with the language and vision communities. On the other hand, we also hope to empower the existing LLMs with the ability to overcome the weaknesses in their performance when handling graph structured data for complex graph reasoning tasks. So, the latest developed LLMs can also benefit the graph learning community for solving various graph reasoning tasks as well.

Considering the current language models and their extremely high pre-training costs, we cannot fundamentally re-design a new LLM with pre-training to equip them with the graph reasoning capabilities. Pre-training such LLMs from scratch is an infeasible task for most research groups in academia and majority of companies in the industry as well. To adapt to the common practices of NLP approaches, we will introduce the Graph Reasoning oriented Toolformer framework ({\our}) by fine-tuning some existing pre-trained LLMs ({\eg} GPT-J or LLaMA) in this paper. Technically, as illustrated in Figure~\ref{fig:problem_illustration}, based on the latest ChatGPT from OpenAI and {\toolformer} model from Meta \cite{Schick2023ToolformerLM}, we propose to provide the existing pre-trained LLMs ({\eg} GPT-J or LLaMA) with the ability to perform various complex graph reasoning tasks by allowing them to use external graph learning tools, such as other pre-trained graph neural network models and existing graph reasoning toolkits. Instead of manually hard-coding the graph data loading and external graph learning tool usage function calls in the reasoning statements, to make {\our} as a general graph reasoning interface, we will fine-tune the LLMs to teach the models to decide not only \textit{where} to retrieve the graph data, but also \textit{what} tools to be used, as well as \textit{when} and \textit{how} to use these tools. More technical details about the {\our} model will be introduced in the following methodology section.

As the first exploration attempt to use LLMs for general graph reasoning tasks, we summarize the contributions of this paper as follows:
\begin{itemize}

\item \textbf{Graph Reasoning with LLMs}: This paper is the first paper that attempts to propose a general LLM, {\ie} {\our}, that can handle graph reasoning tasks. It effectively remedies the weaknesses of existing LLMs on graph reasoning. More importantly, it helps bridge the graph learning community with the latest development on LLMs and AIGC led by the language and vision learning communities. So people in the graph learning community will also have the stage to demonstrate our skills and expertises in the current era of AIGC and the future era AGI.

\item \textbf{Graph Reasoning Prompt Dataset}: In this paper, we create a handful number of human-written language instructions and prompt examples of how graph learning tools can be used. Based on the self-supervised in-context learning, we use ChatGPT to annotate and augment a large graph reasoning dataset with API calls of different external graph learning tools, which will also be post-processed with selective filtering. Via the github page\footnote{https://github.com/jwzhanggy/Graph\_Toolformer/tree/main/data}, we have released both the graph raw datasets and the generated graph reasoning prompt dataset used in this paper with the community for future explorations.

\item \textbf{Extensive Experimental Studies}: We have extensively tested the effectiveness of our proposed {\our} with various graph reasoning based application tasks studied in the real-world, which include the most basic graph data loading and general graph property computation tasks, as well as some more advanced ones. Specifically, we study several challenging advanced graph reasoning tasks in the experiments, which include paper topic inference in bibliographic networks, molecular graph function prediction, online social network community detection, personalized sequential recommendation in recommender systems and knowledge graph entity and relation reasoning.

\end{itemize}

The remaining sections of this paper are organized as follows. We will briefly introduce the related work in Section~\ref{sec:related_work}. The definitions of some terminologies and the formulation of the studied problem will be provided in Section~\ref{sec:formulate}. A detailed introduction about the {\our} framework will be provided in Section~\ref{sec:method}. The effectiveness of {\our} will be tested with extensive experiments on real-world benchmark graph datasets in Section~\ref{sec:experiment}. Finally,  we will conclude this paper in Section~\ref{sec:conclusion} and briefly discuss about some potential future exploration directions in Section~\ref{sec:future_work}.

\section{Related Work}\label{sec:related_work}

In this section, we will discuss about several research topics that are related to our {\our} framework proposed in this paper, which include \textit{graph neural networks}, \textit{language models}, \textit{language model based graph learning} and \textit{prompt tuning}.

\subsection{Graph Neural Networks}

Graph neural networks (GNNs) aim to learn the embedding representations of the graph structured data. Representative examples of GNNs proposed already include {GCN} \cite{Kipf_Semi_CORR_16} and Graph-Bert \cite{Zhang2020GraphBertOA}, based on which various extended variants \cite{Velickovic_Graph_ICLR_18,sun2019adagcn,DBLP:journals/corr/abs-1810-05997} have been introduced as well. As mentioned above, {GCN} and its variant models are all based on the approximated graph convolutional operator \cite{Hammond_2011}, which may lead to the suspended animation problem \cite{Zhang2019GResNetGR} and over-smoothing problem \cite{Li_Deeper_CORR_18} for deep model architectures. Theoretic analyses of the causes are provided in \cite{Li_Deeper_CORR_18,Zhang2019GResNetGR,Merve_An_19}. To handle such problems, \cite{Zhang2019GResNetGR} generalizes the graph raw residual terms and proposes a method based on graph residual learning; \cite{Li_Deeper_CORR_18} proposes to adopt residual/dense connections and dilated convolutions into the GCN architecture. Besides the GCN and Graph-Bert based models, several other work \cite{sun2019adagcn,Huang_Inductive_19} also seeks to involve the recurrent network for deep graph representation learning instead.

\subsection{Language Models}

Since the proposal of Transformer \cite{Vaswani2017AttentionIA}, large language models (LLMs) have become the dominant deep model for various NLP tasks. Assisted with pre-training, the giant tech-companies have also introduced their own versions of different LLMs, like BERT from Google \cite{Devlin2019BERTPO}, BART from Meta \cite{Lewis2019BARTDS}, GPT from OpenAI \cite{Radford2018ImprovingLU, Radford2019LanguageMA, Brown2020LanguageMA}, ELMo from AI2 \cite{Peters2018DeepCW} and MT-DNN from Microsoft \cite{Liu2019MultiTaskDN}. Many of these LLMs have also been open-sourced with both model algorithm and learned parameters released to the community for both research and application purposes. One research paper closely related to this work is Toolformer \cite{Schick2023ToolformerLM} from Meta, which proposes to incorporate external APIs into language models. Equipped with such external APIs, the models will be able to automatically decide how to use which tool. Meanwhile, even prior to the Toolformer model, several other previous papers \cite{Parisi2022TALMTA, Mialon2023AugmentedLM} have also explored to augment language models with external tools.

\subsection{Prompt Tuning}

Prompts have been shown to be effective in tuning the pre-trained language models with zero-shot or few-shot learning \cite{Brown2020LanguageMA}, which can help language models learn faster than traditional fine tuning tasks. By now, we have witnessed three categories of prompt tuning approaches, {\ie}, \textit{discrete prompts} \cite{Schick2020ExploitingCF}, \textit{continuous prompts} \cite{Li2021PrefixTuningOC} and \textit{priming} \cite{Brown2020LanguageMA}. Discrete prompts \cite{Schick2020ExploitingCF} reformat data instances with some template text, like, 

\begin{center}
``\textit{\{ premise \} Should we assume that \{ hypothesis \}? [prediction]}''.
\end{center}
Discrete prompts will typically tune all parameters of the model. On the other hand, continuous prompts \cite{Li2021PrefixTuningOC} will prepend examples with embedding vectors of special tokens, which will only update a much smaller set of model parameters. Very different from the discrete and continuous prompts, priming \cite{Brown2020LanguageMA} initially adopted in GPT-3 will prepend several priming examples to the target evaluation example instead, like

\begin{center}
 ``\textit{Example 1: \{ sentence 1 \} True or False? \{ label 1 \}.\\
 Example 2: \{ sentence 2 \} True or False? \{ label 2 \}.\\
  $\cdots$\\
 Example k:  \{ sentence k \} True or False? \{ label k \}.\\
 Evaluation: \{ eval-sentence \} True or False? [prediction]}.'' 
 \end{center}
According to the analysis reported in \cite{Webson2021DoPM}, discrete prompts works very well in few-shot tuning, continuous prompts have not yet reported success in few-shot setting yet, while priming is very costly and seems to work well for the largest GPT-3 (175B) model.

\section{Notation, Terminology Definition and Problem Formulation}\label{sec:formulate}

In this section, we will first introduce the notations used in this paper. After that, we will provide the definitions of several used terminologies used and the formulations of the graph reasoning tasks studied in this paper.

\subsection{Basic Notations}

In the sequel of this paper, we will use the lower case letters (e.g., $x$) to represent scalars, lower case bold letters (e.g., $\mathbf{x}$) to denote column vectors, bold-face upper case letters (e.g., $\mathbf{X}$) to denote matrices, and upper case calligraphic letters (e.g., $\mathcal{X}$) to denote sets or high-order tensors. Given a matrix $\mathbf{X}$, we denote $\mathbf{X}(i,:)$ and $\mathbf{X}(:,j)$ as its $i_{th}$ row and $j_{th}$ column, respectively. The ($i_{th}$, $j_{th}$) entry of matrix $\mathbf{X}$ can be denoted as $\mathbf{X}(i,j)$. We use $\mathbf{X}^\top$ and $\mathbf{x}^\top$ to represent the transpose of matrix $\mathbf{X}$ and vector $\mathbf{x}$. For vector $\mathbf{x}$, we represent its $L_p$-norm as $\left\| \mathbf{x} \right\|_p = (\sum_i |\mathbf{x}(i)|^p)^{\frac{1}{p}}$. The Frobenius-norm of matrix $\mathbf{X}$ is represented as $\left\| \mathbf{X} \right\|_F = (\sum_{i,j} |\mathbf{X}(i,j)|^2)^{\frac{1}{2}}$. The element-wise product of vectors $\mathbf{x}$ and $\mathbf{y}$ of the same dimension is represented as $\mathbf{x} \otimes \mathbf{y}$, whose concatenation is represented as $\mathbf{x} \sqcup \mathbf{y}$.

\subsection{Terminology Definitions}\label{subsec:definition}

In this paper, we will investigate the reasoning tasks on graph structured data. The graph datasets studied in this paper all come from different domains, which have very different structures and carry very different properties. Here, in this subsection, we will provide the general terminology definitions of these different graph structured data studied in this paper.

\begin{definition}
(Graph): Generally, the graph studied in this paper can be represented as $G = (\mc{V}, \mc{E})$. In the representation, notation $\mc{V} = \left\{v_1, v_2, \cdots, v_n \right\}$ denotes the set of $n$ nodes in the graph and $\mc{E} = \left\{e_{i,j} = (v_i, v_j) \right\}_{v_i, v_j \in \mc{V}}$ denotes the set of $m$ links among these nodes, where $|\mc{V}| = n$ and $|\mc{E}| = m$ are also normally called the \textit{order} and \textit{size} of the graph $G$, respectively.
\end{definition}

Depending on the application domains, the graph data to be studied may have very different property and structural information. For some graph, the nodes may carry some feature and label information, which can be represented via mappings $x: \mc{V} \to \mathbbm{R}^{d_x}$ and $y: \mc{V} \to \mathbbm{R}^{d_y}$, respectively. For each node $v_i \in \mc{V}$, we can represent its features as $x(v_i) = \mb{x}_{v_i} \in \mathbbm{R}^{d_x}$ and its label vector as $y(v_i) = \mb{y}_{v_i} \in \mathbbm{R}^{d_y}$, where $d_x$ and $d_y$ denote the feature and label space dimensions, respectively. If there are also features and labels attached to the links in graph $G$, we can also represent the corresponding feature and label vectors of link $e_{i,j} \in \mc{E}$ in a similar way as $\mb{x}_{e_{i,j}} \in \mathbbm{R}^{d_x}$ and $\mb{y}_{e_{i,j}} \in \mathbbm{R}^{d_y}$, respectively.

For the graph data from many domains, like \textit{bibliographic network}, \textit{online social network}, \textit{recommender systems}, \textit{knowledge graph}, there will exist one single large-scale graph structure in the dataset, but the graph may contain thousands, millions or even billions of nodes and links. Such large-scale graphs can be perfectly represented with the above definition. Meanwhile, for the graphs from many other domains, like the \textit{special graph structures} we learn from the discrete math course, and the \textit{bio-chemical molecular graphs}, there will exist a large number of much smaller graph instances in the dataset, and each graph instance normally contain tens or a few hundred nodes and links instead. To differentiate these two types of graph structured data, some existing work \cite{Zhang2019GraphNN} also names first categories of graphs as the \textit{giant networks} and calls the second categories of graphs as the \textit{small graph instances set}. Meanwhile, to represent the set of such small-sized graph instances, we introduce the concept of \textit{graph set} as follows.

\begin{definition}
(Graph Set): For the generated special graph instances (to be introduced in this paper) and the bio-chemical molecular graph instances, we can represent the set of graph instances in these datasets as $\mc{G} = \left\{g_1, g_2, \cdots, g_l \right\}$, where $g_i = (\mc{V}_{g_i}, \mc{E}_{g_i})$ denotes an individual graph instance and it can be represented according to the above graph definition.
\end{definition}

For some application domains, in the above graph set, each graph instance may also have its unique feature and label information, denoting its topological properties and tags of the graph instance. Formally, for a graph instance $g_i \in \mc{G}$ in the graph set $\mc{G}$, we can represent its raw feature and label vectors as $\mb{x}_{g_i} \in \mathbbm{R}^{d_x}$ and $\mb{y}_{g_i} \in \mathbbm{R}^{d_y}$, respectively.

\begin{figure*}
    \centering
    \begin{minipage}{\textwidth}
    	\includegraphics[width=\linewidth]{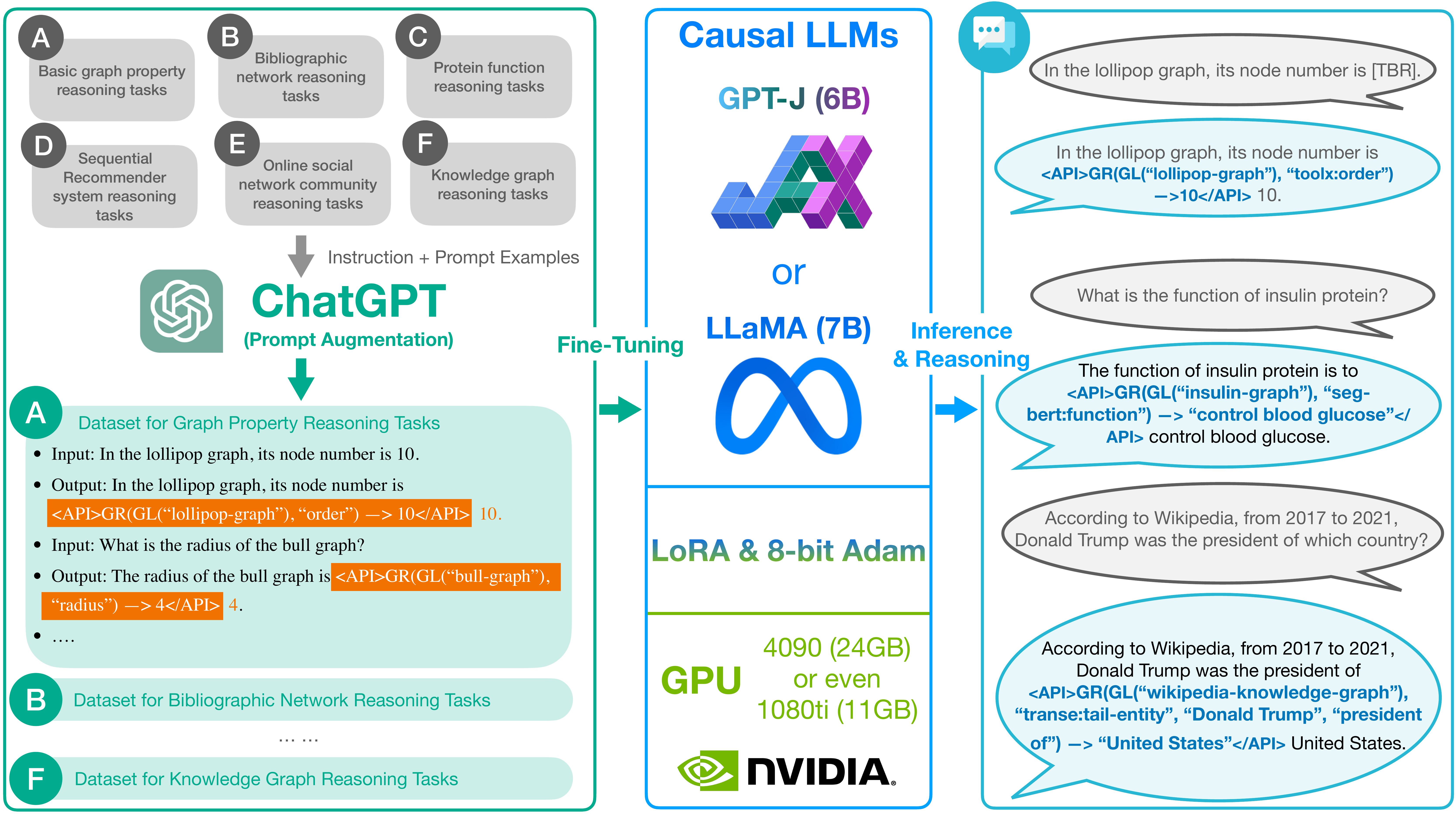}
        \caption{The Outline of the {\our} Framework. The framework has three main parts: (1) prompt data annotation and augmentation with ChatGPT, (2) existing pre-trained causal LLMs fine-tuning with the generated prompt dataset, and (3) inference of the fine-tuned model for adding graph reasoning API calls into statements.}
    	\label{fig:framework}
	\end{minipage}
\end{figure*}

\subsection{Problem Formulation}\label{subsec:formulation}

In this paper, we aim to empower the existing pre-trained LLMs to carry out graph reasoning tasks. As introduced before, the graph reasoning tasks studied in this paper include (1) \textit{basic graph property reasoning}, (2) \textit{bibliographic paper topic reasoning}, (3) \textit{bio-chemical molecular graph function reasoning}, (4) \textit{recommender system sequential recommendation reasoning}, (5) \textit{online social network community reasoning}, and (6) \textit{knowledge graph entity and relation reasoning}. Specifically, these graph reasoning tasks studied in this paper are carefully selected, which can be categorized into six types of the most fundamental graph learning problems listed as follows:

\begin{itemize}

\item \textit{Attribute Calculation}: For the tasks like \textit{basic graph property reasoning}, we actually aim to calculate either explicit or implicit attributes of the input graph data, ranging from the simple \textit{number of nodes/links} in the graph, to the \textit{graph radius} and \textit{diameter}, and the more complex \textit{graph periphery} and \textit{node pairwise short path length}.

\item \textit{Node Classification}: For the \textit{bibliographic paper topic reasoning} task, we aim to predict the topic of the academic papers in the bibliographic network, which can be modeled as the node classification task actually. Via the raw features of the paper nodes and their nearby neighboring nodes, we can classify the papers into different classes, which correspond to the specific topics of these papers.

\item \textit{Graph Classification}: For the \textit{bio-chemical molecular graph function reasoning} task, based on the molecular graph structures, we aim to infer the potential functions of the bio-chemical molecules, which can be defined as the graph instance classification task. Via both the molecular graph structure and raw attributes, we can classify the graph instances into different classes, which correspond to different pre-defined bio-chemical molecule functions.

\item \textit{Link Prediction}: For the \textit{sequential recommender system reasoning} task, based on the historical user-item interaction records, we aim to infer the potential preferences of users towards certain items in the system, which can be defined as the link prediction task (connecting user and item) in graph learning. Depending on the recommender system settings, we can either predict the link existence label denoting whether the user will be interested in the item or not, or infer the potential link weights denoting the rating scores that users will give to the items.

\item \textit{Graph Partition/Clustering}: For the \textit{online social network community reasoning} task, we aim to infer the community structures of online social networks, which can be defined as the graph partition/clustering task. Based on the user social interaction patterns, we want to partition the users in online social networks into different clusters, each of which denote one social community formed by the users with very frequent social interactions.
  
\item \textit{Graph Searching}: For the \textit{knowledge graph reasoning} task, we aim to infer the potential entities or relations based on the input parameters, which can be modeled as the graph searching problem. Starting from the input entity or relation, we aim to expand and search for the related entities or relations for generating the outputs, that can effectively preserve the desired semantics of the inputs.

\end{itemize}

To address these above diverse graph reasoning tasks with one single LLM, we propose to include the API calls of external graph learning tools into to the graph reasoning statements seamlessly. Based on the above notations, we will design a set of graph reasoning API calls for different graph tasks in the real-world. Such API calls include both the external graph learning tool name and the parameters, which will be surrounded with special tokens to differentiate from regular text. Based on a handful human-written prompt examples, with ChatGPT, we will generate a large language modeling prompt dataset containing such API calls, which will be used for fine-tuning the LLMs, like GPT-J and LLaMA. Such LLMs to be studied in this paper have all been pre-trained already and we will only fine-tune them with the generated prompt datasets. More information about the technical details on address these tasks will be introduced in the following methodology section.

{\sloppy


\section{Proposed Method}\label{sec:method}

In this section, we will introduce the {\our} framework proposed in this paper. At the beginning, in Section~\ref{subsec:framework_outline}, we will first briefly describe the {\our} framework outline for readers. After that, we will talk about the graph reasoning API call general representations in Section~\ref{subsec:api_call}, and introduce the specific graph reasoning task oriented API calls in Section~\ref{subsec:graph_reasoning_api_call}. Based on the hand-crafted graph reasoning prompt examples, we will introduce how to use the ChatGPT to augment the prompt dataset in Section~\ref{subsec:prompt_augmentation}. Detailed information about the language model fine-tuning with the augmented prompt datasets will be introduced in Section~\ref{subsec:llm_finetuning}. Meanwhile, to also allow {\our} to handle some basic Q\&A for graph reasoning, we will also introduce a few number of graph reasoning Q\&A prompts in Section~\ref{subsec:qa_prompt}, which will be merged into the statement prompts for LLM fine-tuning. Finally, based on the output statements with API calls generated by the language models, the graph reasoning tasks oriented API call parsing, execution, graph task reasoning and output post-processing will be introduced in Section~\ref{subsec:llm_inference_postprocessing}.


\subsection{Framework Outline}\label{subsec:framework_outline}

In Figure~\ref{fig:framework}, we provide an outline of the {\our} framework illustrating the internal functional components and pipeline of {\our}. According to the framework outline, based on the hand-crafted instructions and a handful number of prompt examples, we use ChatGPT to annotate and augment a large prompt dataset about graph reasoning API call statements. With the generated prompt dataset, we will fine-tune the existing pre-trained causal LLMs, such as GPT-J \cite{gpt-j, gpt-j-8bit} and LLaMA \cite{Touvron2023LLaMAOA} to teach them how to use the external graph reasoning tools. With both LoRA (Low-Rank Adaptation) \cite{Hu2021LoRALA} and 8-bit Adam and the model quantization techniques \cite{Dettmers20218bitOV}, {\our} can be fine-tuned on GPUs with very small memory space, such as Nvidia GeForce RTX 4090 (24GB RAM) and even Nvidia  GeForce RTX 1080Ti (11GB RAM). The fine-tuned {\our} will be used for inference purposes. Given the input query statements and questions, {\our} will add the corresponding graph reasoning API calls into the output statements at the most appropriate positions automatically. Detailed information about these mentioned components and steps in building {\our} will be introduced in detail in the following subsections.




\DeclareRobustCommand{\hlcyan}[1]{{\sethlcolor{cyan}\hl{#1}}}

\begin{table*}[t]
\caption{A summary of API call examples for basic graph Loading and property reasoning studied in this paper. In this table, we use notations $GL(\cdot)$ and $GR(\cdot)$ to represent the graph loading and graph reasoning API calls. Without introducing new special tokens to the pre-trained tokenizer of LLMs, we use ``['', ``]'' and ``-->'' to represent the ``${\bapi}$'', ``${\eapi}$'' and ``$\to$'' tokens introduced in this paper. Notation ``[TBR]'' denotes the ``to be reasoned'' placeholder token. We refer to the top left Lollipop graph (in green color) illustrated in Figure~\ref{fig:problem_illustration} as the ``Lollipop graph'' example in the table. In the graph property reasoning API call notations, we use $G_l$ to represent the result ob API call ``[GL(file-path:``./graphs/lollipop'') $\to$ $G_l$]'' and use the notation ``toolx:desired\_property'' to denote the reasoning of the desired properties with the toolx graph toolkit (to be introduced in the following experiment section).}\label{tab:api_call_summary_upper}
\centering
\small
\setlength{\tabcolsep}{3pt}
\resizebox{\textwidth}{!}{
\begin{tabular}{c | c | p{0.25\textwidth} | p{0.40\textwidth} }
\toprule
\hline
\multirow{2}{*}{\textbf{Tasks}}  & \multirow{2}{*}{\textbf{API Call Templates}}  & \multicolumn{2}{c}{\textbf{Prompt Examples}} \\
\cline{3-4}

&& \textbf{Inputs} & \textbf{Outputs} \\

\hline

\multirow{9}{*}{\makecell{Graph\\ Data\\ Loading} }
& \multirow{2}{*}{${GL(\textit{file-path})}$} 
&``The structure of the {\color{cyan}molecular graph of the benzene ring} contains a hexagon.'' 
&``The structure of the {\color{white}\hlcyan{[GL(file-path:``./graphs/benzene-ring'')]}} molecular graph of the benzene ring contains a hexagon.''  \\

\cline{2-4}
& \multirow{3}{*}{${GL(\textit{file-path}, \textit{node-subset}, \textit{link-subset})}$} 
&``There exist a {\color{cyan}carbon-oxygen double bond in the Acetaldehyde molecular graph}.'' 
&``There exist a {\color{white}\hlcyan{[GL(file-path:``./graphs/acetaldehyde'', node-subset:``all related nodes'', link-subse:\{(C=O)\})]}} carbon-oxygen double bond in the Acetaldehyde molecular graph.''  \\

\cline{2-4}
& \multirow{2}{*}{${GL(\textit{file-path})\to r}$} 
&\multirow{2}{*}{``{\color{cyan}Lollipop graph} looks like a spoon.''}
&``{\color{white}\hlcyan{[GL(file-path:``./graphs/lollipop'') --> $G_l$]}} Lollipop graph looks like a spoon.''  \\

\cline{2-4}
\hline
\multirow{25}{*}{\makecell{Graph\\ Property\\ Reasoning}}
&\multirow{2}{*}{$GR(graph, \text{``} order \text{''}) \to r $}
&``There exist {\color{cyan}[TBR] nodes} in the {\color{cyan}lollipop graph}.'' 
&``There exist {\color{white}\hlcyan{[GR($G_l$, ``toolx:order'') --> $10$]}} nodes in the lollipop graph.'' \\

\cline{2-4}
&\multirow{2}{*}{$GR(graph, \text{``} size \text{''}) \to r$}
&``Via {\color{cyan}[TBR] links}, nodes in the {\color{cyan}lollipop graph} are all connected.'' 
&``Via {\color{white}\hlcyan{[GR($G_l$, ``toolx:size'') --> $12$]}} links, nodes in the example lollipop graph are all connected.'' \\

\cline{2-4}
&\multirow{2}{*}{$GR(graph, \text{``} density \text{''}, is\text{-}directed) \to r$}
&``The {\color{cyan}undirected lollipop graph} has a {\color{cyan}density of $\frac{4}{15}$}.'' 
&``The undirected lollipop graph has a density of {\color{white}\hlcyan{[GR($G_l$, ``toolx:density'', $is\text{-}directed$:False) --> $\frac{4}{15}$]}}.'' \\

\cline{2-4}
&\multirow{3}{*}{$GR(graph, \text{``} eccentricity \text{''}) \to r$}
&``The long `tail' will lead to large {\color{cyan}eccentricity [TBR]} for many nodes in the {\color{cyan} lollipop graph}.''
&``The long `tail' will lead to large eccentricity {\color{white}\hlcyan{[GR($G_l$, ``toolx:eccentricity'') --> $\{0: 7, 1: 7, 2: 7, 3: 6, 4: 5, 5: 4, 6: 4, 7: 5, 8: 6, 9: 7\}$]}} for many nodes in the lollipop graph.''
\\

\cline{2-4}
&\multirow{2}{*}{$GR(graph, \text{``} eccentricity \text{''}, \text{node-subset}) \to r$}
&``The {\color{cyan}eccentricity of node \#$4$} in the {\color{cyan} lollipop graph is [TBR]}.''
&``The eccentricity of node \#$4$ in the lollipop graph is {\color{white}\hlcyan{[GR($G_l$, ``toolx:eccentricity'', node \#4) --> $5$]}}.''
\\

\cline{2-4}
&\multirow{2}{*}{$GR(graph, \text{``} radius \text{''}) \to r$}
&``The {\color{cyan}radius} of the {\color{cyan} lollipop graph is [TBR]}.''
&``The radius of the lollipop graph is {\color{white}\hlcyan{[GR($G_l$, ``toolx:radius'') --> $4$]}}.''
\\

\cline{2-4}
&\multirow{2}{*}{$GR(graph, \text{``} center \text{''}) \to r$}
&``The {\color{cyan}center of the lollipop graph} include node(s) {\color{cyan}[TBR]}.''
&``The center of the lollipop graph include node(s) {\color{white}\hlcyan{[GR($G_l$, ``toolx:center'') --> $\{5, 6\}$]}}.''
\\

\cline{2-4}
&\multirow{3}{*}{$GR(graph, \text{``} shortest\text{-}path \text{''}, node_1, node_2) \to r$}
&``In the {\color{cyan}lollipop graph}, the {\color{cyan}length of shortest path} between {\color{cyan}node \#1 and node \#5 is [TBR]}.''
&``In the lollipop graph, the length of shortest path between node \#1 and node \#5 is {\color{white}\hlcyan{[GR($G_l$, ``toolx:shortest-path'', node \#1, node \#5) --> $3$]}}.''
\\

\cline{2-4}
&\multirow{3}{*}{$GR(graph, \text{``} avg\text{-}shortest\text{-}path \text{''}) \to r$}
&``The {\color{cyan}average length of shortest path} for all nodes in the {\color{cyan}lollipop graph is [TBR]}.''
&``The average length of shortest path for all nodes in the lollipop graph is {\color{white}\hlcyan{[GR($G_l$, ``toolx:avg-shortest-path'') --> $2.86$]}}.''
\\

\cline{2-4}
&\multirow{2}{*}{$GR(graph, \text{``} diameter \text{''}) \to r$}
&``The {\color{cyan}diameter} of the {\color{cyan} lollipop graph is [TBR]} due to the long `tail'.''
&``The diameter of the lollipop graph is {\color{white}\hlcyan{[GR($G_l$, ``toolx:diameter'') --> $7$]}} due to the long `tail'.''
\\

\cline{2-4}
&\multirow{2}{*}{$GR(graph, \text{``} periphery \text{''}) \to r$}
&``The {\color{cyan}periphery} of the {\color{cyan}lollipop graph} includes the nodes {\color{cyan}[TBR]}.''
&``The periphery of the lollipop graph includes the nodes {\color{white}\hlcyan{[GR($G_l$, ``toolx:periphery'') --> $\{0, 1, 2, 9\}$]}}.''
\\
\hline
\bottomrule
\end{tabular}
}
\end{table*}



\subsection{Prompts with API Calls}\label{subsec:api_call}

In this paper, we can represent the API calls of external graph learning tools as $f(args)$, where $f$ is the external tool function name and $args$ denotes the list of parameters of the function. For representation simplicity, we can also represent such an API call as a tuple notation $c = (f, args)$, which will be frequently used in the following part of this paper. Instead of merely generating the external API calls as the output, we propose to inset the API calls into the generated output statements instead, which allows the LLMs to handle and respond the graph reasoning tasks with regular conversations via texts.

Formally, to insert the API calls into the output statements, we can represent the sequence of tokens for API call $c = (f, args)$ as
\begin{equation}
\mb{s}(c) = {{\bapi} f(args) {\eapi}},
\end{equation}
or
\begin{equation}
\mb{s}(c, r) = {{\bapi} f(args) \to r {\eapi}},
\end{equation}
where both ``{\bapi}'' and ``{\eapi}'' surrounding the API call function are the special tokens to differentiate it from other tokens in the generated output statements. For the {\our} framework, when it generates the ``{\bapi}'' and ``{\eapi}'' tokens, the framework parser will recognize that the tokens inside it denotes the API function call. As to the second API call representation, the notation $r$ denotes the return result of $f(args)$. For the API calls with notations ``$\to r$'', {\our} will also replace and insert the API call results into the statements after query parsing and execution; otherwise, the API calls will be executed at the backend with return results recorded into the working memory instead. Detailed information about the output statement API call parsing, execution and post-processing will be introduced later in Section~\ref{subsec:llm_inference_postprocessing}. Depends on both the function $f(args)$ and the contexts for the API call in the reasoning task, the LLMs to be fine-tuned later will automatically decide whether the output results will be inserted into the output statement.

Different from the very simple API calls ({\eg} ``\textit{Calendar}'', ``\textit{Calculator}'' and ``\textit{WikiSearch}'') studied in \cite{Schick2023ToolformerLM}, in graph reasoning, some of the API calls may involve complicated and nested calls of various external functions. For instance, some of the parameters in one API call can actually be the returning results of other API calls, or we may need to call multiple sequential APIs concurrently for accomplishing one graph reasoning task. In the following Section~\ref{subsec:graph_reasoning_api_call}, when discussing about the specific graph reasoning tasks, we will encounter some of such complicated graph reasoning API calls. 

To address such complicated graph reasoning tasks, in this paper, we will also allow {\our} to generated nested and sequential API calls surrounded by the special tokens ``{\bapi}'' and ``{\eapi}''. For instance, given two API calls $c_1 = (f_1, args_1)$ and $c_2 = (f_2, args_2)$ with their own input parameters, the first API function $f_1$ needs to use the return result of the second API function $f_2$ as its input parameter, we can represent such nested API calls as
\begin{equation}
\mb{s}(c_1 | c_2) = {{\bapi} f_1(args_1=f_2(args_2)) {\eapi}},
\end{equation}
or just simply as
\begin{equation}
\mb{s}(c_1 | c_2) = {{\bapi} f_1(f_2(args_2)) {\eapi}},
\end{equation}
where the notation ``$c_1 | c_2$'' denotes these two API calls are nested.

Meanwhile, if a task needs to call multiple sequential APIs simultaneously, {\eg} $c_1 = (f_1, args_1)$ and $c_2 = (f_2, args_2)$, we can represent such sequential API calls as
\begin{align}
&\mb{s}(c_1, c_2) = {{\bapi} f_1(args_1), f_2(args_2) {\eapi}}\\
&= {{\bapi} f_1(args_1) {\eapi}, {\bapi} f_2(args_2) {\eapi}} \\
&= \mb{s}(c_1), \mb{s}(c_2),
\end{align}
which is equivalent to two sequential API calls of $c_1$ and $c_2$ as well.

Meanwhile, for some even more complicated graph reasoning cases, we can rewrite the above API call representations with either more input parameters denoted by other API calls or with deeply nested API calls instead, {\eg}
\begin{equation}
\mb{s}\left(c_1 | (c_2, c_3) \right) = {{\bapi} f_1\left(f_2(args_2), f_3(args_3) \right) {\eapi}},
\end{equation}
or
\begin{equation}
\mb{s}\left(c_1 | (c_2 | c_3)\right) = {{\bapi} f_1\left(f_2 \left(f_3(args_3) \right) \right) {\eapi}},
\end{equation}
where $c_3 = (f_3, args_3)$ denotes the notation of a third API call.

Such graph reasoning function API calls will be inserted into statements for LLMs fine-tuning later. Without modifying the LLMs' vocabulary set and the pre-trained tokenizer, in implementation, we can replace the special tokens ``{\bapi}'', ``{\eapi}'' and $\to$ with some less frequently used tokens like ``['', ``]'' and ``->'' instead. In this paper, we will study several very different graph reasoning tasks involving diverse graph learning API calls, which will be introduced in detail in the following subsection for readers.


\subsection{Graph Reasoning Oriented Prompts}\label{subsec:graph_reasoning_api_call}

We will study several graph reasoning tasks in this paper with {\our}, which include both the very basic tasks, like the general graph property reasoning, and more advanced ones, like the reasoning tasks on graphs from different specific application domains. As introduced before in Section~\ref{sec:formulate}, these graph reasoning tasks studied in this paper are all carefully selected, which can be categorized into different types of fundamental graph learning tasks, {\eg} \textit{graph attribute calculation}, \textit{node classification}, \textit{graph classification}, \textit{link prediction}, \textit{graph partition/clustering} and \textit{graph searching}. All these fundamental graph learning tasks have extensive applications in real-world graph data reasoning tasks. Besides the tasks studied in this paper, with minor changes to the API calls, we can also apply the {\our} to other graph reasoning related application tasks as well.


\subsubsection{Graph Data Loading} 

Different from texts and images, the graph data we have in the real-world may have a relatively larger size, extensively connected structures and complex raw attributes. Except for some small-sized hand-crafted graph examples, it is almost impossible to manually type in the graph structured data as a sequence of token inputs to LLMs for reasoning. Therefore, in this paper, we propose to empower the {\our} model with the ability to automatically load the desired graph data from offline files or online repositories based on the provided the dataset name, local file path or online repository URL link.

Technically, the first API call that we will introduce in this paper is for graph data loading, which can load either the whole graph or just a subgraph involving one or a few nodes and links. Specifically, we can represent the graph loading API call as
\begin{equation}
{{\bapi} GL(\textit{file-path}, \textit{node-subset}, \textit{link-subset}) \to G {\eapi}},
\end{equation}
where ``$GL()$'' denotes abbreviation of the ``Graph Loading'' function name, and the function parameters ``\textit{file-path}'', ``\textit{node-subset}'' and ``\textit{link-subset}'' specify the local graph data file path (or the online repository URL if the data is stored on the web), subsets of specific nodes and links, respectively. The notation ``$\to G$'' explicitly represents the loaded graph data with the reference variable $G = (\mc{V}, \mc{E})$, which is optional actually depending on the application task and settings. What's more, if the local file directory or the online repository root URL has been pre-provided to the ``$GL()$'' function already, then we can just simplify the ``\textit{file-path}'' with the specific ``graph-name'' instead when calling this API function. 

Furthermore, when the parameters ``\textit{node-subset}'' and ``\textit{link-subset}'' are either omitted or assigned with the strings ``all nodes'' and ``all links'', respectively, then the API function call will just load the whole graph. For some cases, we can only specify the subset of nodes to be loaded ({\eg} $\{v_i, v_j, \cdots, v_k\} \subset \mc{V}$ in the graph) but cannot enumerate all the related links, we can just assign the ``\textit{node-subset}'' and ``\textit{link-subset}'' parameters with values ``$\{v_i, v_j, \cdots, v_k\}$'' and ``all related links'' (or the ``\textit{link-subset}'' parameter is just omitted). It will provide us with more flexibility in loading sub-graphs based on the provided node set and their internal links. Similarly, we can also only specify the subset of links, by assigning the ``\textit{node-subset}'' with ``all related nodes'' or just omitted it, it will automatically load the nodes composing those provided links in the graph data, like the second graph data loading prompt example shown in Table~\ref{tab:api_call_summary_upper}.

Besides that example, as shown at the top part of Table~\ref{tab:api_call_summary_upper}, we also provide a few other prompt examples of the graph data loading API calls, which can retrieve and load the requested graph data from the (local) files according to the input textual statements.

\subsubsection{Graph Property Reasoning}\label{subsubsec:graph_property_reasoning}

Graph structured data may have various properties, such as \textit{diameter}, \textit{density}, \textit{center} and \textit{shortest path}, which can capture different characteristics of the graph data and have extensive applications in real-world graph structured data. For reasoning such graph properties, it usually requires the model to not only know the property definitions but also has very strong logic reasoning and mathematical calculation abilities to compute such properties. For the existing language models, either \textit{masked language models} or \textit{autoregressive language models}, it will be very hard (almost impossible) for them to conduct the reasoning process for such complex properties based on the input graphs.

In this paper, to empower LLMs with the graph property reasoning ability, we introduce a group of external APIs, which can be called by the language models for reasoning about those properties. To illustrate how {\our} handles such graph property reasoning tasks, we will use the small-sized lollipop graph shown in Figure~\ref{fig:problem_illustration} (the top-left graph in green color) as an example in this part, which can be loaded via the following API calls as introduced before:
\begin{equation}\label{equ:lollipop_load}
{\bapi}GL(\text{``} lollipop \text{''}) \to G_l {\eapi},
\end{equation}
where the loaded the graph can also be referred to by notation $G_l$. For simplicity, in the following part, we will also use the above loaded lollipop graph $G_l$ as an example to introduce the graph property reasoning APIs for readers.


\vspace{10pt}

\noindent \textbf{Order and Size}: Formally, given a graph, like the loaded lollipop graph $G_l = (\mc{V}, \mc{E})$, its \textit{order} denotes the number of nodes in the graph, {\ie} $|\mc{V}|$, and its \textit{size} is the number of links in the graph, {\ie} $|\mc{E}|$. We can represent the API calls for reasoning the \textit{order} and \textit{size} properties of the lollipop graph as
\begin{align}
&{{\bapi} GR(GL(\text{``} lollipop \text{''}), \text{``} toolx\text{:}order \text{''}) \to r {\eapi}},\\
&{{\bapi} GR(GL(\text{``} lollipop \text{''}), \text{``} toolx\text{:}size \text{''}) \to r {\eapi}}.
\end{align}
If the lollipop graph has been pre-loaded via other API calls already and can be referred to as $G_l$, the above API calls can also be simplified as follows:
\begin{align}
&{{\bapi} GR(G_l, \text{``} toolx\text{:}order \text{''}) \to r {\eapi}},\\
&{{\bapi} GR(G_l, \text{``} toolx\text{:}size \text{''}) \to r {\eapi}},
\end{align}
where the notation $GR()$ denotes the abbreviated ``Graph Reasoning'' function name and the parameters ``order'' and ``size'' represent the graph properties to be reasoned. The notation ``\textit{toolx\text{:}desired\_property}'' denotes the desired graph property reasoning with the \textit{toolx} toolkit. The \textit{toolx} is a graph property calculation toolkit created in this paper for {\our} based on the networkx, and we will introduce more information about the graph reasoning models and toolkits used in this paper in the next experiment section instead. The notation ``$\to r$'' specifies the output result $r$ by the graph property reasoning API call to be included into the output statements. As introduced before, the returning output result tag ``$\to r$'' of the API calls is actually optional, inclusion of which depends on both the reasoning context and application task.

\vspace{10pt}

\noindent \textbf{Density}: Graph \textit{density} denotes the ratio of existing links in a graph compared with the maximal number of potential links among nodes in a graph. If the input lollipop graph $G_l = (\mc{V}, \mc{E})$ is \textit{directed}, its \textit{density} can be represented as $\frac{|\mc{E}|}{|\mc{V}| (|\mc{V}| - 1)}$; while if $G_l$ is \textit{undirected}, its \textit{density} can be represented as $\frac{2|\mc{E}|}{|\mc{V}| (|\mc{V}| - 1)}$. Formally, the API calls that can be used for computing the density of graph can be represented as follows:
\begin{equation}
{{\bapi} GR(G_l, \text{``} toolx\text{:}density \text{''}, is\text{-}directed) \to r {\eapi}},
\end{equation}
where the boolean ``\textit{is\text{-}directed}'' parameter differentiates directed graph from undirected ones in the density calculation.


\vspace{10pt}

\noindent \textbf{Shortest Path}: The \textit{shortest path} between two nodes in a graph is a path of shortest possible length connecting them via the nodes and links in the graph. The API call for reasoning the length of the \textit{shortest path} from $node_1$ to $node_2$ in a graph can be represented as
\begin{align}
&{{\bapi} GR(G_l, \text{``} toolx\text{:}shortest\text{-}path \text{''}, node_1, node_2) \to r {\eapi}}.
\end{align}
Meanwhile, the average length of \textit{shortest path} for all nodes in the graph can be obtained via the following API call instead
\begin{align}
&{{\bapi} GR(G_l, \text{``} toolx\text{:}avg\text{-}shortest\text{-}path \text{''}) \to r {\eapi}}.
\end{align}
Besides the average \textit{shortest path length}, we can also reason for the largest \textit{shortest path length} and the smallest \textit{shortest path length} of a graph as follows:
\begin{align}
&{{\bapi} GR(G_l, \text{``} toolx\text{:}max\text{-}shortest\text{-}path \text{''}) \to r {\eapi}},\\
&{{\bapi} GR(G_l, \text{``} toolx\text{:}min\text{-}shortest\text{-}path \text{''}) \to r {\eapi}}.
\end{align}


\vspace{10pt}

\noindent \textbf{Eccentricity}: Given a connected graph, like the lollipop graph $G_l = (\mc{V}, \mc{E})$, for node $v_i \in \mc{V}$, its \textit{eccentricity} denotes the maximum graph distance between $v_i$ and any other node $v_j \in \mc{V}$ in the graph. According to such a definition, for disconnected graph, all nodes are defined to have infinite \textit{eccentricity}. We can compute the \textit{eccentricity} either for the whole graph ({\ie} for all nodes in the graph) or for specific node(s) via the following two API calls:
\begin{align}
&{{\bapi} GR(G_l, \text{``} toolx\text{:}eccentricity \text{''}) \to r {\eapi}},\\
&{{\bapi} GR(G_l, \text{``} toolx\text{:}eccentricity \text{''}, \text{node-subset}) \to r {\eapi}}.
\end{align}


\vspace{10pt}

\noindent \textbf{Diameter}: The \textit{diameter} of a graph denotes the ``longest shortest path'' between any two nodes in the graph, whose API call can be represented as
\begin{align}
&{{\bapi} GR(G_l, \text{``} toolx\text{:}diameter \text{''}) \to r {\eapi}},
\end{align}
whose result will be equal to the result of the above API call ${{\bapi} GR(G_l, \text{``} max\text{-}shortest\text{-}path \text{''}) {\eapi}}$ actually.


\vspace{10pt}

\noindent \textbf{Radius}: Graph \textit{radius} denotes the is the minimum graph \textit{eccentricity} of any node in a graph. A disconnected graph therefore has infinite radius. The API call for computing a graph \textit{radius} can be represented as
\begin{align}
&{{\bapi} GR(G_l, \text{``} toolx\text{:}radius \text{''}) \to r {\eapi}}.
\end{align}


\vspace{10pt}

\noindent \textbf{Center}: Formally, the \textit{center} of a graph denotes the set of nodes whose eccentricity is equal to the graph radius. The API call for identifying a graph \textit{center} can be represented as
\begin{align}
&{{\bapi} GR(G_l, \text{``} toolx\text{:}center \text{''}) \to r {\eapi}}.
\end{align}


\vspace{10pt}

\noindent \textbf{Periphery}: The \textit{periphery} of a graph is the subgraph of the graph induced by nodes that have the \textit{eccentricities} equal to the graph diameter, whose API call can be represented as
\begin{align}
&{{\bapi} GR(G_l, \text{``} toolx\text{:}periphery \text{''}) \to r {\eapi}}.
\end{align}


\vspace{10pt}

\noindent \textbf{A Summary of Basic Graph Reasoning API Calls}: According to our descriptions above, the readers should have observed that those graph properties may require very complex logic reasoning. Via some preliminary experimental testings, the current LLMs (such as ChatGPT and LLaMA) cannot handle them very well. At the same time, the above reasoning properties, like the \textit{shortest-path}, also have very extensive applications in the real-world graph reasoning tasks, {\eg} \textit{traffic network reasoning} and \textit{traffic route planning}. Currently, the LLMs have been criticized since they cannot provide the correct reasoning results for the spatial traffic data, {\eg} estimating the traveling distance and time between different locations. Equipped with the above shortest path based property API calls on traffic networks, we will be able to provide more precise reasoning results for LLMs in handling such queries. To incorporate them into language models, we also show some examples of the above API calls in Table~\ref{tab:api_call_summary_upper}, which can load the graph data from specified data sources and conduct the reasoning of some general graph properties as discussed above. 




\DeclareRobustCommand{\hlcyan}[1]{{\sethlcolor{cyan}\hl{#1}}}

\begin{table*}[t]
\caption{A summary of API call examples for advanced graph reasoning tasks studied in this paper. In this table, we use notations $GL(\cdot)$ and $GR(\cdot)$ to represent the graph loading and graph reasoning API calls. Similarly, we use ``['', ``]'' and ``-->'' to represent the ``${\bapi}$'', ``${\eapi}$'' and ``$\to$'' tokens, and use notation ``[TBR]'' to denote the ``to be reasoned'' placeholder token.}\label{tab:api_call_summary_lower}
\centering
\small
\setlength{\tabcolsep}{3pt}
\resizebox{\textwidth}{!}{
\begin{tabular}{c | c | p{0.28\textwidth} | p{0.40\textwidth} }
\toprule
\multirow{2}{*}{\textbf{Tasks}}  & \multirow{2}{*}{\textbf{API Call Templates}}  & \multicolumn{2}{c}{\textbf{Prompt Examples}} \\
\cline{3-4}

&& \textbf{Inputs} & \textbf{Outputs} \\

\hline


\multirow{15}{*}{\makecell{Bibliographic\\ Paper Topic\\ Reasoning}}
&\multirow{15}{*}{$GR(graph, \text{``} topic \text{''}, paper\text{-}node) \to r$}
&In the {\color{cyan}core bibliographic network}, {\color{cyan}paper \#31366} focuses on the {\color{cyan}topic of [TBR]}.
&In the core bibliographic network, paper \#31366 focuses on the topic of {\color{white}\hlcyan{[GR(GL(``cora''), ``graph-bert:topic'', paper\#31366)-->Neural Networks]}}.\\
\cline{3-4}
&&Within {\color{cyan}cora}, {\color{cyan}paper \#13195} is dedicated to the {\color{cyan}study of [TBR]}.
&Within cora, paper \#13195 is dedicated to the study of {\color{white}\hlcyan{[GR(GL(``cora''), ``graph-bert:topic'', paper\#13195)-->Reinforcement Learning]}}.\\
\cline{3-4}
&&The {\color{cyan}citeseer bibliographic network's paper \#2} is concerned with the {\color{cyan}area of [TBR]}.
&The citeseer bibliographic network's paper \#2 is concerned with the area of {\color{white}\hlcyan{[GR(GL(``citeseer''), ``graph-bert:topic'', paper\#2)-->Agents]}}.\\
\cline{3-4}
&&{\color{cyan}Paper \#3 in the citeseer network} investigates the {\color{cyan}field of [TBR]}.
&Paper \#3 in the citeseer network investigates the field of {\color{white}\hlcyan{[GR(GL(``citeseer''), ``graph-bert:topic'', paper\#3)-->DB]}}.\\
\cline{3-4}
&&{\color{cyan}Paper \#7}, situated {\color{cyan}in the pubmed bibliographic network}, is centered around {\color{cyan}the [TBR] topic}.
&Paper \#7, situated in the pubmed bibliographic network, is centered around the {\color{white}\hlcyan{[GR(GL(``pubmed''), ``graph-bert:topic'', paper\#7)-->1]}} topic.\\

\hline
\multirow{14}{*}{\makecell{Protein\\ Function\\ Reasoning}}
&\multirow{14}{*}{$GR({graph}, \text{``} protein\text{-}function \text{''}, g_i) \to r$}
&The {\color{cyan}protein molecular graph instance \#63 in the PROTEIN dataset} has {\color{cyan}a function of [TBR]} for the disease.
&The protein molecular graph instance \#63 in the PROTEIN dataset has a function of {\color{white}\hlcyan{[GR(GL(``PROTEIN''), ``seg-bert:molecule{-}function'', instance\#63)-->0]}} for the disease.\\
\cline{3-4}
&&In {\color{cyan}PROTEIN}, {\color{cyan}instance \#985} of the protein molecular graph demonstrates {\color{cyan}a function of [TBR]} for the disease.
&In PROTEIN, instance \#985 of the protein molecular graph demonstrates a function of {\color{white}\hlcyan{[GR(GL(``PROTEIN''), ``seg-bert:molecule{-}function'', instance\#63)-->1]}} for the disease.\\
\cline{3-4}
&&The chemical {\color{cyan}molecular graph numbered 63 in PTC} is characterized by {\color{cyan}a function of [TBR]}.
&The chemical molecular graph numbered 63 in PTC is characterized by a function of {\color{white}\hlcyan{[GR(GL(``PTC''), ``seg-bert:molecule{-}function'', instance\#63)-->1]}}.\\
\cline{3-4}
&&For chemical {\color{cyan}molecular graph instance \#63 in NCI1}, its {\color{cyan}function is [TBR]}.
&For chemical molecular graph instance \#63 in NCI1, its function is {\color{white}\hlcyan{[GR(GL(``NCI1''), ``seg-bert:molecule{-}function'', instance\#63)-->0]}}.\\
\cline{3-4}
&&The {\color{cyan}molecular graph of chemical compound \#121 in MUTAG} possesses {\color{cyan}a function of [TBR]}.
&The molecular graph of chemical compound \#121 in MUTAG possesses a function of {\color{white}\hlcyan{[GR(GL(``MUTAG''), ``seg-bert:molecule{-}function'', instance\#121)-->2]}}.\\

\hline
\multirow{10}{*}{\makecell{Sequential \\Recommender\\ System\\ Reasoning} }
&\multirow{10}{*}{$GR({graph}, \text{``}recommendation\text{''}, u_j, i_l) \to r$}
&In the {\color{cyan}Amazon recommender system}, {\color{cyan}user \#A240ORQ2LF9LUI} rates {\color{cyan}item \#0077613252} with {\color{cyan}a score of [TBR]}.
&In the Amazon recommender system, user \#A240ORQ2LF9LUI rates item \#0077613252 with a score of {\color{white}\hlcyan{[GR(GL(``amazon''), ``bpr:recommendation'', user\#A240ORQ2LF9LUI, item\#0077613252)-->4.0]}}.\\
\cline{3-4}
&&{\color{cyan}Within Last.fm}, {\color{cyan}user \#2} awards {\color{cyan}item \#52} with {\color{cyan}a [TBR] tag}.
&Within Last.fm, user \#2 awards item \#52 with a {\color{white}\hlcyan{[GR(GL(``last.fm''), ``bpr:recommendation'', user\#2, item\#52)-->41]}} tag.\\
\cline{3-4}
&&{\color{cyan}User \#196} gives {\color{cyan}a rating of [TBR]} to {\color{cyan}item \#251 at MovieLens}.
&User \#196 gives a rating of {\color{white}\hlcyan{[GR(GL(``movielens''), ``bpr:recommendation'', user\#196, item\#251)-->3]}} to item \#251 at MovieLens.\\

\hline
\multirow{11}{*}{\makecell{Online\\ Social Network\\ Reasoning} }
&\multirow{11}{*}{$GR({graph}, \text{``}community\text{''}) \to r$}
&In the {\color{cyan}academic collaboration network dblp}, {\color{cyan}scholar \#355233} is involved in {\color{cyan}[TBR] local community} formed by his/her collaborators.
&In the academic collaboration network dblp, scholar \#355233 is involved in {\color{white}\hlcyan{[GR(GL(``dblp''), ``kmeans:community-count'', scholar\#355233)-->6]}} local community formed by his/her collaborators.\\
\cline{3-4}
&&In the {\color{cyan}email communication social network}, there exist {\color{cyan}a number of [TBR] local communities} formed by users.
&In the email communication social network, there exist a number of {\color{white}\hlcyan{[GR(GL(``email''), ``kmeans:community-count'')-->42]}} local communities formed by users.\\
\cline{3-4}
&&The {\color{cyan}video sharing social network youtube} houses the {\color{cyan}largest user-formed local community}, which {\color{cyan}consists of [TBR] users}.
&The video sharing social network youtube houses the largest user-formed local community, which consists of {\color{white}\hlcyan{[GR(GL(``youtube''), ``kmeans:max-community-size'')-->3001]}} users.\\

\hline
\multirow{7}{*}{\makecell{Knowledge\\ Graph\\ Reasoning} }
&\multirow{7}{*}{$GR({graph}, reasoning\text{-}type, inputs) \to r$}
&According to the {\color{cyan}Freebase knowledge graph}, {\color{cyan}the relation between entity /m/027rn and entity /m/06cx9 is [TBR]}.
&According to the Freebase knowledge graph, the relation between entity /m/027rn and entity /m/06cx9 is {\color{white}\hlcyan{[GR(GL(``freebase''), ``transe:relation'', entity:/m/027rn, entity:/m/06cx9)-->/location/country/form\_of\_government]}}.\\
\cline{3-4}
&&According to the {\color{cyan}WordNet knowledge graph}, {\color{cyan}from entity plaything.n.01, via relation \_hyponym, we can derive entity [TBR]}.
&According to the WordNet knowledge graph, from entity plaything.n.01, via relation \_hyponym, we can derive entity {\color{white}\hlcyan{[GR(GL(``freebase''), ``transe:tail-entity'', entity:plaything.n.01, relation:\_hyponym)-->swing.n.02]}}.\\
\bottomrule
\end{tabular}
}
\end{table*}



\subsubsection{Advanced Graph Reasoning Tasks}

Besides the basic graph property reasoning tasks, we will also study several advanced reasoning tasks on real-world graph data with more complex structures in this paper, which include (1) \textit{academic paper topic reasoning on bibliographic network}, (2) \textit{protein function reasoning based on protein graph structures}, (2) \textit{sequential product recommendation reasoning based on recommender systems}, (4) \textit{social community reasoning from online social networks} and (5) \textit{semantics reasoning on knowledge graphs}. 

For many other advanced graph reasoning tasks not studied in this paper, via very minor changes to the {\our} framework, they can also be effectively incorporated into {\our} as well by adding the corresponding API calls into the reasoning prompts. The {\our} framework can serve as the backbone for hosting various graph reasoning application tasks with LLMs as the general interface. In Section~\ref{sec:future_work}, we will also describe some potential future research opportunities for the readers at the very end of this paper.


\vspace{10pt}

\noindent \textbf{Bibliographic Paper Topic Reasoning}: Bibliographic network \cite{10.14778/3402707.3402736} defines a complex graph structured data involving diverse entities, such as academic papers, authors, institutions and publication venues, as well as diverse links among these entities, such as the citation links, authorship links, affiliation links and publication links. In this part, we will discuss about the academic paper topic reasoning task based on the bibliographic network. The topics of a paper can be inferred with not only its own textual descriptions but also the other papers cited by/citing it, which requires the graph reasoning model to utilize both the raw textual features of the papers and the extensive citation links among the papers.

Formally, based on the terminology definition provided in the previous Section~\ref{subsec:definition}, we can represent the \textit{bibliographic network} as $G = (\mc{V}, \mc{E})$, which can be loaded via the API call 
\begin{equation}
{\bapi}GL(\text{``}bibliographic\text{-}network\text{''}) \to G {\eapi}.
\end{equation}

Each paper is represented as a node $v_i \in \mc{V}$ in the bibliographic network, which has both its raw feature vector $\mb{x}_{v_i}$ and label vector $\mb{y}_{v_i}$. The raw feature vector includes the textual information about the paper (like its title or abstract), and its label vector indicates the topics of the paper. Existing graph neural networks (GNNs) infer the paper topics by learning their representations with both raw features and connected neighbors' information \cite{Kipf_Semi_CORR_16, Zhang2020GraphBertOA}, which can be further used to infer the topic label vector. For the {\our} model introduced in this paper, we will use the pre-trained Graph-Bert \cite{Zhang2020GraphBertOA} as the default topic inference model for bibliographic networks. Based on the above descriptions, we can represent the paper topic reasoning via graph neural network model with the following API call:
 \begin{equation}
 {{\bapi} GR(G, \text{``} graph\text{-}bert\text{:}topic \text{''}, paper\text{-}node) \to r {\eapi}}.
 \end{equation}
 The function notation ``$GR(\cdot, \text{``} graph\text{-}bert\text{:}topic \text{''}, \cdot)$'' denotes it is a paper topic reasoning API with the Graph-Bert model \cite{Zhang2020GraphBertOA}. Actually, the {\our} framework proposed in this paper is a general framework. Besides the Graph-Bert model, many other existing graph neural network models can also be used here for academic paper topic inference as well. Based on the provided source code, the readers can customize the {\our} to include more different graph models that can be used for accomplishing their own graph reasoning tasks.
 

\vspace{10pt}

\noindent \textbf{Protein Molecule Function Reasoning}: Protein and chemical molecule function inference \cite{doi:10.1142/S0219633602000117} has been a classic problem studied in bio-chemical research for decades, which has fundamental applications in the real-world, such as helping design some new drugs for curing some existing rare diseases. Protein function inference is not an easy task, because homologous proteins often have several different functions at the same time. Also such a prediction needs to be fine-tuned with respect to some mutations but robust with respect to others. Researchers have been exploring on this problem with machine learning models, and have also developed a relatively large protein function database \cite{10.1093/nar/gkaa1074} already. However, compared with the number of protein existing in the real world, the specific proteins with known functions included in the database is still very limited. In graph learning, inferring the function of protein molecules based on its structure has also be extensively studied as well. Therefore, in this part, we also include it as a graph reasoning task into {\our} as well. 

Different from the \textit{bibliographic network}, the \textit{protein molecular graphs} have much smaller sizes and there will also exist multiple such graph instances in the dataset. What's more, the features and labels of \textit{protein molecular graphs} are both about the whole molecular graph, not about the individual nodes anymore. As introduced in Section~\ref{subsec:definition}, we can represent the set of studied \textit{protein molecular graphs} as $\mc{G} = \{g_1, g_2, \cdots, g_l\}$, which can be loaded with the following graph loading API call:
\begin{equation}
{\bapi}GL(\text{``}protein\text{-}graph\text{-}set\text{''}) \to \mc{G} {\eapi}.
\end{equation}
For each molecular graph instance $g_i = (\mc{V}_{g_i}, \mc{E}_{g_i})$ in the dataset $\mc{G}$, there will also be raw features and labels related to each protein molecular graph instance. For instance, for the graph instance $g_i \in \mc{G}$, we can represent its raw feature as $\mb{x}_{g_i}$ and its label as $\mb{y}_{g_i}$, where the label vector will indicate its corresponding functions. Based on the protein graph structure and its raw features, we can define the following API call for \textit{protein molecule function reasoning} as follows:
 \begin{equation}
 {{\bapi} GR(\mc{G}, \text{``} seg\text{-}bert\text{:}molecule\text{-}function \text{''}, g_i) \to r {\eapi}},
 \end{equation}
which will call the pre-trained graph neural network SEG-Bert proposed in \cite{Zhang2020SegmentedGF}. The SEG-Bert with full name ``Segmented Graph-Bert'' \cite{Zhang2020SegmentedGF} extends the Graph-Bert model for molecular graph instance representation learning. Besides the SEG-Bert model used in {\our}, the readers can also customize the {\our} framework to include other graph models for addressing the molecular graph reasoning tasks as well.

\vspace{10pt}

\noindent \textbf{Sequential Recommender System Reasoning}: In the era of big data, as more and more data are generated both online and offline, manual search of information from such big data sources has become infeasible nowadays and we may need recommender systems \cite{PremRec} to automatically recommend desired information for us instead. Based on the historical records, sequential recommender system aims to infer the next item(s) that users may be interested in, which may lead to either the future purchase action or the review rating scores of those items. When studying the sequential recommender systems, it is a common way to model recommender systems as the bipartite graphs, where the user-item interaction record also has an attached timestamp. With considerations about the timestamps, sequential recommender systems aim to infer the potential existence (or the weight) of links between user and their interested items for the next future timestamp. In other words, we can define the sequential recommendation problem in recommender systems as a link prediction task with considerations about the temporal factor.

Formally, according to the above description, we can represent the sequential recommender system as a bipartite graph $G = (\mc{V}, \mc{E})$, where the node set $\mc{V} = \mc{U} \cup \mc{I}$ covers both users and items and the links in set $\mc{E} \subset \mc{M} \times \mc{I}$ only exist between users and item instead. For each user-item pair $(u_j, i_l) \in \mc{E}$ in the link set, we can also obtain its timestamp. The sequential recommender system data can be loaded with the following API call:
\begin{equation}
{\bapi}GL(\text{``}recommender\text{-}system\text{''}) \to {G} {\eapi}.
\end{equation}
For each user $u_j$ and item $i_l$ in the recommender system $G$, based on the historical interaction records (before the current timestamp), we can learn the embedding representations of them, which will be used to infer the label between them in the future. Depending on the modeling approach, the label vector can indicate either whether the user will purchase the item or not ({\ie} binary classification task) or the rating score of the user for the item ({\ie} the regression task). Regardless of the specific modeling settings, we can represent the recommender system reasoning API call in LLMs as follows:
 \begin{equation}
 {{\bapi} GR({G}, \text{``}bpr\text{:}recommendation\text{''}, u_j, i_l) \to r {\eapi}},
 \end{equation}
 which will return either the probability scores that the user $u_j$ will be interested in the item $i_l$ or the specific rating scores that $u_j$ will give to $i_l$. We use BPR (Bayesian Personalized Ranking) \cite{10.5555/1795114.1795167} as the default recommendation model in {\our} in this paper, but other recommendation models can also be used for defining the above \textit{recommendation} API calls as well. Besides the recommendation API calls to infer the scores between user and item, for one specific user $u_j$, we can also return the list of top-$k$ recommended items with the following API call:
  \begin{equation}
 {{\bapi} GR({G}, \text{``}bpr\text{:}topk\text{-}recommendation\text{''}, u_j, k) \to r {\eapi}},
 \end{equation}
 where the notation $k$ denotes a hyper-parameter to be extracted from the input statements for the recommendation reasoning.

\vspace{10pt}

\noindent \textbf{Online Social Network Community Reasoning}: Online social networks \cite{10.1145/1298306.1298311}, like Facebook, Twitter and Tiktok, provide different online services for their users to facilitate their online socialization with friends, family members and colleagues. Users in online social networks tend to interact more frequently with their online friends, and they will naturally form their online social communities based on their online social behaviors. Reasoning for the social communities of users in online social networks is a complicated problem. In this part, we will introduce the API calls to empower LLMs to detect social communities from online social networks.

Formally, we can represent the online social network studied in this paper as $G= (\mc{V}, \mc{E})$, where $\mc{V}$ denotes the set of user nodes and $\mc{E}$ denotes the social interactions among the users in the network. The online social network data can be loaded with the following API call:
\begin{equation}
{\bapi}GL(\text{``}social\text{-}network\text{''}) \to {G} {\eapi}.
\end{equation}
Based on $G$, we can represent its detected social community structure as $\mc{C} = \{C_1, C_2, \cdots, C_k\}$, where $\bigcup_{i=1}^k C_i = \mc{V}$. Depending on the problem setting, the social communities to be detected can be based on either hard partition or soft partition of the user node set. For hard partition, we will have $C_i \cap C_j = \emptyset, \forall i, j \in \{1, 2, \cdots, k\}$ ({\ie} there exist no overlap between any two communities); whereas for the soft partition, the communities may have overlaps and one user node may belong to multiple social communities simultaneously. 

Based on the above description, given on the loaded online social network $G$, we can infer both the communities of the whole loaded social network or only the local community for a specific user ({\eg} $u_i \in \mc{V}$) with the following API calls:
\begin{align}
&{{\bapi} GR({G}, \text{``}kmeans\text{:}community\text{''}) \to r {\eapi}}, \\
&{{\bapi} GR({G}, \text{``}kmeans\text{:}community\text{''}, u_i) \to r {\eapi}},
\end{align}
which will call various pre-trained social network community detection algorithms to identify the community structures. In this paper, we will use the KMeans algorithm to partition the user node set into different communities by calculating the number of common neighbors among them as the affinity score. 

\vspace{10pt}

\noindent \textbf{Knowledge Graph Entity and Relation Reasoning}: Compared with unstructured documents, knowledge graph \cite{9416312} aggregates information about entities and their relations from textual data sources in a well-organized representation. Knowledge graph is a powerful tool for supporting a large spectrum of applications in the real-world, like searching, ranking, Q\&A and chatbot dialogue systems. Reasoning of knowledge graphs helps provide the evidences for providing the results with factual basis. At the same time, such a reasoning process will also provide the justification and explanation for the obtained results by the current natural language processing systems. In this paper, we will not study how to build the knowledge graph from textual document sources. Instead, we assume the knowledge graph has been built and is ready to be used for reasoning in the downstream applications. 

Formally, we can represent the built knowledge graph ({\eg} Wikipedia) as $G = (\mc{V}, \mc{E})$, where the node set $\mc{V}$ covers the set of named entities and the link set $\mc{E}$ includes the set of relations among these entities instead. The knowledge graph can be loaded with the following API call:\begin{equation}
{\bapi}GL(\text{``}knowledge\text{-}graph\text{''}) \to {G} {\eapi}.
\end{equation}
Such a loaded knowledge graph structure can be effectively used in the reasoning tasks to infer the potential head-entities, the relations between a pair of entities, and the tail-entities. Formally, we can represent the knowledge graph reasoning API calls used in this paper as:
 \begin{align}
& {\bapi} GR({G}, \text{``}transe\text{:}head\text{-}entity\text{''}, relation, tail\text{-}entity) \to r {\eapi},\\
& {\bapi} GR({G}, \text{``}transe\text{:}relation\text{''}, head\text{-}entity, tail\text{-}entity) \to r {\eapi},\\
& {\bapi} GR({G}, \text{``}transe\text{:}tail\text{-}entity\text{''}, head\text{-}entity, relation) \to r {\eapi}.
 \end{align}
 For the $(\textit{head-entity, relation, tail-entity})$ tuples in the knowledge graph, given any two of them, we can infer the remaining one based on their learned representations. Various pre-trained knowledge graph representation learning models can be used to define the function called in the above API. In this paper, we will use the TransE \cite{NIPS2013_1cecc7a7} as the default knowledge graph embedding and reasoning model.

\vspace{10pt}

\noindent \textbf{A Summary of Advanced Graph Reasoning API Calls}: we also provide a summary of API call examples of the advanced graph reasoning tasks mentioned above in Table~\ref{tab:api_call_summary_lower}. For each of the tasks, we provide several different input reasoning statements, and insert the corresponding reasoning API calls at the most appropriate positions in the output statements. As introduced above, some of the API calls introduced above can be used in different ways to reason for different types of desired information, like based on the online social network \textit{community} reasoning results, we can further define the functions to reason for the \textit{community-count}, \textit{community-size}. Some examples of which have also been provided in Table~\ref{tab:api_call_summary_lower} as well.


\subsection{Prompt Augmentation with ChatGPT}\label{subsec:prompt_augmentation}

For the prompt examples provided in Table~\ref{tab:api_call_summary_upper} and Table~\ref{tab:api_call_summary_lower}, they can only cover a handful number of examples about how to use the API calls for different graph reasoning tasks. Such a small number of instances are not sufficient for the fine-tuning of the existing LLMs. In this paper, we propose to augment the prompt instances with ChatGPT (gpt-3.5-turbo), which has demonstrated excellent few-shot and zero-shot in-context learning ability \cite{Brown2020LanguageMA} in many different language learning tasks already.


\subsubsection{Graph Loading Prompt Dataset Generation}

Similar to \cite{Ouyang2022TrainingLM}, to help the generation of prompt examples, we also provide a detailed instruction for ChatGPT to specify its system role. Here, we can take the graph data loading API call as an example. The instruction together with the prompt examples fed to ChatGPT are provided as follows. Based on both the instruction and prompt examples, we will ask the ChatGPT to generate the graph data loading prompt dataset.\\

{
\sloppy
{
\fontfamily{qcr}\selectfont

\hrule \vspace{.2em} \hrule
\vspace{1em}

\noindent Instruction:  Your task is to add API calls of Graph Loading functions to a piece of input text for concrete graph data loading.\\
The function should help load required graph structured data based on the mentioned graph name and its nodes and links.\\
You can call the Graph Loading API by writing "[GL(graph-name, nodes, links)]", where the "graph-name" denotes the target graph data, and "nodes" and "links" are the mentioned nodes and links.\\
If no specific nodes or links are mentioned, then the API will write "all nodes" and "all links" for the "nodes" and "links" parameters.\\
If only nodes are specified, the API will list the mentioned nodes for the "nodes" parameter entry, and write "all related links" for the "links" parameter entry.\\
If only links are specified, the API will and write "all related nodes" for the "nodes" parameter entry, and list the mentioned links for the "links" parameter entry.\\

\noindent Here are some examples of the API call for loading graph structured data. In the examples, the output will repeat the input, and also insert the API call at the most appropriate position.

\vspace{1em}
\hrule \vspace{.2em} \hrule
\vspace{1em}

\noindent $\bullet$ Input: The structure of the benzene ring molecular graph of benzene ring contains a hexagon.

\noindent $\bullet$ Output: The structure of the [GL("benzene-ring")] molecular graph of benzene ring contains a hexagon.

\noindent \hrulefill

\noindent $\bullet$ Input: There exist a carbon-oxygen double bond in the Acetaldehyde molecular graph.

\noindent $\bullet$ Output: There exist a [GL("acetaldehyde-molecular-graph", \{Carbon, Oxygen\}, \{(Carbon, Oxygen)\})] carbon-oxygen double bond in the Acetaldehyde molecular graph.

\noindent \hrulefill
    
\noindent $\bullet$ Input: The lollipop graph looks like a spoon.

\noindent $\bullet$ Output: The [GL("lollipop-graph", "all nodes", "all links")] lollipop graph looks like a spoon.

\noindent \hrulefill

\noindent $\bullet$ Input: The paper\#10 in the Cora bibliographic network introduces the Transformer model.

\noindent $\bullet$ Output: The [GL("cora", \{Paper\#10\}, "all related citation links")] paper\#10 in the bibliographic network introduces the Transformer model.

\noindent \hrulefill

\noindent $\bullet$ Input: Insulin is a small globular protein containing two long amino acid chains.

\noindent $\bullet$ Output: [GL("insulin-protein-graph", "all atom nodes", "all atom bond links")] Insulin is a small globular protein containing two long amino acid chains.

\noindent \hrulefill

\noindent $\bullet$ Input: At the IMDB recommender system, David rates the "The Avengers" movie with a 10-star review score.

\noindent $\bullet$ Output: At the [GL("imdb-recommender-system", \{"David", "The Avengers"\}, \{("David", "The Avengers")\})] IMDB recommender system, David rates the "The Avengers" movie with a 10-star review score.

\noindent \hrulefill

\noindent $\bullet$ Input: Among the existing online social apps, Tiktok makes it easy for users to socialize with each other online via livestream videos.

\noindent $\bullet$ Output: Among the existing online social apps, [GL("tiktok-social-network", "all user and video nodes", "all user-video links and user-user links")] Tiktok makes it easy for users to socialize with each other online via livestream videos.

\noindent \hrulefill

\noindent $\bullet$ Input: According to the Freebase knowledge graph, Donald Trump was born in 1946 at the Jamaica Hospital Medical Center in New York.

\noindent $\bullet$ Output: According to the [GL("freebase", \{"Donald Trump", "Jamaica Hospital Medical Center", "New York"\}, \{("Donald Trump", "Jamaica Hospital Medical Center"), ("Jamaica Hospital Medical Center", "New York")\})] Freebase knowledge graph, Donald Trump was born in 1946 at the Jamaica Hospital Medical Center in New York.\\

\hrule \vspace{.2em} \hrule
\vspace{1em}

\noindent Query: Based on the instruction and examples, please generate 5000 such input-output pairs for real-world graph data loading. Please make sure the data loaded are in graph structures and the API call is insert ahead of the mentioned graphs or the mentioned nodes or links.

\vspace{1em}
\hrule \vspace{.2em} \hrule
\vspace{1em}
}
}

Based on the instruction, examples and query, by calling ChatGPT API, we obtained a prompt dataset with $5,000$ input-output pair instances. With manual removal the incomplete instances and brief proofreading, about $2,803$ graph data loading API call input-output pairs are preserved in the dataset, which will be used for the fine-tuning to be introduced later.

{\sloppy

\subsubsection{Graph Reasoning Prompt Dataset Generation}

As to the other graph reasoning prompts, with similar instruction and prompt examples, we can use ChatGPT to generate a large number of similar input-output pairs. Meanwhile, slightly different from graph loading API calls, to ensure the graph reasoning prompts are valid, we propose to compose all the inputs statements manually by calling the graph reasoning toolkits in advance. For instance, for the first paper in the Cora bibliographic network, its topic is about ``Neural Networks'' and we will compose its input statement as follows:

{\sloppy
\fontfamily{qcr}\selectfont
\vspace{1em}
\noindent $\bullet$ Input: The first paper in Cora has a topic of Neural Networks.
\vspace{1em}
}

\noindent We will feed such input to ChatGPT and ask it helps insert the graph reasoning API calls to the statement with the query.

{\sloppy
\fontfamily{qcr}\selectfont
\vspace{1em}
\noindent Query: Based on the instruction and examples, generate the output with graph reasoning API calls for the input. Please make sure the API call is insert at the most appropriate position.
\vspace{1em}
}

\noindent Based on the query and input statement, ChatGPT will return the following output:

{\sloppy 
\fontfamily{qcr}\selectfont
\vspace{1em}
\noindent $\bullet$ Output: The first paper in Cora has a topic of [GR(GL("cora", "all paper nodes", "all citation links"), "topic", \{Paper\#1\}) --> r] Neural Networks.
\vspace{1em}
}

\noindent Besides using ChatGPT to annotate the API calls and generate the above output, we also use ChatGPT to rewrite the input statement in another way without changing its semantic meanings. For instance, for the input statement shown above, we also obtain several of its rephrased versions as follows:

{\sloppy
\fontfamily{qcr}\selectfont
\vspace{1em}
\noindent $\bullet$ Input: The initial article in Cora focuses on the subject of Neural Networks.

\vspace{1em}
\noindent $\bullet$ Input: In Cora, the premier paper addresses Neural Networks as its main theme.

\vspace{1em}
\noindent $\bullet$ Input: The foremost paper in the Cora collection pertains to the field of Neural Networks.

\vspace{1em}
\noindent $\bullet$ Input: Cora's inaugural publication delves into the subject matter of Neural Networks.

\vspace{1em}
}

These rephrased input will also be fed to ChatGPT again for the API call annotation as well. Such a process will be done for all the node/graph instances studied in both the basic graph property reasoning tasks and the advanced graph reasoning tasks for generating the input-output prompt pair datasets. Based on the generated dataset, we will run the API calls generated by ChatGPT and compare the return result of the graph reasoning API functions with the true values in the statements. For the outputs whose API calls (1) are not runnable or (2) cannot return the correct result, they will be filtered from the dataset. Finally, after the filtering, the ChatGPT augmented generated datasets will be used for the LLMs fine-tuning, whose statistical information will be provided later in the following experiment section. Meanwhile, for the other graph reasoning tasks not studied in this paper, their reasoning API call datasets can be generated in a similar way as described above.


\subsection{LLMs Fine-Tuning for Graph Reasoning}\label{subsec:llm_finetuning}

Based on the above augmented graph reasoning prompt datasets, in this part, we will introduce how to fine-tune existing pre-trained LLMs, so the LLMs can learn how to use the API tools to address the graph reasoning tasks. Formally, as shown by the prompt examples in Table~\ref{tab:api_call_summary_upper} and Table~\ref{tab:api_call_summary_lower}, given the input statements with a sequence of tokens, {\ie} $\mb{w} = [w_1, w_2, \cdots, w_n]$, the LLMs in {\our} aim to identify the most appropriate position where we can insert the API calls, {\ie} $\mb{s}(c) = {{\bapi} f(args) {\eapi}}$ as introduced in the previous Section~\ref{subsec:api_call}. The major challenges lie in (1) precisely identify the most appropriate positions to insert the API call, (2) correctly the choose the API functions to be used for the call, and (3) also accurately extract the parameters from the context and feed them to the functions. In this section, we will address all these three challenges.


\subsubsection{API Call Insertion Position Prediction}

For the provided input statement, there may exist multiple potential positions for inserting the API calls. Different from \cite{Schick2023ToolformerLM} that choose top-k positions for API call data generation, in this paper, we aim to identify the most likely position to insert the API calls instead. Formally, based on a pre-trained language model $M$, given the input statement $\mb{w} = [w_1, w_2, \cdots, w_n]$, we can insert an API call at the $i_{th}$ (where $i \in \{1, 2, \cdots, n\}$) position ({\ie} right between the tokens $w_{i-1}$ and $w_i$) with the probability
\begin{equation}
P(i | \mb{w}(1:i-1)) = P_M({\bapi} | \mb{w}(1:{i-1})).
\end{equation}
Once the LLM $M$ generates the special beginning token ${\bapi}$ at the $i_{th}$ position, the model will know it should insert an API call here. All the tokens generated after the ${\bapi}$ token and before the special ending token ${\eapi}$ will be the API call function name or the input parameters. For the position index with the latest probability, $\arg\max_{i \in \{1, 2, \cdots, n\}} P(i | \mb{w}(1:i-1))$, it will be selected as the most appropriate position to insert the API calls. 


\subsubsection{API Call Domain and Function Selection}

Different from the very few API call functions studied in \cite{Schick2023ToolformerLM}, the graph reasoning APIs studied in this paper are much more diverse, which may create challenges in the framework implementation and tuning. On the one hand, as more graph reasoning tasks and API functions are incorporated into tuning the LLMs, the graph reasoning API function search space will grow exponentially, which makes it harder to select the correct and the best API functions into the call. On the other hand, some API functions for different graph reasoning tasks may even share similar function names, which may mislead {\our} in selecting the correct ones in the API calls. What's more, different graph reasoning tasks may call different API functions from different toolkits, and some may require different graph functions and trained models to be pre-loaded at the backend. We may also want to specify the trained graph models and graph toolkits to be loaded in the API calls, so {\our} can pre-load the models and toolkits in the generation stage in advance to lower down the overall graph reasoning time costs.

Therefore, according to the graph reasoning API call examples shown before, we propose to slightly change the graph reasoning API call templates introduced in Section~\ref{subsec:api_call} as follows:
\begin{equation}
\mb{s}(c) = {{\bapi} GR(G, domain\text{:}func, args) {\eapi}},
\end{equation}
or
\begin{equation}
\mb{s}(c, r) = {{\bapi} GR(G, domain\text{:}func, args) \to r {\eapi}},
\end{equation}
where the corresponding ``domain'' of the API function is prepend to the specific graph reasoning tasks in the API function call. The domain can be either the used toolkit names or the specific pre-trained model names. 

For instance, for the graph reasoning API calls shown in Table~\ref{tab:api_call_summary_upper} and Table~\ref{tab:api_call_summary_lower}, we will use \textit{toolx} developed in this paper based on networkx toolkit\footnote{https://networkx.org/} for graph property reasoning and can represent the corresponding parameter as ``toolx:property-names''; as to the bibliographic paper topic reasoning with Graph-Bert \cite{Zhang2020GraphBertOA}, we can represent the corresponding parameter as ``\textit{graph-bert:topic}''. For some other cases, if the domain is not specified, we will just use the function in the default domain for the graph reasoning task. More information about the used pre-trained graph models for defining the ``\textit{domain:function}'' entry in the API call will be provided in the following Section~\ref{sec:experiment} for readers. 

In other words, when inserting the API function calls into the statement, we need to infer both the domain and function name of the API call. At the inference stage, as the LLMs generate the domain, the system can pre-load the domain code at the backend even before the whole API call output statement generation is completed. At the same time, it will also allow the LLMs to choose the optimal domain to be used in the API call, since to accomplish the same graph reasoning task, there will exist several different approaches with different performance in terms of effectiveness and efficiency. 

Technically, within the function parameters, we may need to select the best domain from the available candidate domain set, {\ie} $\mc{D} = \{d_1, d_2, \cdots, d_n\}$, according to the prefix context, {\ie} the selected domain after the sequence ``$\mb{w}(1:{i-1}) {\bapi} GR(G, $'' can be represented as
\begin{equation}
d^i = \arg \max_{d_j \in \mc{D}} P_M(d_j | \mb{w}(1:{i-1}) {\bapi} GR(G, ).
\end{equation}
Furthermore, based on the selected domain $d^i$, we may need to select the best function from it that may meet our needs. Formally, we can represent the available functions from the selected domain $d^i$ as set $\mc{F}_{d^i} = \{f_1, f_2, \cdots, f_m\}$, and the function which can maximize the generation probability will be selected, {\ie}
\begin{equation}
f^i = \arg \max_{f_l \in \mc{F}_{j}^i} P_M(f_l | \mb{w}(1:{i-1}) {\bapi} GR(G, d^i\text{:}),
\end{equation}
where the ``$:$'' mark is appended after domain $d^i$ automatically. Once the domain and function are selected, the model may also need to fill in the remaining parameters for the functions accordingly, which will be introduced in the following subsection for readers.


\subsubsection{API Call Function Parameter Completion}

Once the domain and function tokens $d^i\text{:}f^i$ are determined, {\our} will also need to provide the parameters for the selected function based on the statement context. There exist two different ways for completing the parameter entries, {\ie} masked parameter completion and causal parameter completion. 

For the masked parameter completion, once the domain and function are selected, {\our} will automatically generate and insert the remaining tokens, include the function parameter names, the parentheses, the comma and colon marks, and API call ending special token ${\eapi}$. For instance, based on the current token sequence ``$\mb{w}(1:{i-1}) {\bapi} GR(G, d^i\text{:}f^i,$'', {\our} will automatically complete the API call as follows:
\begin{align}
\mb{w}(1:{i-1}) &{\bapi} GR(G, d^i\text{:}f^i, arg_1\text{:}[MASK_1], \\
&arg_2\text{:}[MASK_2], \cdots, arg_n\text{:}[MASK_n] ) {\eapi},
\end{align}
where terms $arg_1$, $arg_2$, $\cdots$, $arg_n$ are the list of parameter names of function $d^i\text{:}f^i$. In the API call, we mask the parameter values, which will be inferred based on both the prefix context and the function parameter names.

The disadvantages of the above masked parameter completion is that the model need to complete the full list of parameters of the function. However, in the real world function calls, only a few number of parameters will be provided actually, whereas the remaining parameters will use their default values instead. Also the masked parameter completion is inconsistent with the previous autoregressive special token and domain/function prediction process. Therefore, in this paper, we propose to use the consistent autoregressive parameter completion with causal language models instead.

For the causal language model based completion of the function parameters, its completion process is similar to the above special token and domain/function selection process. Based on the provided statement and generated tokens, {\our} can generate the list of provided parameter values for the API call, e.g., 
\begin{align}
&arg_j^i = \arg \max P_M(arg | \mb{w}(1\text{:}{i-1}) {\bapi} GR(G, d^i\text{:}f^i, ),\\
&val_j^i = \arg \max P_M(val | \mb{w}(1\text{:}{i-1}) {\bapi} GR(G, d^i\text{:}f^i, arg_j^i\text{:}).
\end{align}
Such a process continues until the end API call special token ``{\eapi}'' is generated. By adding the generated parameter name and value into the token list, we can get the model generation result to be ``$\mb{w}(1\text{:}{i-1}) {\bapi} GR(G, d^i\text{:}f^i, arg_j^i\text{:}val_j^i , \cdots ){\eapi} \mb{w}(i\text{:}n)$'' or ``$\mb{w}(1\text{:}{i-1}) {\bapi} GR(G, d^i\text{:}f^i, arg_j^i\text{:}val_j^i , \cdots ) \to r{\eapi} \mb{w}(i\text{:}n)$'' (with the output tag ``$\to r$''). For the parameters that are not generated by the {\our} model, we will use their default parameter values in the API function calls in the follow-up graph reasoning process.


\subsubsection{LLMs Fine-Tuning with Augmented API Call Dataset}

Formally, given the ChatGPT augmented graph reasoning prompt dataset $\mc{D} = \{(\mb{w}_1, \bar{\mb{w}}_1), (\mb{w}_2, \bar{\mb{w}}_2), \cdots, (\mb{w}_{|\mc{D}|}, \bar{\mb{w}}_{|\mc{D}|})\}$ involving a set of input-output prompt pairs, where the notation $\bar{\mb{w}}_i$ ($\forall i \in \{1, 2, \cdots, |\mc{D}|\}$) is the output statement of $\mb{w}_i$ with inserted API call annotations, according to the above generation process, by feeding the input statement $\mb{w}_i$ to LLMs, we can represent the generated output by the model as
\begin{equation}
\hat{\mb{w}}_i = LLM(\mb{w}_i), \forall i \in \{1, 2, \cdots, |\mc{D}|\}.
\end{equation}

Furthermore, by comparing the generation output $\hat{\mb{w}}_i$ with the ChatGPT annotated output statement $\bar{\mb{w}}_i$, we can define the loss function for fine-tuning the LLM as
\begin{align}
\ell(\mc{D}) &= \frac{1}{|\mc{D}|} \sum_{(\mb{w}_i, \bar{\mb{w}}_i) \in \mc{D}} \ell(\hat{\mb{w}}_i,  \bar{\mb{w}}_i) \\
&=\frac{1}{|\mc{D}|} \sum_{(\mb{w}_i, \bar{\mb{w}}_i) \in \mc{D}} \sum_j \text{cross-entropy}(\hat{\mb{w}}_i(j),  \bar{\mb{w}}_i(j)).
\end{align}


\subsection{Graph Reasoning Q\&A Prompts}\label{subsec:qa_prompt}

What's more, to provide {\our} with basic Q\&A ability for graph reasoning, besides the above statements based graph reasoning prompts, we will also design some Q\&A based prompts for fine-tuning {\our} as well. The graph reasoning Q\&A based prompts are created in a very similar way as above, but we will replace the input statements with other reasoning questions instead, and the output will still be a statement with graph reasoning API calls corresponding to the input question. 

Formally, we also list of the graph reasoning Q\&A prompt examples used in this paper as follows, which will be merged into the previous prompts for fine-tuning {\our}. Different from the previous input-output statement prompts (where the output is almost a duplicated copy of the input but with API calls), the inputs and outputs in question-answer prompts are not duplicated copies of each other anymore. However, with the above autoregressive generation of the desired output statement introduced before, the {\our} using causal language models as the backbone is still capable to generate the desired output statements for the input question queries.

\vspace{.5em}

\noindent \textbf{1. Graph Property Reasoning Q\&A Prompt Examples:}

\vspace{1em}

{\sloppy
\fontfamily{qcr}\selectfont

\hrule \vspace{.2em} \hrule

\vspace{.5em}
\noindent $\bullet$ Input: What is the order of the barbell graph?

\noindent $\bullet$ Output: The order of the barbell graph is [GR(GL("gpr", {"barbell\_graph"}), "toolx:order")-->r].

\noindent \hrulefill

\vspace{.5em}

\noindent $\bullet$ Input: What is the size of the star graph?

\noindent $\bullet$ Output: The size of the star graph is [GR(GL("gpr", {"star\_graph"}), "toolx:size")-->r].

\noindent \hrulefill

\vspace{.5em}

\noindent $\bullet$ Input: What is the density of the dodecahedral graph?

\noindent $\bullet$ Output: The density of dodecahedral graph is [GR(GL("gpr", {"dodecahedral\_graph"}), "toolx:density")-->r].

\noindent \hrulefill

\vspace{.5em}

\noindent $\bullet$ Input: What is the eccentricity of node \#25 in the balanced tree?

\noindent $\bullet$ Output: The eccentricity of node \#25 in the balanced tree is [GR(GL("gpr", {"balanced\_tree"}), "toolx:eccentricity", "node\#25")-->r].

\noindent \hrulefill

\vspace{.5em}

\noindent $\bullet$ Input: What is the radius of the lollipop graph?

\noindent $\bullet$ Output: The radius of the lollipop graph is [GR(GL("gpr", {"lollipop\_graph"}), "toolx:radius")-->r].

\noindent \hrulefill

\vspace{.5em}

\noindent $\bullet$ Input: What is the center of the star graph?

\noindent $\bullet$ Output: The center of the star graph includes node(s) [GR(GL("gpr", {"star\_graph"}), "toolx:center")-->r].

\noindent \hrulefill

\vspace{.5em}

\noindent $\bullet$ Input: What is the length of shortest path between node \#5 and node \#0 in the octahedral graph?

\noindent $\bullet$ Output: In the octahedral graph, the length of shortest path between node \#5 and node \#0 is [GR(GL("gpr", {"octahedral\_graph"}), "toolx:shortest\_path", "node\#5", "node\#0")-->r].

\noindent \hrulefill

\vspace{.5em}

\noindent $\bullet$ Input: What is the diameter of the binomial tree?

\noindent $\bullet$ Output: The diameter of the binomial tree is [GR(GL("gpr", {"binomial\_tree"}), "toolx:diameter")-->r].

\noindent \hrulefill

\vspace{.5em}

\noindent $\bullet$ Input: What is the periphery of the house x graph?

\noindent $\bullet$ Output: The periphery of the house x graph includes node(s) [GR(GL("gpr", {"house\_x\_graph"}), "toolx:periphery")-->r].

\vspace{1em}

\hrule \vspace{.2em} \hrule

\vspace{1em}
}

\noindent \textbf{2. Bibliographic Network Reasoning Q\&A Prompt Examples:}

\vspace{1em}

{\sloppy
\fontfamily{qcr}\selectfont

\hrule \vspace{.2em} \hrule

\vspace{.5em}
\noindent $\bullet$ Input: What is the topic of paper \#83826 in the cora bibliographic network?

\noindent $\bullet$ Output: The topic of paper \#83826 in the cora bibliographic network is [GR(GL("cora"), "graph\_bert:topic", paper\#83826)-->r].

\noindent \hrulefill

\vspace{.5em}
\noindent $\bullet$ Input: What is the topic of paper \#5832 in the pubmed bibliographic network?

\noindent $\bullet$ Output: The topic of paper \#5832 in the pubmed bibliographic network is [GR(GL("pubmed"), "graph\_bert:topic", paper\#5832)-->r].

\noindent \hrulefill

\vspace{.5em}
\noindent $\bullet$ Input: What is the topic of paper \#3230 in the citeseer bibliographic network?

\noindent $\bullet$ Output: The topic of paper \#3230 in the citeseer bibliographic network is [GR(GL("citeseer"), "graph\_bert:topic", paper\#3230)-->r].

\vspace{1em}

\hrule \vspace{.2em} \hrule

\vspace{1em}

}

\noindent \textbf{3. Molecular Graph Reasoning Q\&A Prompt Examples:}

\vspace{1em}

{\sloppy
\fontfamily{qcr}\selectfont

\hrule \vspace{.2em} \hrule

\vspace{.5em}
\noindent $\bullet$ Input: What is the function for the protein molecular graph \#138 in proteins?

\noindent $\bullet$ Output: The function for the protein molecular graph \#138 in proteins is [GR(GL("proteins"), "seg\_bert:molecule\_function", instance\#138)-->r].

\noindent \hrulefill

\vspace{.5em}
\noindent $\bullet$ Input: What is the function for the chemical molecular graph \#129 in mutag?

\noindent $\bullet$ Output: The function for the chemical molecular graph \#129 in mutag is [GR(GL("mutag"), "seg\_bert:molecule\_function", instance\#129)-->r].

\noindent \hrulefill

\vspace{.5em}
\noindent $\bullet$ Input: What is the function for the chemical molecular graph \#322 in nci1?

\noindent $\bullet$ Output: The function for the chemical molecular graph \#322 in nci1 is [GR(GL("nci1"), "seg\_bert:molecule\_function", instance\#322)-->r].

\noindent \hrulefill

\vspace{.5em}
\noindent $\bullet$ Input: What is the function for the chemical molecular graph \#44 in ptc?

\noindent $\bullet$ Output: The function for the chemical molecular graph \#44 in ptc is [GR(GL("ptc"), "seg\_bert:molecule\_function", instance\#44)-->r].

\vspace{1em}

\hrule \vspace{.2em} \hrule

\vspace{1em}
}

\noindent \textbf{4. Social Network Reasoning Q\&A Prompt Examples:}

\vspace{1em}

{\sloppy
\fontfamily{qcr}\selectfont

\hrule \vspace{.2em} \hrule

\vspace{.5em}
\noindent $\bullet$ Input: In foursquare, what is the id of user sparkey215's community?

\noindent $\bullet$ Output: In foursquare, the id of user sparkey215's community is [GR(GL("foursquare"), "kmeans:community", user\#sparkey215)-->r].

\noindent \hrulefill

\vspace{.5em}
\noindent $\bullet$ Input: In the online social network foursquare, are user \#user/9674821 and user \#ljaniszewski8 belong to the same community?

\noindent $\bullet$ Output: In the online social network foursquare, user \#user/9674821 and user \#ljaniszewski8 belong to [GR(GL("foursquare"), "kmeans:common\_community\_check", user\#user/9674821, user\#ljaniszewski8)-->r] community.

\vspace{1em}

\hrule \vspace{.2em} \hrule

\vspace{1em}

}

\noindent \textbf{5. Recommender System Reasoning Q\&A Prompt Examples:}

\vspace{1em}

{\sloppy
\fontfamily{qcr}\selectfont

\hrule \vspace{.2em} \hrule

\vspace{.5em}
\noindent $\bullet$ Input: How likely user \#A23E9QQHJLNGUI will be interested in item \#B004PIPG2A in Amazon?

\noindent $\bullet$ Output: The likelihood that user \#A23E9QQHJLNGUI will be interested in item \#B004PIPG2A in Amazon is [GR(GL("amazon"), "bpr:recommendation", user\#A23E9QQHJLNGUI, item\#B004PIPG2A)-->r].

\noindent \hrulefill

\vspace{.5em}
\noindent $\bullet$ Input: How likely user \#u329 will be interested in music of artist \#i8323 in Last-fm?

\noindent $\bullet$ Output: The likelihood that user \#u329 will be interested in music from artist \#i8323 in Last-fm is [GR(GL("last-fm"), "bpr:recommendation", user\#u329, artist\#i8323)-->r].

\noindent \hrulefill

\vspace{.5em}
\noindent $\bullet$ Input: How likely user \#u650 will be interested in movie \#i671 in Movielens?

\noindent $\bullet$ Output: The likelihood that user \#u650 will be interested in movie \#i671 in Movielens is [GR(GL("movielens"), "bpr:recommendation", user\#u650, movie\#i671)-->r].

\vspace{1em}

\hrule \vspace{.2em} \hrule

\vspace{1em}

}

\noindent \textbf{6. Knowledge Graph Reasoning Q\&A Prompt Examples:}

\vspace{1em}

{\sloppy
\fontfamily{qcr}\selectfont

\hrule \vspace{.2em} \hrule

\vspace{.5em}
\noindent $\bullet$ Input: According to the Freebase knowledge graph, what is the relation between entity\#/m/053yx and entity\#/m/015\_1q?

\noindent $\bullet$ Output: According to the Freebase knowledge graph, the relation between entity\#/m/053yx and entity\#/m/015\_1q is [GR(GL("freebase"), "transe:relation", entity\#/m/053yx, entity\#/m/015\_1q)-->r].

\noindent \hrulefill

\vspace{.5em}
\noindent $\bullet$ Input: According to the WordNet knowledge graph, via relation \#\_hypernym, we derive entity \#imagination.n.02 from what entity?

\noindent $\bullet$ Output: According to the WordNet knowledge graph, via relation \#\_hypernym, we can obtain entity \#imagination.n.02 from entity [GR(GL("wordnet"), "transe:head\_entity", relation\#\_hypernym, entity\#imagination.n.02)-->r].

\vspace{1em}

\hrule \vspace{.2em} \hrule

\vspace{1em}

}


\begin{figure*}
    \centering
    \begin{minipage}{.95\textwidth}
    	\includegraphics[width=0.95\linewidth]{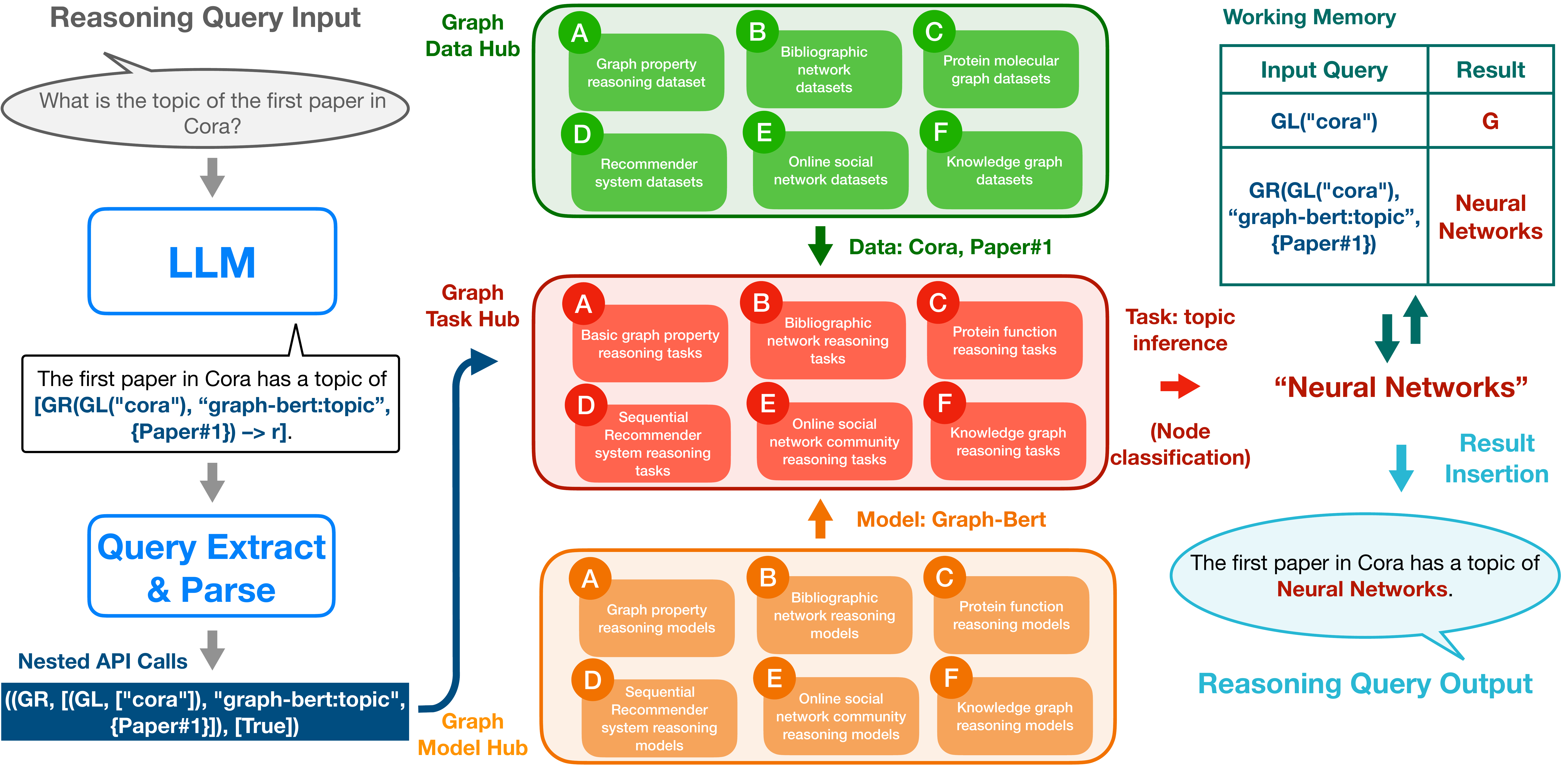}
        \caption{An Illustration of Graph Reasoning Query Processing. The graph reasoning query processing component in {\our} has several modules/hubs: (1) LLM based query statement generation, (2) query extraction and parsing model, (3) query execution model, (4) working memory module, and (5) output post-processing module. The graph reasoning query execution module is built based on the (6) graph reasoning task hub, (7) graph data hub, and (8) graph model hub. The {\our} framework will recognize the parsed graph reasoning queries, load the corresponding graph data, call the corresponding graph reasoning model, execute the reasoning task API function to generate the result, store the result into working memory and insert the output result to replace the reasoning query in the LLM generated reasoning response statement as well.}
    	\label{fig:inference_illustration}
	\end{minipage}
\end{figure*}

\subsection{LLMs Inference and Graph Reasoning Query Parsing, Execution and Post-Processing}\label{subsec:llm_inference_postprocessing}

Finally, at the end of this section, we will introduce the details about how to use the fine-tuned LLMs in {\our} for addressing various graph reasoning tasks. As shown in Figure~\ref{fig:inference_illustration}, the graph reasoning process has several important steps based on several functional modules, which include (1) \textit{LLMs based output query statement generation}, (2) \textit{query extraction and parsing}, (3) \textit{query execution}, (4) \textit{graph data hub}, (5) \textit{graph model hub}, (6) \textit{graph task hub}, (7) \textit{working memory} and (8) \textit{reasoning output post-processing}. In this subsection, we will introduce these steps and the involved functional modules/hubs used in {\our} for readers.

\subsubsection{LLMs Inference}\label{subsubsec:llm_inference}

Based on the prompt datasets, we have discussed about how to fine-tune the LLMs in {\our} in the previous subsections already, which is capable to generate the graph reasoning query statement outputs for the input statements. To apply the the fine-tuned LLMs for the inference, given any graph reasoning input statement, the LLMs will project the input statement to the corresponding output statement annotated with the API calls. We also provide an example about the inference process as follows:
 
{\sloppy
\fontfamily{qcr}\selectfont
\vspace{1em}
\noindent $\bullet$ Input: The first paper in Cora has a topic of [TBR].

\vspace{1em}
\noindent $\bullet$ Output: The first paper in Cora has a topic of [GR(GL("cora"), "graph\text{-}bert:topic", \{Paper\#1\}) --> r].
\vspace{1em}
}

Meanwhile, by including the Q\&A based prompt datasets for LLMs fine-tuning, the {\our} will also be capable to generate the graph reasoning statements for the input question queries as well, such as

{\sloppy
\fontfamily{qcr}\selectfont
\vspace{1em}
\noindent $\bullet$ Input: What is the topic of the first paper in Cora bibliographic network?

\vspace{1em}
\noindent $\bullet$ Output: The first paper in Cora has a topic of [GR(GL("cora"), "graph\text{-}bert:topic", \{Paper\#1\}) --> r].
\vspace{1em}
}

The LLMs in {\our} can add the correct graph reasoning API calls into the output at the correct position for majority of the graph reasoning input statements and questions (we will illustrate the experimental results in the following Section~\ref{sec:experiment}). Such generated output statements with API calls will be fed to the following parser module to extract the graph reasoning queries.

\subsubsection{Query Parser Module}

Since we allow both nested and sequential API calls in {\our}, the parsing of the LLMs' generated output graph reasoning queries is never an easy task. Here, we can take the output query ``{\fontfamily{qcr}\selectfont[GR(GL("cora"),"graph\text{-}bert:topic",\{Paper\#1\})-->r]}'' generated by the LLMs introduced in the above subsection as an example. The {\our} framework introduce a query parser module that is capable to identify and parse the queries to a standard format that is recognizable and executable by the executor module. The expected parsing result of the query will involve two parts:
\begin{itemize}

\item \textbf{Function Call Parsing}: To differentiate normal textual tokens in the statement from the graph reasoning API call queries, we will use the regular expression to identify and parse the queries. The function call part of the query can be effectively identified with the following regular expression in python:
\begin{lstlisting}[language=Python]
function_call_pattern = r'\b([a-zA-Z_][a-zA-Z0-9_]*)\s*\(([^\]]*)\)'
\end{lstlisting}
which will detect both the API GR/GL function tokens, as well as the parameters in the API call. For the parameters in the API call, if we identify there exist any nested API calls (via detecting the parentheses marks $($ and $)$), we will recursively parse the nested API call.

\item \textbf{Output Insertion Parsing}: The extraction of the output insertion tag ``{\fontfamily{qcr}\selectfont-->r}'' will be much easier, which can be identified with the regular expression as well, {\ie} 
\begin{lstlisting}[language=Python]
output_variable_pattern = r'\)\s*([-->]*)([a-zA-Z0-9_\s]*)\s*\]'
\end{lstlisting}
For the API calls studied in this paper, if the tag ``{\fontfamily{qcr}\selectfont-->r}'' exists, we will replace the query text to insert the graph reasoning results into the original text; otherwise, the query will be executed in the backend only, whose result will be recorded in the working memory to be introduced later.

\end{itemize}

For instance, for the example query generated by the LLM ``{\fontfamily{qcr}\selectfont[GR(GL("cora"),"graph\text{-}bert:topic",\{Paper\#1\})-->r]}'', its nested parsing result by the query parser module in the {\our} framework can be represented as ``{\fontfamily{qcr}\selectfont ((GR, [(GL, ["cora"]), "graph\text{-}bert:topic", \{Paper\#1\}]), [True])}'', where the ``{\fontfamily{qcr}\selectfont True}'' tag denotes the existence of the output insertion tag ``{\fontfamily{qcr}\selectfont-->r}'', {\ie} we need to replace the query text with the reasoning result into the output statement.

\subsubsection{Graph Reasoning Hubs}\label{subsubsec:graph_reasoning_hub}

Prior to the reasoning stage to execute the parsed queries, all the graph loading and reasoning API toolkits and models will be pre-loaded and ready-to-use. Also all the graph data to be loaded will be organized into a unified format that the graph loading API functions can handle. Specifically, we introduce several hubs in the {\our} framework, that will host the various \textit{graph datasets}, \textit{pre-trained graph models} and \textit{graph reasoning tasks}, respectively.

\begin{itemize}

\item \textbf{Graph Dataset Hub}: A set of pre-processed graph datasets to be used in the {\our} framework will be organized into the \textit{graph dataset hub}. All these datasets will have a unified format, and it will allow both the {\our} framework and the pre-trained graph models to access the desired information in the reasoning process. In the Appendix, we will describe the standard data organization format used by the source code of {\our} in the experiment. Specifically, the datasets hosted in the \textit{graph dataset hub} include
\begin{itemize}
\item \textit{Graph Property Reasoning Dataset}: GPR;
\item \textit{Bibliographic Network Datasets}: Cora, Pubmed, Citeseer;
\item \textit{Molecular Graph Datasets}: Proteins, Mutag, Nci1, Ptc;
\item \textit{Online Social Network Datasets}: Twitter, Foursquare;
\item \textit{Recommender System Datasets}: Amazon, Last-FM, Movielens;
\item \textit{Knowledge Graph Datasets}: WordNet, Freebase.
\end{itemize}
More detailed information about the graph datasets studied in this paper will be introduced in the following Section~\ref{sec:experiment} when talking about the experiments.

\item \textbf{Graph Model Hub}: In the {\our} framework, we also define a \textit{graph model hub} for hosting several (ready-to-use) graph tools and pre-trained graph neural network models. Specifically, the graph models included in the {\our} framework include
\begin{itemize}
\item \textit{Toolx}: created in this paper based on networkx for property calculation;
\item \textit{Graph-Bert} \cite{Zhang2020GraphBertOA}: built for graph representation and node classification;
\item \textit{SEG-Bert} \cite{Zhang2020SegmentedGF}: built for graph representation and graph instance classification;
\item \textit{KMeans} \cite{MacQueen1967}: built for graph partitioning and node clustering;
\item \textit{BPR} \cite{10.5555/1795114.1795167}: built for link ranking and recommendation;
\item \textit{TransE} \cite{NIPS2013_1cecc7a7}: built for graph entity/relation searching.
\end{itemize} 
More detailed information about these models will be introduced in the following Section~\ref{sec:experiment}. These graph models will implement the basic graph reasoning functions, which will be called in the specific graph reasoning tasks on the provided graph datasets.

\item \textbf{Graph Task Hub}: Finally, the \textit{graph task hub} will define the specific graph reasoning tasks to be studied in the {\our} framework. As introduced in the previous Section~\ref{subsec:formulation}, most of the graph reasoning tasks can be reduced to several very fumdamental graph learning tasks, {\eg} (1) \textit{graph attribute calculation}, (2) \textit{node classification}, (3) \textit{graph classification}, (4) \textit{graph partition/clustering}, (5) \textit{link prediction/ranking} and (6) \textit{graph searching} tasks. For all the application oriented graph reasoning tasks as introduced in Section~\ref{subsec:graph_reasoning_api_call}, {\ie} (1) \textit{graph property reasoning}, (2) \textit{bibliographic paper topic reasoning}, (3) \textit{molecular graph function reasoning}, (4) \textit{social network community reasoning}, (5) \textit{recommender system reasoning} and (6) \textit{knowledge graph reasoning}, we will reduce them to the very fumdamental graph learning tasks in the \textit{graph task hub}.

\end{itemize}

Besides the hubs we mention above, within the {\our} framework, we also have an extra hub for hosting the LLMs to be used for the graph reasoning API  generation based on the inputs received from the interaction with the end users. The LLM hub will host a set of fine-tuned language models for the graph reasoning tasks. Specifically, within the {\our} framework studied in this paper, several LLMs (like \textit{GPT-J 6B 8bit} can be included in the hub for the output graph reasoning statement generation.


\subsubsection{Query Executor Module}

Furthermore, the generation output will be further post-processed by detecting and initiating the API calls in it. Depending on whether the API call return result needs to be outputted or not, the executor {\our} will also further replace the API calls with its return result in the statement. For instance, for the example mentioned in Section~\ref{subsubsec:llm_inference}, based on the parsing result ``{\fontfamily{qcr}\selectfont ((GR, [(GL, ["cora"]), "graph\text{-}bert:topic", \{Paper\#1\}]), [True])}'', {\our} will recognize and execute the query as follows:
\begin{itemize}

\item \textbf{Outer Function}: ``{\fontfamily{qcr}\selectfont GR}'', {\ie} the outer function is a for graph reasoning.

\item \textbf{Outer Function Parameters}: ``{\fontfamily{qcr}\selectfont [(GL, ["cora"]), "graph\text{-}bert:topic", \{Paper\#1\}] }''.

\begin{itemize}

\item \textbf{Parameter 1}:  ``{\fontfamily{qcr}\selectfont (GL, ["cora"])}'', the first parameter is a nested API call.

\begin{itemize}

\item \textbf{Inner Function}: ``{\fontfamily{qcr}\selectfont GL}'', {\ie} the inner function is a for graph loading.

\item \textbf{Inner Function Parameter(s)}: ``{\fontfamily{qcr}\selectfont ["cora"]}'', {\ie} the graph loading API will load the Cora network dataset.

\end{itemize}

\item \textbf{Parameter 2}:  ``{\fontfamily{qcr}\selectfont "graph\text{-}bert:topic"}'', the second parameter denotes the outer reasoning function aims to infer the topic with the Graph-Bert model.

\item \textbf{Parameter 3}:  ``{\fontfamily{qcr}\selectfont \{Paper\#1\}}'', the third parameter denotes the outer reasoning function focuses on the ``Paper\#1'' in the input graph dataset.

\end{itemize}

\item \textbf{Output Insertion Tag}: ``{\fontfamily{qcr}\selectfont True}'', {\ie} this query requires the replacement and insertion of this query result back into the statement. 

\end{itemize}

After parsing the query in the text, the query executor module in {\our} will execute the query by calling the corresponding API function with the provided graph data and parameters, {\ie} ``Graph-Bert.topic(cora, \{Paper\#1\})'', which will return the graph reasoning query result, {\ie} {\fontfamily{qcr}\selectfont Neural Networks}, as the output bibliographic paper topic reasoning query. Furthermore, since the output insertion token ``{\fontfamily{qcr}\selectfont -->r}'' exist in the query, the query executor module in {\our} will also replace the query token sequence with the reasoning results, which will generate the final output by the {\our} as follows:
 
{\sloppy
\fontfamily{qcr}\selectfont
\vspace{1em}
\noindent $\bullet$ Input: The first paper in Cora has a topic of [TBR].

\vspace{1em}
\noindent $\bullet$ Post-processed Output: The first paper in Cora has a topic of Neural Networks.
\vspace{1em}

}

\subsubsection{Working Memory Module}

What's more, within the {\our} model, we also maintain a small-sized working memory, which keeps records of both the recent external API function calls (including both GL and GR API function calls) and their output results for the model in the reference stage. For instance, if the API calls on the graph loading $GL(file\text{-}path, node\text{-}subset, link\text{-}subset)$ or on the graph reasoning $GR(G, d^i\text{:}f^i, arg_1\text{:}{val}_1, arg_2\text{:}{val}_2, \cdots, arg_n\text{:}{val}_n )$ has been executed, then its read-only result will be stored into the working memory. In the future, if we have similar queries on the same graph dataset again, the stored results can be retrieved from the working memory directly as the output. Such a working memory is very helpful, especially for the API function calls like data loading or other graph reasoning API calls with large time or space costs. For instance, as shown in Table~\ref{tab:api_call_summary_upper}, after the example lollipop graph (the top left green graph) shown in Figure~\ref{fig:problem_illustration} has been loaded as $G_l$, we will just replace the data loading file name path with the graph $G_l$ directly. The reuse of pre-stored result from the working memory will save lots of time costs on graph reasoning tasks. The working memory has a pre-defined memory capacity in {\our}, and will maintain its stored information similar to a queue ({\ie} FIFO). Once the stored information exceeds the working memory capacity, the result of the oldest API calls or the results which has been rewritten already will be removed from the working memory by {\our}.

}

{\sloppy
\section{Experiments}\label{sec:experiment}

In this section, we will conduct extensive experiments to evaluate the performance of {\our} on various graph reasoning tasks that we have discussed before. According to the previous method section, we will use ChatGPT to generate a large-size of graph reasoning prompt dataset based on both the textual instructions and a small number of hand-crafted prompt reasoning examples. To ensure {\our} can handle diverse graph reasoning tasks, we will merge the generated prompt datasets for different graph reasoning tasks on different graph datasets together to obtain a mixed prompt dataset. By partitioning the mixed prompt dataset into training and testing sets, we will fine-tune existing pre-trained LLMs ({\eg} GPT-J or LLaMA) on the training set, and evaluate its generation performance on the testing set. What's more, the fine-tuned LLMs will be further plugged into {\our} for conducting graph reasoning based on the textual input statement or question queries. More details about the experimental settings and some experimental results will be provided in the following parts of this section. All the source code, datasets, and checkpoints of all the pre-trained graph models and fine-tuned LLMs have been released and shared to the community, which can be accessed via the github link provided at the beginning of this paper.

\begin{table*}[h]
\caption{A statistical summary of graph datasets used in the experiments of this paper. For the GPR and molecular graph datasets (including PROTEIN, PTC, NCI1 and MUTAG), the ``Node\#'' and ``Edge\#'' denote the average numbers of nodes and edges for the graph instances in the datasets, respectively. For the graphs without features or labels, we will fill the entries with ``NA'' in the table.}\label{tab:data_statistics}
\centering
\setlength{\tabcolsep}{.3em}
\renewcommand{\arraystretch}{1.6}
\begin{tabular}{lcccccccc}
\toprule
\textbf{Tasks} &\textbf{Datasets} & \textbf{Graph Types} & \textbf{Node\#} & \textbf{Edge\#} & \textbf{Graph\#} & \textbf{Feature\#} & \textbf{Class\#} &\textbf{Prompt\#}\\
\hline
\textbf{Graph Loading} &GL-Prompt &NA &NA &NA &NA &NA &NA &2,802\\
\hline
\multirow{2}{*}{\parbox{2.5cm}{\textbf{Property\\ Reasoning}}}
&\multirow{2}{*}{GPR-Prompt} & \multirow{2}{*}{Generated classic graphs} & \multirow{2}{*}{14.70 (avg)} & \multirow{2}{*}{28.27 (avg)} & \multirow{2}{*}{37} & \multirow{2}{*}{NA} & \multirow{2}{*}{NA} & \multirow{2}{*}{2,587} \\
&&&&&&&&\\
\hline
\multirow{3}{*}{\parbox{2.5cm}{\textbf{Paper\\ Topic\\ Reasoning}}}
&Cora & Bibliographic network & 2,708 & 5,429 & 1 & 1,433 & 7 &18,956 \\
\cline{2-9}
&Citeseer & Bibliographic network & 3,327 & 4,732 & 1 & 3,703 & 6 &23,184 \\
\cline{2-9}
&Pubmed & Bibliographic network & 19,717 & 44,338 & 1 & 500 & 3 &138,019 \\
\hline
\multirow{4}{*}{\parbox{2.5cm}{\textbf{Molecule\\ Function\\ Reasoning}}}
&PROTEINS & Protein molecular graphs & 39.05 (avg) & 72.82 (avg) & 1,113 & NA &2  &6,678 \\
\cline{2-9}
&PTC & Chemical molecular graphs & 25.56 (avg) & 25.96 (avg) & 344 & NA &2  &2,064 \\
\cline{2-9}
&NCI1 & Chemical molecular graphs & 29.86 (avg) & 32.30 (avg) & 4,110 & NA &2  &24,660 \\
\cline{2-9}
&MUTAG & Chemical molecular graphs & 17.93 (avg) & 19.79 (avg) & 188 & NA &2  &1,128 \\
\hline
\multirow{3}{*}{\parbox{2.5cm}{\textbf{Sequential\\ Recommendation\\ Reasoning}}}
&MovieLens & Recommender system & 2,625 & 100,000 & 1 & NA & 5 (rating) &500,000 \\
\cline{2-9}
&Last.FM & Recommender system & 19,524 & 118,268 & 1 & NA & 2 (binary) &355,320 \\
\cline{2-9}
&Amazon & Recommender system & 396,810 & 450,578 & 1 & NA & 5 (rating) &2,252,890 \\
\hline
\multirow{2}{*}{\parbox{2.5cm}{\textbf{Social\\ Community\\ Reasoning}}}
&Foursquare & Social network & 5,392 & 76,972 & 1 & NA & NA &64,710 \\
\cline{2-9}
&Twitter & Social network & 5,223 & 164,920 & 1 & NA & NA &52,240 \\
\hline
\multirow{2}{*}{\parbox{2.5cm}{\textbf{Knowledge\\ Graph\\ Reasoning}}}
&Freebase & Knowledge graph & 14,951 & 592,213 & 1 & NA & NA &1,695,651 \\
\cline{2-9}
&WordNet & Knowledge graph & 41,105 & 151,442 & 1 & NA & NA &454,326 \\
\bottomrule
\end{tabular}
\end{table*}



\subsection{Graph Benchmark Dataset Descriptions}

As introduced in the previous Section~\ref{subsubsec:graph_reasoning_hub}, about $15$ different graph benchmark datasets are studied in this paper, which include
{
\begin{itemize}
\item \textbf{Graph Property Reasoning Dataset}: We create a toy dataset named ``GPR'' in this paper containing $37$ special connected graph instances generated by networkx toolkit, which include the ``\textit{bull graph}'', ``\textit{wheel graph}'', ``\textit{lollipop graph}'', etc. These generated graph instances all have a relatively small size with about $15$ nodes and and $28$ links on average.

\item \textbf{Bibliographic Network Datasets}: We use three benchmark bibliographic network datasets in the experiment for to infer the paper topics with {\our}, which include Cora, Pubmed, Citeseer \cite{Kipf_Semi_CORR_16, Zhang2020GraphBertOA}. Each node in these bibliographic networks denotes an academic paper, which are annotated with both numerical features and categorical labels indicating the paper topics.

\item \textbf{Molecular Graph Datasets}: We use four molecular graph benchmark datasets in the experiments, which include PROTEINS, MUTAG, NCI1, PTC \cite{10.1145/2783258.2783417, Zhang2020SegmentedGF}. For the molecular graphs in these four datasets, we can obtain their topological structures and categorical label indicating the molecular graph functions.

\item \textbf{Social Network Datasets}: Two online social network benchmark datasets Twitter, Foursquare \cite{10.1145/2505515.2505531} are investigated in this experiment. We can obtain both the social connections among the users and other diverse heterogeneous information. In the experiment, we will only use the social connections among the users to reason for the social communities from the social networks.

\item \textbf{Recommender System Datasets}: Three sequential recommender system datasets Amazon (Software) \cite{McAuley2015ImageBasedRO}, Last-FM \cite{Bertin-Mahieux2011}, Movielens (100K) \cite{10.1145/2827872} are studied in this paper. For each recommender system, we have the user-item interaction records annotated with the timestamps. Existing sequential recommender systems will partition the datasets into training/testing sets by the timestamps: the historical records will be used for model training and the last interaction is used for testing. We will follow the same settings in the experiment.

\item \textbf{Knowledge Graph Datasets}: We also obtain and use two knowledge graph benchmark datasets WordNet \cite{10.1145/219717.219748}, Freebase \cite{10.1145/2567948.2577016} in the experiments. These datasets will be partitioned into training/testing sets for knowledge graph embedding and model training.

\end{itemize}
}
To make it easier for the graph data hub to load the graph datasets for various reasoning tasks, we will pre-process the datasets and organize the graph information into a unified format, which has been described in detail in the Appendix. We also provide the dataset statistical information in Table~\ref{tab:data_statistics}, which include both the statistics of these graph datasets (including numbers of nodes, links, graph instances, features and classes) and the obtained prompt dataset size. The raw graph datasets and the generated graph reasoning prompt datasets have been shared with the community and released at the github page\footnote{data github link: https://github.com/jwzhanggy/Graph\_Toolformer/tree/main/data}.


\subsection{Pre-Trained Graph Models}

As introduced in the previous Section~\ref{subsubsec:graph_reasoning_hub}, {\our} also includes a set of pre-trained graph models in the framework for various graph reasoning tasks. According to the previous Section~\ref{subsec:formulation}, since most of the application oriented graph reasoning tasks can be reduced to the very fundamental graph learning tasks, like \textit{attribute calculation}, \textit{node classification}, \textit{graph classification}, \textit{link prediction}, \textit{graph partition/clustering} and \textit{graph searching}, these pre-trained graph models in {\our} will implement these fundamental graph learning functions correspondingly.

\begin{itemize}
\item \textbf{Toolx}: The current toolx model in {\our} is implemented based on networkx, and toolx will implement different API functions to calculate different graph properties mentioned in the paper, including \textit{order}, \textit{size}, \textit{density}, \textit{eccentricity}, \textit{radius}, \textit{diameter}, \textit{center}, \textit{shortest-path}, \textit{avg-path-length}, \textit{min-path-length}, \textit{max-path-length}, \textit{periphery}.

\item \textbf{Graph-Bert}: The Graph-Bert model proposed in \cite{Zhang2020GraphBertOA} can effectively learn the representations of graph data. The Graph-Bert model will be used to implement the bibliographic network paper topic inference function in {\our}, which will implement the corresponding API function of \textit{node-classification}.

\item \textbf{SEG-Bert}: The SEG-Bert model original proposed in \cite{Zhang2020SegmentedGF} can both embed and classify graph instances. It will be used to implement the molecular graph function inference function and implement the corresponding \textit{graph-classification} API function.

\item \textbf{KMeans}: In {\our}, we extend the KMeans algorithm \cite{MacQueen1967} to partition the social network data for detecting the social communities. Specifically, we calculate the \textit{common neighbor} to define the affinity matrix among users in social networks, and define the \textit{community-detection} function with KMeans to partition the nodes into clusters.

\item \textbf{BPR}: The BPR (Bayesian Personalized Ranking) proposed in \cite{10.5555/1795114.1795167} will compare and rank the personalized positive and negative user-item pairs for model learning. In {\our}, we use BPR to define the API functions about recommender systems, which include \textit{recommendation} (to calculate the scores for provided user-item pair) and \textit{top k-recommendation} (to recommend top-k items for provided user).

\item \textbf{TransE}: The The TransE model proposed in \cite{NIPS2013_1cecc7a7} will be used to implement the knowledge graph entity/relation reasoning functions in {\our}, will be reduced to several graph searching functions like \textit{search-head-entity}, \textit{search-tail-entity} and \textit{search-relation}.
\end{itemize} 

These graph models will be pre-trained with the datasets introduced above by following the identical train/test set partition. Via some testing, the performance of these models are comparable to the scores reported in the existing papers \cite{Zhang2020GraphBertOA, Zhang2020SegmentedGF, 10.5555/1795114.1795167, NIPS2013_1cecc7a7}. The source code and pre-trained checkpoints of these graph models have been shared with the community, which can be accessed at the github page mentioned above.


\subsection{Pre-Trained Language Models}

As the base model to be used for building {\our}, we have tried several pre-trained language models with open-source model architectures, configurations, tokenizers and parameter checkpoints. Specifically, the base language models used in this experiment include
\begin{itemize}

\item \textbf{{\our} (GPT-J 6B, 8bit)}: The EleutherAI's GPT-J (6B) is a transformer model trained using Ben Wang's ``Mesh Transformer JAX'' \cite{gpt-j} that has 6 billion parameters. To load GPT-J in float 32, it will require 22+GB RAM to load the model and the fine-tuning will require at least 4x RAM size. To further lower-down the RAM consumption for GPT-J (6B), researchers also propose to quantize it with 8-bit weights \cite{gpt-j-8bit}, which allows scalable fine-tuning with LoRA (Low-Rank Adaptation) and 8-bit Adam and GPU quantization from bitsandbytes. The {\our} (GPT-J 6B, 8bit) model will use GPT-J 6B 8bit as the base model for fine-tuning.


\end{itemize}

More base LLMs will be added and compared in the experiments. Both the source code and the checkpoints of the fine-tuned LLMs used in the experiment have been shared to the community as well.


\subsection{Experimental Settings}

\begin{table*}[h]
\caption{A summary of the experimental results of {\our} on various graph reasoning tasks on the corresponding benchmark datasets. The results are evaluated by the Rouge scores, BLEU and BP scores. Except for the graph loading task, we also evaluate the results on other tasks/datasets by comparing the graph reasoning API calls (other textual contents are excluded) with the ground-truth API calls, and report the Accuracy on reasoning API calls in the table as well.}\label{tab:experimental_result}
\centering
\setlength{\tabcolsep}{.3em}
\renewcommand{\arraystretch}{1.5}
\begin{tabular}{lcc|ccccccc}
\toprule
\hline
\multirow{2}{*}{\textbf{Tasks}} &\multirow{2}{*}{\textbf{Datasets}} &\multirow{2}{*}{\textbf{Methods}} & \multicolumn{6}{c}{\textbf{Evaluation Metrics}}\\
\cline{4-10}
&&&\textbf{Rouge-1} & \textbf{Rouge-2} & \textbf{Rouge-L} & \textbf{Rouge-LSum} & \textbf{BLEU} &\textbf{BP} &\textbf{API-Gen Acc} \\
\hline
\multirow{2}{*}{\parbox{2.5cm}{\textbf{Graph\\ Loading}}}
&\multirow{2}{*}{GL-Prompt} &\multirow{2}{*}{{\our}} & \multirow{2}{*}{ 82.28 } & \multirow{2}{*}{ 67.74 } & \multirow{2}{*}{ 70.93 } & \multirow{2}{*}{ 70.85 } & \multirow{2}{*}{ 63.53 } & \multirow{2}{*}{ 89.98 }	&\multirow{2}{*}{ 4.38 } \\
&&&&&&&&& \\
\hline
\multirow{2}{*}{\parbox{2.5cm}{\textbf{Property\\ Reasoning}}}
&\multirow{2}{*}{GPR-Prompt} &\multirow{2}{*}{{\our}} & \multirow{2}{*}{ 94.56 } & \multirow{2}{*}{ 92.10 } & \multirow{2}{*}{ 91.69 } & \multirow{2}{*}{ 91.69 } & \multirow{2}{*}{ 91.53 } & \multirow{2}{*}{ 99.93 } & \multirow{2}{*}{ 80.00 }\\
&&&&&&&&& \\
\hline
\multirow{3}{*}{\parbox{2.5cm}{\textbf{Paper\\ Topic\\ Reasoning}}}
&Cora &{\our} & 99.69 & 99.68 & 99.69 & 99.69 & 99.2 & 100.0 & 100.0  \\
\cline{3-10}
&Citeseer &{\our} & 100.0 & 100.0 & 100.0 & 100.0 & 99.39 & 100.0 & 97.5  \\
\cline{3-10}
&Pubmed &{\our} & 99.91 & 99.84 & 99.91 & 99.91 & 99.04 & 100.0 & 99.38  \\
\hline
\multirow{4}{*}{\parbox{2.5cm}{\textbf{Molecule\\ Function\\ Reasoning}}}
&PROTEINS &{\our} & 99.61 & 99.19 & 99.61 & 99.61 & 98.27 & 100.0 & 100.0  \\
\cline{3-10}
&PTC &{\our} & 100.0 & 100.0 & 100.0 & 100.0 & 98.52 & 100.0 & 100.0 \\
\cline{3-10}
&NCI1 &{\our} & 100.0 & 100.0 & 100.0 & 100.0 & 98.28 & 100.0 & 100.0  \\
\cline{3-10}
&MUTAG &{\our} & 100.0 & 100.0 & 100.0 & 100.0 & 98.72 & 100.0 & 100.0  \\
\hline
\multirow{3}{*}{\parbox{2.5cm}{\textbf{Sequential\\ Recommendation\\ Reasoning}}}
&MovieLens &{\our} & 97.47 & 96.56 & 97.47 & 97.47 & 94.63 & 95.31 & 93.12 \\
\cline{3-10}
&Last.FM &{\our} & 89.24 & 86.69 & 88.75 & 88.79 & 83.43 & 89.67 & 85.62  \\
\cline{3-10}
&Amazon &{\our}  & 99.9 & 99.8 & 99.9 & 99.9 & 99.74 & 100.0 & 100.0  \\
\hline
\multirow{2}{*}{\parbox{2.5cm}{\textbf{Social\\ Community\\ Reasoning}}}
&Foursquare &{\our} & 98.6 & 98.01 & 98.51 & 98.46 & 97.41 & 100.0 & 95.0\\
\cline{3-10}
&Twitter &{\our} & 99.86 & 99.71 & 99.78 & 99.76 & 99.75 & 99.89 & 98.75 \\
\hline
\multirow{2}{*}{\parbox{2.5cm}{\textbf{Knowledge\\ Graph\\ Reasoning}}}
&Freebase &{\our} & 91.98 & 91.79 & 91.97 & 92.0 & 78.17 & 78.29 & 53.75 \\
\cline{3-10}
&WordNet &{\our} & 98.73 & 98.73 & 98.73 & 98.73 & 97.99 & 98.69 & 96.88  \\
\hline
\bottomrule
\end{tabular}
\end{table*}


\subsubsection{Experimental Setups}

Based on generated graph reasoning prompt datasets, we will fine-tune the base language models. Considering the high-cost in generation for evaluation, we split the prompt datasets for each graph reasoning task into training/testing with the size ratio ``$\min(N-160, 1,600):160$'', where $N$ is the complete prompt dataset size. For the dataset with more than $1,760$ instances, $1,600$ instances will be randomly sampled from it as the training set; whereas for the dataset with less than $1,760$ instances, by excluding the randomly sampled testing set (with $160$ instances), all the remaining $N-160$ instances will be used as the training set. To ensure {\our} will be able to handle diverse graph reasoning queries, those sampled training/testing sets for each graph reasoning task on these graph datasets will be merged together for the base language model fine-tuning and evaluation in {\our}.

Specifically, the hardware and software setups for the fine-tuning of the language models in this experiment are provided as follows:
\begin{itemize}
\item \textbf{Hardware}: We run the fine-tuning experiments on a stand-along workstation with several Nvidia GPUs.  Detailed hardware information about the workstation is as follows: ASUS WS X299 SAGE/10G LGA motherboard, Intel Core i7 CPU 6850K@3.6GHz (6 cores), 1 Nvidia Ampere A100 GPU (80 GB HBM2e DRAM), 1 Nvidia GeForce RTX 4090 Founders Edition GPU (24GB GDDR6X RAM), and 96 GB DDR4 memory and 128 GB SSD swap.

\item \textbf{System and Software}: We run the experiment on Ubuntu 22.04, with CUDA toolkit version 11.8, Nvidia Driver version 520, PyTorch version 1.13.1 and Python 3.9. For the optimizer of {\our} (GPT-J 6B, 8bit), we use the 8-bit AdamW from bitsandbytes with version 0.37.1. We load the pre-trained GPT-J 6B 8bit from Huggingface with weight parameter checkpoint ``hivemind/gpt-j-6B-8bit'' and config/tokenizer checkpoint ``EleutherAI/gpt-j-6b'' as the base model of {\our}, and the installed transformer toolkit version is 4.28.0.dev0. More information about other system and software configurations can be found at the shared anaconda environment file\footnote{https://github.com/jwzhanggy/Graph\_Toolformer/blob/main/environment.yml}.


\item \textbf{Hyper-parameters}: For fine-tuning {\our} (GPT-J 6B, 8bit), we use AdamW with a very small learning rate 1e-5 with weight decay 1e-2, and a max-epoch of 3. Both the training and testing instances are divided into batches with shuffle with batch size 32 and we set the max input/output token length as 128. For the generation function of the language model, the following hyper-parameters are used, {\ie} num-beams: 5, top-k: 5, top-p: 0.95, temperature: 1.9, num-return-sequence: 1, max-length: 128.
\end{itemize}

Especially, when the batch size and input/output max token length are assigned with small values ({\eg} batch-size: 1 or 2 and max-length: 64), we can also fine-tune {\our} (GPT-J 6B, 8bit) model on GPUs with smaller RAM (like Nvidia 1080ti with 11GB memory). It will allow most research groups and individuals to tune and deploy {\our} to provide LLMs based graph reasoning functions and services.

\subsubsection{Performance Evaluation}

For the preliminary performance evaluation of {\our}, we have used several evaluation metrics as follows in the experiments:

\begin{itemize}
\item \textbf{ROUGE scores}: By comparing the outputs of the {\our} framework with the ground-truth, we calculate the Rouge-1, Rouge-2, Rouge-L and Rouge-LSum scores obtained by the model.

\item \textbf{BLEU scores}: Besides the ROUGE scores, we also evaluate the performance of {\our} with the BLEU and BP metrics by comparing the generation output with the ground-truth. 

\item \textbf{API Generation Accuracy}: From the generated statement by the LLMs in {\our}, we will extract and parse the API call queries to compare with the ground-truth. The accuracy of the generated API call queries will also be reported as an extra evaluation metric in this experiment.


\end{itemize}



\subsection{Graph Reasoning Output Statement Generation}



\DeclareRobustCommand{\hlcyan}[1]{{\sethlcolor{cyan}\hl{#1}}}

\begin{table*}[t]
\caption{Case analysis of graph reasoning API call statement generation, query parsing, and query execution and final output generation in {\our}.}\label{tab:successful_cases}
\vspace{-10pt}
\centering
\setlength{\tabcolsep}{.3em}
\resizebox{\textwidth}{!}{
\begin{tabular}{c | p{0.22\textwidth} | p{0.35\textwidth} | p{0.25\textwidth} | p{0.13\textwidth} | p{0.23\textwidth} }
\toprule
\hline
\multirow{2}{*}{\textbf{Tasks}}  & \multicolumn{4}{c}{\textbf{Graph Reasoning Examples with {\our}}} \\
\cline{2-6}

& \textbf{Input Statements} & \textbf{Generated Statements with API calls} & \textbf{Query parsing} & \textbf{Reasoning}& \textbf{Final Statement Output} \\
\hline

\multirow{9}{*}{\makecell{Graph\\ Property\\ Reasoning} }
&What is the order of the diamond graph?
&The order of the diamond graph is [GR(GL("gpr", \{"diamond\_graph"\}), "toolx:order")-->r].
&(('GR', [('GL', ['gpr', '\{"diamond\_graph"\}']), 'toolx:order']), [True])
& 4
&The order of the diamond graph is 4.\\
\cline{2-6}
&The path graph has a center at nodes [TBR].
&The path graph has a center at nodes [GR(GL("gpr", \{"path\_graph"\}), "toolx:center")-->r].
&(('GR', [('GL', ['gpr', '\{"path\_graph"\}']), 'toolx:center']), [True])
&[5, 6]
&The path graph has a center at nodes [5, 6].\\
\cline{2-6}
&The nodes in the wheel graph have eccentricity values of [TBR].
&The nodes in the wheel graph have eccentricity values of [GR(GL("gpr", \{"wheel\_graph"\}), "toolx:eccentricity")-->r].
&(('GR', [('GL', ['gpr', '\{"wheel\_graph"\}']), 'toolx:eccentricity']), [True])
&\{0: 1, 1: 2, 2: 2, 3: 2, 4: 2, 5: 2\}
&The nodes in the wheel graph have eccentricity values of \{0: 1, 1: 2, 2: 2, 3: 2, 4: 2, 5: 2\}.\\
\hline
\hline
\multirow{10}{*}{\makecell{Bibliographic\\ Network\\ Reasoning} }
&What is the topic of paper \#83826 in the cora bibliographic network?
&The topic of paper \#83826 in the cora bibliographic network is [GR(GL("cora"), "graph\_bert:topic", paper\#83826)-->r].
&(('GR', [('GL', ['cora']), 'graph\_bert:topic', 'paper\#83826']), [True])
&Neural\_Networks
&The topic of paper \#83826 in the cora bibliographic network is Neural\_Networks.\\
\cline{2-6}
&Paper \#2998, situated in the citeseer bibliographic network, is centered around the [TBR] topic.
&Paper \#2998, situated in the citeseer bibliographic network, is centered around the [GR(GL("citeseer"), "graph\_bert:topic", paper\#2998)-->r] topic.
&(('GR', [('GL', ['citeseer']), 'graph\_bert:topic', 'paper\#2998']), [True])
&AI
&Paper \#2998, situated in the citeseer bibliographic network, is centered around the AI topic.\\
\cline{2-6}
&Within the pubmed bibliographic network, paper \#16466 focuses on the subject of [TBR].
&Within the pubmed bibliographic network, paper \#16466 focuses on the subject of [GR(GL("pubmed"), "graph\_bert:topic", paper\#16466)-->r].
&(('GR', [('GL', ['pubmed']), 'graph\_bert:topic', 'paper\#16466']), [True])
&1
&Within the pubmed bibliographic network, paper \#16466 focuses on the subject of 1.\\
\hline
\hline
\multirow{12}{*}{\makecell{Molecular\\ Graph\\ Reasoning} }
&What is the function for the protein molecular graph \#138 in proteins?
&The function for the protein molecular graph \#138 in proteins is [GR(GL("proteins"), "seg\_bert:molecule\_function", instance\#138)-->r].
&(('GR', [('GL', ['proteins']), 'seg\_bert:molecule\_function', 'instance\#138']), [True])
&0
&The function for the protein molecular graph \#138 in proteins is 0.\\
\cline{2-6}
&In mutag, instance \#30 of the chemical molecular graph demonstrates a function of [TBR].
&In mutag, instance \#30 of the chemical molecular graph demonstrates a function of [GR(GL("mutag"), "seg\_bert:molecule\_function", instance\#30)-->r].
&(('GR', [('GL', ['mutag']), 'seg\_bert:molecule\_function', 'instance\#30']), [True])
&2
&In mutag, instance \#30 of the chemical molecular graph demonstrates a function of 2.\\
\cline{2-6}
&For chemical molecular graph instance \#652 in nci1, its function is [TBR].
&For chemical molecular graph instance \#652 in nci1, its function is [GR(GL("nci1"), "seg\_bert:molecule\_function", instance\#652)-->r].
&(('GR', [('GL', ['nci1']), 'seg\_bert:molecule\_function', 'instance\#652']), [True])
&0
&For chemical molecular graph instance \#652 in nci1, its function is 0.\\
\cline{2-6}
&The chemical molecular graph numbered 239 in ptc is characterized by a function of [TBR].
&The chemical molecular graph numbered 239 in ptc is characterized by a function of [GR(GL("ptc"), "seg\_bert:molecule\_function", instance\#239)-->r].
&(('GR', [('GL', ['ptc']), 'seg\_bert:molecule\_function', 'instance\#239']), [True])
&0
&The chemical molecular graph numbered 239 in ptc is characterized by a function of 0.\\
\hline
\hline
\multirow{12}{*}{\makecell{Recommender\\ System\\ Reasoning} }
&How likely user \#A1HOLE9R6WPT85 will be interested in item \#B00005MOTF in Amazon?
&The likelihood that user \#A1HOLE9R6WPT85 will be interested in item \#B00005MOTF in Amazon is [GR(GL("amazon"), "bpr:recommendation", user\#A1HOLE9R6WPT85, item\#B00005MOTF)-->r].
&(('GR', [('GL', ['amazon']), 'bpr:recommendation', 'user\#A1HOLE9R6WPT85', 'item\#B00005MOTF']), [True])
&0.008
&The likelihood that user \#A1HOLE9R6WPT85 will be interested in item \#B00005MOTF in Amazon is 0.008.\\
\cline{2-6}
&The likelihood that user \#u1527 will be interested in music from artisit \#i5422 in Last-fm is [TBR].
&The likelihood that user \#u1527 will be interested in music from artisit \#i5422 in Last-fm is [GR(GL("last-fm"), "bpr:recommendation", user\#u1527, artisit\#i5422)-->r].
&(('GR', [('GL', ['last-fm']), 'bpr:recommendation', 'user\#u1527', 'artisit\#i5422']), [True])
&0.248
&The likelihood that user \#u1527 will be interested in music from artisit \#i5422 in Last-fm is 0.248.\\
\cline{2-6}
&In Movielens, the top 10 movies that user \#u272 likes include [TBR].
&In Movielens, the top 10 movies that user \#u272 likes include [GR(GL("movielens"), "bpr:topk\_recommendation", user\#u272, 10)-->r].
&(('GR', [('GL', ['movielens']), 'bpr:topk\_recommendation', 'user\#u272', '10']), [True])
&['i286', 'i288', 'i258', 'i294', 'i300', 'i50', 'i313', 'i100', 'i181', 'i269']
&In Movielens, the top 10 movies that user \#u272 likes include ['i286', 'i288', 'i258', 'i294', 'i300', 'i50', 'i313', 'i100', 'i181', 'i269'].\\
\hline
\hline
\multirow{9}{*}{\makecell{Social\\ Network\\ Reasoning} }
&In the online social network foursquare, which community is user user/1265481 involved in?
&In the online social network foursquare, user user/1265481 is involved in the [GR(GL("foursquare"), "kmeans:community", user\#user/1265481)-->r] communities formed by users.
&('GR', [('GL', ['foursquare']), 'kmeans:community', 'user\#user/1265481'])
&\#2
&In the online social network foursquare, user user/1265481 is involved in the \#2 communities formed by users.\\
\cline{2-6}
&In the online social network twitter, are user \#deeprogress and user \#alejandro1254 belong to the same community?
&In the online social network twitter, user \#deeprogress and user \#alejandro1254 belong to [GR(GL("twitter"), "kmeans:common\_community\_check", user\#deeprogress, user\#alejandro1254)-->r] community.
&('GR', [('GL', ['twitter']), 'kmeans:common\_community\_check', 'user\#deeprogress', 'user\#alejandro1254'])
&The Same
&In the online social network twitter, user \#deeprogress and user \#alejandro1254 belong to the same community.\\
\hline
\hline
\multirow{10}{*}{\makecell{Knowledge\\ Graph\\ Reasoning} }
&According to the Freebase knowledge graph, what is the relation between entity\#/m/053yx and entity\#/m/015\_1q?
&According to the Freebase knowledge graph, the relation between entity\#/m/053yx and entity\#/m/015\_1q is [GR(GL("freebase"), "transe:relation", entity\#/m/053yx, entity\#/m/015\_1q)-->r].
&('GR', [('GL', ['freebase']), 'transe:relation', 'entity\#/m/053yx', 'entity\#/m/015\_1q'])
&/music/artist/label
&According to the Freebase knowledge graph, the relation between entity\#/m/053yx and entity\#/m/015\_1q is /music/artist/label.\\
\cline{2-6}
&According to the WordNet knowledge graph, via relation \#\_hypernym, what entity can we obtain from entity \#imagination.n.02?
&According to the WordNet knowledge graph, via relation \#\_hypernym, we can obtain entity \#imagination.n.02 from entity [GR(GL("wordnet"), "transe:head\_entity", relation\#\_hypernym, entity\#imagination.n.02)-->r].
&('GR', [('GL', ['wordnet']), 'transe:head\_entity', 'relation\#\_hypernym', 'entity\#imagination.n.02'])
&chimera.n.02
&According to the WordNet knowledge graph, via relation \#\_hypernym, we can obtain entity \#imagination.n.02 from entity chimera.n.02.\\

\hline
\bottomrule
\end{tabular}
}
\end{table*}




\DeclareRobustCommand{\hlcyan}[1]{{\sethlcolor{cyan}\hl{#1}}}

\begin{table*}[t]
\caption{Analysis of generated API query cases different from ground-truth queries in the testing set. In this table, we illustrate the examples of input statements, generated statements by {\our} and the desired ground-truth statements of different graph reasoning tasks studied in this paper.}\label{tab:failed_cases}
\vspace{-10pt}
\centering
\small
\setlength{\tabcolsep}{.3em}
\renewcommand{\arraystretch}{1.6}
\resizebox{\textwidth}{!}{
\begin{tabular}{c | p{0.2\textwidth} | p{0.4\textwidth} | p{0.4\textwidth} }
\toprule
\hline
\multirow{2}{*}{\textbf{Tasks}}  & \multicolumn{3}{c}{\textbf{Failed case examples}} \\
\cline{2-4}
& \textbf{Input Statements} & \textbf{Generated Statements} & \textbf{Ground-Truth Statements}\\
\hline
\hline
\multirow{8}{*}{\makecell{Graph\\ Property\\ Reasoning} }
&Nodes have the largest eccentricity [TBR] in the lollipop graph, which make them part of its periphery.
&root>'s eccentricity is [GR(GL("gpr", {"lollipop\_graph"}), "toolx:eccentricity", <root>)-->r].root>'s eccentricity is [GR(GL("gpr", {"lollipop\_graph"}), "toolx:eccentricity", <root>)-->r1].root>'s eccentricity is [GR(GL("gpr
&Nodes [GR(GL("gpr", {"lollipop\_graph"}), "toolx:periphery")-->r] have the largest eccentricity [GR(GL("gpr", {"lollipop\_graph"}), "toolx:eccentricity")] in the lollipop graph, which make them part of its periphery.\\
\cline{2-4}
&The nodes with the smallest eccentricity in the house x graph are [TBR], which are also the center of the tree.
&center>the nodes with the smallest eccentricity in the house x graph are [GR(GL("gpr", {"house\_x\_graph"}), "toolx:eccentricity")-->r], which are also the [GR(GL("gpr", {"house\_x\_graph"}), "toolx:center")-->r].
&The nodes with the smallest eccentricity [GR(GL("gpr", {"house\_x\_graph"}), "toolx:eccentricity")] in the house x graph are [GR(GL("gpr", {"house\_x\_graph"}), "toolx:center")-->r], which are also the center of the tree.\\
\cline{2-4}
&What is the length of shortest path between node \#3 and node \#2 in the diamond graph?
&The length of shortest path between node \#3 and node \#2 in the diamond graph is [GR(GL("gpr", {"diamond\_graph"}), "toolx:shortest\_path", "node\#3", "node\#2")-->r].
&In the diamond graph, the length of shortest path between node \#3 and node \#2 is [GR(GL("gpr", {"diamond\_graph"}), "toolx:shortest\_path", "node\#3", "node\#2")-->r].\\
\hline
\hline
\multirow{5}{*}{\makecell{Bibliographic\\ Network\\ Reasoning} }
&The cora bibliographic network' paper \#28487 is concerned with the area of [TBR].
&ROOT>'s cora bibliographic network' paper \#28487 is concerned with the area of [GR(GL("cora"), "graph\_bert:topic", paper\#28487)-->r].
&The cora bibliographic network' paper \#28487 is concerned with the area of [GR(GL("cora"), "graph\_bert:topic", paper\#28487)-->r].\\
\cline{2-4}
&What is the topic of paper \#5832 in the pubmed bibliographic network?
&The topic of paper \#paper\_number> in the pubmed bibliographic network is [GR(GL("pubmed"), "graph\_bert:topic", paper\#paper\_number>)-->r].
&The topic of paper \#5832 in the pubmed bibliographic network is [GR(GL("pubmed"), "graph\_bert:topic", paper\#5832)-->r].\\
\hline
\hline
\multirow{2}{*}{\makecell{Molecular\\ Graph\\ Reasoning} }
&What is the function for the protein molecular graph \#573 in proteins?
&The function of the protein molecular graph \#573 in proteins is [GR(GL("proteins"), "seg\_bert:molecule\_function", instance\#573)-->r].
&The function for the protein molecular graph \#573 in proteins is [GR(GL("proteins"), "seg\_bert:molecule\_function", instance\#573)-->r].\\
\hline
\hline
\multirow{7}{*}{\makecell{Recommender\\ System\\ Reasoning} }
&In Amazon, what is the item that user \#A3C08BZRVV500V will be most likely to purchase next?
&In Amazon, what is the item that user \#A3C08BZRVV500V will be most likely to purchase next is [GR(GL("amazon"), "bpr:topk\_recommendation", user\#A3C08BZRVV500V, 1)-->r].
&In Amazon, the item that user \#A3C08BZRVV500V will be most likely to purchase next is [GR(GL("amazon"), "bpr:topk\_recommendation", user\#A3C08BZRVV500V, 1)-->r].\\
\cline{2-4}
&In Movielens, which movie user \#u273 will be most likely to watch next?
&In Movielens, which movie user \#u273 will be most likely to watch next?
&In Movielens, the movie that user \#u273 will be most likely to watch next is [GR(GL("movielens"), "bpr:topk\_recommendation", user\#u273, 1)-->r].\\
\cline{2-4}
&In Last-fm, which artisit user \#u1156 will be most likely to listen to next?
&artisit> artisit> artisit> artisit> artisit> artisit> artisit> artisit> artisit> artisit> artisit> artisit> artisit> artisit> 
&In Last-fm, the artisit that user \#u1156 will be most likely to listen to next is [GR(GL("last-fm"), "bpr:topk\_recommendation", user\#u1156, 1)-->r].\\
\hline
\hline
\multirow{6}{*}{\makecell{Social\\ Network\\ Reasoning} }
&In the online social network foursquare, are user \#victorcarbonero and user \#user/11979222 belong to the same community?
&In the online social network foursquare, user \#user/victorcarbonero and user \#user/11979222 belong to [GR(GL("foursquare"), "kmeans:common\_community\_check", user\#user/victorcarbonero, user\#user/11979222)-->r] community.
&In the online social network foursquare, user \#victorcarbonero and user \#user/11979222 belong to [GR(GL("foursquare"), "kmeans:common\_community\_check", user\#victorcarbonero, user\#user/11979222)-->r] community.\\
\cline{2-4}
&In the online social network twitter, are user \#iancr and user \#ClassyIndeed belong to the same community?
&iancr and user \#ClassyIndeed belong to [GR(GL("twitter"), "kmeans:common\_community\_check", user\#iancr, user\#ClassyIndeed)-->r] community.
&In the online social network twitter, user \#iancr and user \#ClassyIndeed belong to [GR(GL("twitter"), "kmeans:common\_community\_check", user\#iancr, user\#ClassyIndeed)-->r] community.\\
\hline
\hline
\multirow{8}{*}{\makecell{Knowledge\\ Graph\\ Reasoning} }
&According to the WordNet knowledge graph, via relation \#\_derivationally\_related\_form, we can obtain entity \#scaremonger.n.01 from entity [TBR].
&
&According to the WordNet knowledge graph, via relation \#\_derivationally\_related\_form, we can obtain entity \#scaremonger.n.01 from entity [GR(GL("wordnet"), "transe:head\_entity", relation\#\_derivationally\_related\_form, entity\#scaremonger.n.01)-->r].\\
\cline{2-4}
&According to the Freebase knowledge graph, from entity\#/m/03r8tl, via relation \#/award/award\_category /nominees./award/award \_nomination/award\_nominee, what entity can we derive?
&According to the Freebase knowledge graph, from entity\#/m/03r8tl, via relation \#/award/award\_category/nominees./award/award\_nomination /award\_nominee, we can derive entity [GR(GL("freebase"), "tr
&According to the Freebase knowledge graph, from entity\#/m/03r8tl, via relation \#/award/award\_category/nominees./award/award\_nomination /award\_nominee, we can derive entity [GR(GL("freebase"), "transe:tail\_entity", entity\#/m/03r8tl, relation\#/award/award\_category/nominees./award /award\_nomination/award\_nominee)-->r].\\
\hline
\bottomrule
\end{tabular}
}
\end{table*}


To evaluate the performance of the fine-tuned LLMs in {\our} on generating the statements with graph reasoning API calls, we have obtained the results of the LLM (GPT-J) on the prompt testing set and the evaluation scores are reported in Table~\ref{tab:experimental_result}. 
\subsubsection{Graph Data Loading}

As shown in Table~\ref{tab:data_statistics}, based on both the instruction and input-output prompt example pairs, we apply ChatGPT to generate about $5,000$ input-output pairs. Via necessary filtering and clean, about $2,802$ are used for the model fine-tuning in the experiment. According to the experimental settings introduced before, we partition the pairs into training and testing sets, and evaluate the performance of the fine-tuned model to evaluate the performance of the {\our} framework for graph data loading.

The experimental results of {\our} on graph data loading evaluated by Rouge and BLEU metrics are provided in Table~\ref{tab:experimental_result}. According to the provided scores, compared with the ground-truth, the outputs generated by {\our} are not bad, which obtained the R1 score of $82.28$ and R2 score of $67.74$. The BLEU scores obtained by {\our} are also very high, which is $63.53$ with BP at $89.98$. Meanwhile, since the ChatGPT generated graph loading API calls have very diverse formats, it is hard for {\our} to obtain precisely the same API calls in the generated output statements, which lead to the API Accuracy to be $4.38$ only.

\subsubsection{General Graph Property Loading}

For the \textit{graph property reasoning} task, we manually create a graph dataset, involving $27$ small-sized classic graph instances, such as \textit{barbell graph}, \textit{wheel graph} and \textit{lollipop graph}, etc. For each graph instance in the dataset, we manually design a few number of reasoning prompts with API calls for its properties (as discussed in Section~\ref{subsubsec:graph_property_reasoning}). Based on both the property reasoning instructions and the hand-crafted prompt examples, with ChatGPT, we further augment the prompt examples and generate a large set of annotated graph property reasoning API call dataset, which will be used for fine-tuning the LLMs in the {\our} framework. 

Some basic statistics about the raw graph dataset and the prompt dataset are provided in the Table~\ref{tab:data_statistics}. On average, the generated graph instances have about $14.70$ nodes and $28.27$ links. We have created a set of $2,587$ API call input-output pairs for reasoning their different properties. Based on the API call instances, different from the graph data loading task, the performance of {\our} on the graph property reasoning tasks is almost perfect, with a $94.56$ R1 score and over $91$ BLEU score as shown in Table~\ref{tab:experimental_result}. Since the API calls for the graph property reasoning follows a standard format, the API generation Accuracy of {\our} is $80\%$ for the graph property reasoning task.



\subsubsection{Bibliographic Paper Topic Reasoning}

We have studied three bibliographic network datasets in this experiment, which include Cora, Citeseer and Pubmed, which are all the frequently used benchmark datasets studied in graph neural network research work. For each of the bibliographic network dataset, we hand-craft a few prompt examples and also augment them with ChatGPT to rephrase and rewrite more diverse input-output prompt pairs. After data filtering, the valid data instances (which works and can obtain the correct graph reasoning results with the API calls) will be used for the model fine-tuning, whose statistical information are provided in Table~\ref{tab:data_statistics}. The performance of the LLMs studied in the experiments are provided in Table~\ref{tab:experimental_result}. Among the three datasets, the performance of {\our} on all these three datasets are very close to $100$ for all these R1, BLEU and API generation Accuracy metrics.



\subsubsection{Protein Molecule Function Reasoning}

For the bio-chemical molecular graph function reasoning task studied in this paper, we will use four bio-chemical graph classification benchmark dataset in the experiments, which include PROTEINS, PTC, NCI1 and MUTAG. In these dataset, each graph instance has both its molecular graph structure and a label indicating its function. For each graph instance, we hand-craft a few prompt examples about its functions, which will be augmented by ChatGPT to rephrase and generate similar data instances. The statistical information about these four datasets are provided in Table~\ref{tab:data_statistics}. We provide the experimental results of the comparison language models in Table~\ref{tab:experimental_result}, and {\our} works perfectly on PROTEINS, PTC, NCI1.

\subsubsection{Sequential Recommender System Reasoning}

For the sequential recommendation reasoning task, we have used three benchmark datasets studied in recommender systems, which include MovieLens, Last.FM and Amazon Review (Software). In these recommender system datasets, both users and items are represented as individual nodes and the interaction between users and items are represented as the links connecting them annotated with the timestamps. Similar to the previous reasoning tasks, we also design a few reasoning prompt examples and further augment them with ChatGPT to generate a large sequential recommender system reasoning dataset. The statistical information about both the raw datasets and the generated reasoning input-output pairs are provided in Table~\ref{tab:data_statistics}. By comparing the studied language models with each, we provide the experimental results of these comparison methods in Table~\ref{tab:experimental_result}. {\our} works very well on generating the correct API calls to the recommender system reasoning tasks on both Movielens and Amazon, and the scores on Last.FM are slightly lower than the other two.

\subsubsection{Online Social Network Reasoning}

In the social network community reasoning task, we use two online social network benchmark datasets, which include Foursquare (an location based online social network) and Twitter (an microblog online social network). These two datasets are initially crawled and used in the social network alignment paper \cite{10.1145/2505515.2505531}, and there exist no ground-truth community information about them. Both Foursquare and Twitter are not very big, containing more than $5,000$ user nodes, from them we generate $64,710$ prompt instances for Foursquare and $52,240$ prompt instance for Twitter. Some other information about the Foursquare and Twitter are also provided in Table~\ref{tab:data_statistics}. The experimental performance results of comparison methods are provided in Table~\ref{tab:experimental_result} and {\our} can correctly create the desired output statements and generate more than $95\%$ of the API calls in the outputs.

\subsubsection{Knowledge Graph Reasoning}

For the knowledge graph reasoning, we use two benchmark datasets, Freebase and WordNet, in the experiments, which provides both entities and their internal relations. In Table~\ref{tab:data_statistics}, we provide the statistical information about the knowledge graphs and the generated reasoning prompts. For the Freebase knowledge graph, there exist $1,345$ types of edges in the graph, and the total number of ``entity-relation-entity'' tuples in the graph is $592,213$; whereas the numbers for the WordNet are $18$ and $151,442$ instead. Among all these tasks studied in this paper, we observe that {\our} achieves slightly lower scores on the knowledge graph reasoning API generation. Partial reasons are due to the special tokens used in the knowledge graph datasets to represent the entity and relation IDs and names, which render the LLM in {\our} fails to reproduce the output statement for many of the instances.


\subsection{Graph Reasoning Case Studies}

In Table~\ref{tab:successful_cases}, we also provide a list of different graph reasoning statements generated by {\our}, we report not only the correctly generated the output statements with API calls but also execute the APIs to obtain the correct reasoning results. 

For the graph property reasoning task, we illustrate the generated outputs by {\our} on three inputs: (1) ``\textit{What is the order of the diamond graph?}'', (2) ``\textit{The path graph has a center at nodes [TBR].}'' and (3) ``\textit{The nodes in the wheel graph have eccentricity values of [TBR].}'', where the \textit{diamond graph}, \textit{path graph} and \textit{wheel graph} mentioned in these three inputs are all the graph instances in the GPR dataset. These three inputs aim to reason about the \textit{order}, \textit{center} and \textit{eccentricity} of the graphs, respectively. As illustrated in the table, {\our} can correctly insert the API calls (1) ``{[GR(GL("gpr", \{"diamond\_graph"\}), "toolx:order")-->r]}'', (2) ``[GR(GL("gpr", \{"path\_graph"\}), "toolx:center")-->r]'', and (3) ``[GR(GL("gpr", \{"wheel\_graph"\}), "toolx:eccentricity")-->r]'' into the output statements, which will call the corresponding API functions in the ``toolx'' toolkit. Since these query statement all have the output tag ``-->r'', {\our} will also replace the final reasoning result by the ``toolx'' toolkit into the output statements as well.

For the bibliographic network based academic paper topic reasoning tasks, we illustrate three reasoning examples on the ``Cora'', ``Citeseer'' and ``Pubmed'' bibliographic networks, respectively. For paper nodes in both ``Cora'' and ``Citeseer'', they are annotated with textual labels, {\eg} ``Neural Networks'' for paper node \#83826 in ``Cora'' and ``AI'' for paper node \#2998 in ``Citeseer'' as shown in the table. Meanwhile, for the paper nodes in ``Pubmed'', we only have the integer annotated class labels, {\eg} ``1'' for paper node \#16466 in ``Pubmed''. Similarly, we also illustrate the molecular graph function reasoning examples on the four molecular graph datasets, {\ie} ``PROTEINS'', ``MUTAG'', ``NCI1'' and ``PTC'', where the graph instances are also annotated with the integer class labels that indicating their functions as well.

For the sequential recommender system reasoning, we illustrate the reasoning examples on both calculating the preference scores of users on certain items and the top-k recommendation for users. For both users and items in the ``Last-fm'' and ``Movielens'' datasets, they are denoted by the integer IDs, to differentiate users from items, we specifically add the ``u'' and ``i'' before their integer IDs in both the released graph datasets and the prompt datasets, {\eg}, ``\#u1527'' and ``\#i5422'' for the example in the table. For the social network community reasoning, we provide the examples on reasoning both community IDs for users and common community checking for input users. Finally, for the reasoning on the knowledge graphs, we illustrate the examples for searching both entities and relations in the table, which illustrate how {\our} works on handling such different knowledge graph reasoning tasks.


\subsection{Inconsistent Generation Result Analysis}

Although {\our} is capable to generate the correct graph reasoning queries for most of the input query statements and questions, but {\our} may still make some mistakes or generate the outputs that are inconsistent with the ground-truth statements. In Table~\ref{tab:failed_cases}, we summarize several types of inconsistent cases of {\our} in generating the graph reasoning queries from the prompt testing set.

For a very small number of the inputs, {\our} will generate very messy and duplicated outputs with random API calls, like the first graph property reasoning example in Table~\ref{tab:failed_cases}. Given the input statement ``Nodes have the largest eccentricity [TBR] in the lollipop graph, which make them part of its periphery.'', the generated output by {\our} is very different from the ground-truth output. It is similar for the last recommender system reasoning example in Table~\ref{tab:failed_cases} as well, whose generation results contains the duplicated wrong token list of ``artisit>''.

For a few inputs, the generation results by {\our} are very close to the ground-truth will has some extra tokens prepend to the output that don't exist in the ground-truth statements. For instance, like second graph property reasoning example and the first bibliographic network reasoning example in Table~\ref{tab:failed_cases}, the generated statements by {\our} are prepended by the tokens ``root>'s'' and ``ROOT>'s'', which don't exist in the prompt datasets actually.

For some of the examples in Table~\ref{tab:failed_cases}, the generated graph reasoning API calls are the same as the desired API calls in the ground-truth, but the statements are presented in another way, like the last graph property reasoning example, the molecular graph function reasoning example, and the first recommender system reasoning example.

In addition to the above inconsistency cases, we also observe several other very rare inconsistency in the generated statements, like (1) fail to generate the key parameters, {\eg} the second bibliographic network reasoning example, (2) fail to add the API call query, {\eg} the second recommender system reasoning example, (3) incomplete statements, {\eg} the second social network community reasoning example and the last knowledge graph reasoning example, and (4) fail to provide the generation statements, {\eg} the first knowledge graph reasoning example.


\subsection{Zero-Shot Inference}



\DeclareRobustCommand{\hlcyan}[1]{{\sethlcolor{cyan}\hl{#1}}}

\begin{table*}[t]
\caption{Zero-shot graph reasoning API query statement generation on new graph datasets and with new tasks.}\label{tab:zero_shot_cases}
\centering
\small
\setlength{\tabcolsep}{.3em}
\resizebox{\textwidth}{!}{
\begin{tabular}{c | p{0.4\textwidth} | p{0.4\textwidth} | c}
\toprule
\hline
\multirow{2}{*}{\textbf{Categories}}  & \multicolumn{3}{c}{\textbf{Failed case examples}} \\
\cline{2-4}
& \textbf{Input Statements} & \textbf{Generated Statements} & \textbf{Correctness Tag} \\

\hline

\multirow{18}{*}{{\makecell{New\\ Graph\\ Data}} }
&What is the density of the Random Graph?
&The density of the random graph is [GR(GL("gpr", {"random\_graph"}), "toolx:density")-->r].
&\multirow{2}{*}{\tikzcmark}\\
\cline{2-4}
&What is the diameter of the online social network Twitter?
&The diameter of the online social network Twitter is [GR(GL("twitter"), "seg\_bert:molecule\_gtopic")].
&\multirow{2}{*}{\tikzxmark}\\
\cline{2-4}
&What is the topic of paper \#1111 in the DBLP bibliographic network?
&The topic of paper \#1111 in the DBLP bibliographic network is [GR(GL("dblp"), "graph\_bert:topic", paper\#1111)-->r].
&\multirow{2}{*}{\tikzcmark}\\
\cline{2-4}
&In AlphaFold, instance \#30 of the protein molecular graph demonstrates a function of [TBR].
& The protein molecular graph instance \#30 has the function of [GR(GL("alphafold"), "graph\_bert:molecule\_function", instance\#30)-->r].
&\multirow{3}{*}{\tikzcmark}\\
\cline{2-4}
&What is the rating score user \#Bob gives to movie \#Titanic at IMDB?
&The user \#Bob gives the movie \#Titanic the average rating score of [GR(GL("imdb"), "bpr:recommendation", user\#Bob, movie\#Titanic)-->r].
&\multirow{3}{*}{\tikzcmark}\\
\cline{2-4}
&In Facebook, what is the id of user \#random\_user's community?
&In Facebook, the id of user \#random\_user's community is [GR(GL("facebook"), "kmeans:community", user\#random\_user, "community")-->r].
&\multirow{3}{*}{\tikzcmark}\\\cline{2-4}
&According to the wikipedia knowledge graph, what is the relation between entity \#Donald\_Trump and entity \#United\_State\_of\_America?
&According to the wikipedia knowledge graph, the relation between entity \#Donald\_Trump and entity \#United\_State\_of\_America is [GR(GL("wikis"), "transe:relation", entity\#Donald\_Trump, entity\#United\_State\_of\_America)-->r].
&\multirow{4}{*}{\tikzcmark}\\

\hline
\hline
\multirow{18}{*}{{\makecell{New\\ Reasoning\\ Tasks}} }
&What is the new\_attribute of the bull graph?
&The new\_attribute of the bull graph is [GR(GL("gpr", {"bull\_graph"}), "toolx:new\_attribute", "graph")-->r].
&\multirow{2}{*}{\tikzcmark}\\
\cline{2-4}
&Who are the authors of the paper \#111 in the cora bibliographic network?
&ids of the authors of the paper \#111 in the cora bibliographic network include [GR(GL("cora"), "graph\_bert:topic", paper\#111)-->r].
&\multirow{3}{*}{\tikzxmark}\\
\cline{2-4}
&How many atoms are there in the protein molecular graph \#111 in mutag?
&The protein molecular graph \#111 in mutag has 
[GR(GL("mutag"), "seg\_bert:molecule\_function", instance\#111)-->r] 
--[GR(GL("mutag"), "seg\_bert:molecule\_function\_relation", instance\#111)-->r].
center>The protein molecular graph \#111 in mutag has [GR(GL("mutag"), "seg\_bert:molecule\_function
&\multirow{5}{*}{\tikzxmark}\\
\cline{2-4}
&In Amazon, what is the brand that user \#Bob will be mostly likely to purchase next?
&In Amazon, the brand that user \#Bob will be mostly likely to purchase next is [GR(GL("amazon"), "bpr:topk\_recommendation", user\#Bob, 1)-->r].
&\multirow{3}{*}{\tikzxmark}\\
\cline{2-4}
&In online social network Twitter, are user \#Alex and user \#Bob both following Elon Musk?
&In online social network Twitter, are user \#Alex and user \#Bob both following user \#ElonMusk? 
&\multirow{2}{*}{\tikzxmark}\\
\cline{2-4}
&According to the Freebase knowledge graph, how many relations are there between entity \#Donald\_Trump and entity \#United\_States\_of\_America?
&According to the Freebase knowledge graph, how many relations are there between entity \#Donald\_Trump and entity [GR(GL("freebase"), "transe:relation", entity\#Donald\_Trump, entity\#United\_States\_of\_America)-->r].
&\multirow{4}{*}{\tikzxmark}\\

\hline

\bottomrule
\end{tabular}
}
\end{table*}


What's more, we also analyze the zero-shot inference ability of {\our} in generating the API calls for graph reasoning tasks on either \textit{new graph datasets} or for \textit{new reasoning tasks}. The case studies of some examples are provided in Table~\ref{tab:zero_shot_cases}.

According to the case studies shown in Table~\ref{tab:zero_shot_cases}, we observe that among the seven provided reasoning tasks on new graph datasets, six of them can generate the desired statements with the correct API calls. The only wrong generated statement is for the graph property reasoning statement ``What is the diameter of the online social network Twitter?'', which generates a statement unexpected API calls ``The diameter of the online social network Twitter is [GR(GL("twitter"), "seg\_bert:molecule\_gtopic")].'' 

Meanwhile, among the cases on new reasoning tasks, only one of these six provided examples can generate the desired outputs, which is for calculating some ``new\_attribute'' of the bull graph. However, for the other new tasks that are not included in the prompt training set, {\eg} the ``paper author reasoning'', ``atom node number reasoning of molecular graphs'', ``brand reasoning in recommender systems'', ``common followee reasoning of users in social networks'' and ``relation count reasoning in knowledge graphs'', the generated output statements by {\our} are all wrong.


\subsection{Language Generation Ability Revisit}



\DeclareRobustCommand{\hlcyan}[1]{{\sethlcolor{cyan}\hl{#1}}}

\begin{table*}[t]
\caption{Analysis of generated API query cases different from ground-truth queries in the testing set.}\label{tab:language_generation_cases}
\centering
\small
\setlength{\tabcolsep}{.3em}
\renewcommand{\arraystretch}{1.6}
\resizebox{\textwidth}{!}{
\begin{tabular}{c | p{0.35\textwidth} | p{0.35\textwidth} | p{0.35\textwidth} }
\toprule
\hline
\multirow{2}{*}{\textbf{\makecell{Input\\ Sources}}}  & \multicolumn{3}{c}{\textbf{Language Generation Examples}} \\
\cline{2-4}
&\textbf{Input Payload} & \textbf{Generated Outputs Before Fine-Tuning} & \textbf{Generated Outputs After Fine-Tuning}\\
\hline
\multirow{9}{*}{\makecell{{Pile}\\ {Data}} }
&Roman Catholic Diocese of Tambacounda$\backslash$n$\backslash$nThe Roman Catholic Diocese of Tambacounda () is a diocese located in the city of Tambacounda in the Ecclesiastical province of Dakar in Senegal.
&$\backslash$n$\backslash$nHistory$\backslash$n$\backslash$nThe Diocese of Tambacounda was erected by Pope Pius XI on December 21, 1933, on territory split off from the Diocese of Saint-Louis du Sénégal.$\backslash$n$\backslash$nBishops$\backslash$n$\backslash$nDiocese of Tambacounda$\backslash$nErected: December 21, 1933$\backslash$nMetropolitan: Archdiocese of Dakar$\backslash$n
&center>The Roman Catholic Diocese of Tambacounda () is a diocese located in the city of Tambacounda in the <center>Ecclesiastical province of Dakar in Senegal.</center>
center>Output: </center>
center>The <center>Roman</center> <center>Catholic\\
\cline{2-4}
&Mud Hens pitcher Evan Reed charged with sexual assault$\backslash$n$\backslash$nMud Hens pitcher Evan Reed was charged July 30 with sexual assault related to a March incident in Detroit when he was a member of the Detroit Tigers.
&$\backslash$n$\backslash$nMud Hens pitcher Evan Reed was charged July 30 with sexual assault related to a March incident in Detroit when he was a member of the Detroit Tigers.$\backslash$n$\backslash$nMud Hens pitcher Evan Reed was charged July 30 with sexual assault related to a March incident in Detroit when he was a member of the Detroit Tigers.$\backslash$n$\backslash$nMud Hens pitcher Evan Reed was charged July 30 with sexual assault related to
&center>\#\#\#<centercentercentercentercentercentercenter
centercentercentercentercentercentercentercenter
centercentercentercentercentercentercentercenter
centercentercentercentercentercentercentercenter
centercentercentercentercenter\\
\hline
\hline
\multirow{7}{*}{\makecell{{Random}\\ {Sources}} }
&Thursday’s verdict only added to the intrigue surrounding the gravest legal and political unknown from the 2021 Capitol insurrection that hangs over the 2024 campaign: will Donald Trump, the president who inspired the uprising, face his own legal and political price?
&I don’t think there’s any question that he’s going to be impeached,” Rep. Al Green (D-TX) told The Daily Beast. “I think he’s going to be impeached, and I think he’s going to be removed from office.”$\backslash$n$\backslash$nGreen’s
&will be the verdict only added to the intrigue surrounding the gravest legal and political unknown from the 2021 Capitol insurrection that hangs over the 2024 campaign: will Donald Trump, the president who inspired the uprising, face his own legal and political price?br>\\
\cline{2-4}
&A source of debate for centuries, the village of Ponte Buriano, a suburb of Arezzo in the Tuscany region of Italy, is so convinced the bridge behind Mona Lisa is the Ponte Buriano that they've made it a key feature of their local tourism campaign, even laying claim to the bridge on the village's welcome sign.
&$\backslash$n$\backslash$nThe Ponte Buriano is one of the oldest bridges in Tuscany, dating back to the 12th century. It's also one of the most famous, thanks to Leonardo da Vinci's Mona Lisa, the most famous painting in the world.$\backslash$n$\backslash$nThe
&The village of [GR(GL("mona\_lisa"), "bpr:search", GR(GL("ponte\_buriano"), GR(GL("search"), GR(GL("mona\_lisa"), GR(GL("search\\
\hline

\bottomrule
\end{tabular}
}
\end{table*}


At the end of this section, we also want to provide more analyses about the impacts of the fine-tuning on LLM's language generation abilities. In Table~\ref{tab:language_generation_cases}, we illustrate some examples about the generation results by the LLM in {\our} before and after the fine-tuning with the graph reasoning prompt data. Specifically, we select the inputs from two different sources, {\ie} two instances from the Pile testing set (Pile was the data used for GPT-J pre-training) and two instances from the recent news articles on the web.

By comparing the generation results, we can observe very large (negative) impacts of the fine-tuning with the prompt datasets on LLM's language generation ability. For these four input payloads, after the fine-tuning, the outputs generated by the LLM in {\our} are either some random tokens or contains the unexpected API calls, and only for the third input, the output by the LLM in {\our} is till closely related to the inputs. 

Therefore, if we plan to make {\our} a very general language interface that can not only handle graph reasoning tasks but also still possess the language generation ability for the inputs not related to graph reasoning, some new continual learning techniques will be needed in model fine-tuning, so the LLM in {\our} will not suffer from the catastrophic forgetting problem after fine-tuning.

}

\section{Conclusion}\label{sec:conclusion}

In this paper, we investigate the principles, methodologies and algorithms to empower existing LLMs with graph reasoning ability. We introduce the {\our} framework, and propose to teach LLMs to use external graph data loading and graph reasoning tools for addressing the tasks. Specifically, several representative graph reasoning tasks are studied in this paper, including the basic graph property reasoning task, and the advanced tasks, like bibliographic paper topic reasoning, molecular graph function reasoning, sequential recommender system reasoning, social network community reasoning and the knowledge graph reasoning. For each of these graph reasoning tasks, we select several benchmark graph datasets in the experiment, and design the reasoning API call prompt examples manually. In addition, with the help of ChatGPT, we can further augment the reasoning API call templates to generate large-sized graph reasoning datasets. Via fine-tuning of existing pre-trained LLMs with the generated prompt dataset, the {\our} framework has been demonstrated to be effective with the extensive experimental results on many of these studied graph reasoning tasks.

\section{Future Research Directions}\label{sec:future_work}

The explorations of this paper try to help identify and develop one potential framework {\our} that can help bridge the graph learning community with the latest research work on LLMs and AIGC. The experimental results reported in this paper also demonstrate that it is feasible to use LLMs as a unified and general interface to conduct various graph reasoning tasks. Based on this paper, researchers in the graph learning communities may consider to further propose new models and frameworks that integrate graph reasoning abilities to the latest LLMs. Along with the explorations in this paper, we also identify several potential research opportunities with the current framework and the readers may consider to further explore in the future, which are listed as follows.

\begin{itemize}

\item \textbf{GNNs Transfer}: Graph neural networks (GNNs) pre-training has been studied for years. Different from the pre-trained language and vision models, pre-trained GNNs has very limited applications in the real-world actually, since GNNs are harder to be transferred to new graph reasoning tasks on new graph datasets. Currently, pre-trained GNN models are normally dedicatedly used in the same (or similar) task(s) that pre-train them. In this paper, {\our} can serve as a hosting platform for deploying various pre-trained GNN models, and we expect to see more pre-trained GNN models to be added to the {\our} framework in the future. At the same time, we also identify a potential problem with the current graph models. As more and more graph data and graph reasoning tasks are added to {\our}, the number of required pre-trained GNN models in {\our} will grow quadratically, since a new pre-trained GNN will be needed for a specific graph reasoning task on a certain graph datasets. So, the number of required graph models in {\our} is approximately equal to ``$|\textit{graph datasets}| \times |\textit{graph reasoning tasks}|$''. Improving the transferability of pre-trained GNN models both across graph reasoning tasks and different graph datasets should be one of major the research exploration focus of the graph learning community for the future.

\item \textbf{Integrated Learning of LLMs and GNNs}: In {\our}, the pre-training of GNNs and LLMs are separated from each other. The LLMs in {\our} are used as more like a reasoning language interface, that will call the pre-trained GNN models for accomplishing certain tasks. In other words, the LLMs in {\our} has no access to the internal components of the GNNs, and will not use any hidden representations learned for the input graphs/nodes/links in the text output generation. For the graph reasoning result oriented tasks, the current framework {\our} works very well. Meanwhile, when it comes to some tasks that may require the LLMs to access the graph reasoning process and internal (intermediate) embedding representations, the current framework will become insufficient. For instance, the following graph reasoning interpretability studies will require the LLMs provide a detailed textual explanation of not only the graph reasoning results but also the reasoning process. Without the access the GNNs' internal components and representations, it will be hard or infeasible for LLMs to generate such textual explanations.

\item \textbf{Regulation and Interpretability}: Nowadays, lots of people have been calling for setting up stricter regulation policies and laws on the AI systems, models, products and services. Impressed by the current AI products and services, human have presented very complicated reactions to the fast growing new AI models: we want to enjoy the services from AI systems but also expect that the systems are safe, reliable and robust. To bridge people's expectations and the current AI models, the model performance and result interpretability plays a critical role. As mentioned above, the current LLMs in {\our} can automatically call the graph reasoning API queries to get the results and replace the results with the generated queries as the final output. Meanwhile, as to the the reasoning process and reasoning result interpretability, the current {\our} cannot provide the textual explanations, which will be one of the most important research focus for the future projects.

\item \textbf{Efficiency}: With LoRA and quantized models/optimizers, we can reduce the model fine-tuning memory capacity requirement to less than 11GB and the memory capacity requirement even lower for the model inference stage. Meanwhile, integrated with the large-sized graph data, pre-trained graph models, and necessary pre-processed data, the efficiency of {\our} for various graph reasoning task can still be a problem. In this paper, we introduce a tentative approach to make the problem less severe with the working memory. However, if we plan to deploy {\our} on devices with very small memories, like cell-phones or embedded equipments, new techniques will still be needed to improve the model learning and inference efficiency.

\item \textbf{Diverse Applications}: Due to the limited space, we can only study a few number of the graph reasoning tasks with {\our} in this paper. Meanwhile, in the real-world, we have lots of graph structured data that may require the LLMs to handle them to reason for the desired outputs. Therefore, a very promising future work direction is to apply {\our} to study diverse real-world graph/network data oriented reasoning tasks with LLMs. We list a few of them here just for the readers' information, and the readers may explore more diverse reasoning tasks according to your own backgrounds and expertises.
\begin{itemize}

\item \textbf{Urban Computing and Smart City}: In the offline world, we have extensively connected traffic networks that bridge different local communities, cities and countries by local roads, national highways, international fights and ocean freight corridors. Applying LLMs for knowledge extraction and reasoning based on such traffic networks is critical for the current urban computing and smart city projects.

\item \textbf{IoT and Smart Home}: Assisted with the 5G, the IoT network effectively bridges the cyber world with the physical devices and equipments together via extremely fast communication channels. The LLMs provide the opportunity for us to utilize language models as the general interface for controlling the devices within the IoT networks, which is also the main objective for building the smart home system.

\item \textbf{Healthcare}: During the past years, the world has suffered a lot from the covid-19 pandemic. Similar to the protein molecules studied in this paper, both the virus and the vaccines can also be represented as the molecular graphs. LLMs with the molecular graph reasoning ability have the potential to improve our current healthcare system in many perspectives, like \textit{early identification of virus}, \textit{analysis of the virus pathogenicity} and \textit{creation of vaccines}. What's more, the LLMs with the social network reasoning ability will also help \textit{infer the potential virus propagation among people}, \textit{early prediction of highly infectious communities} and \textit{identify rumors and misinformation about the pandemic} (at the online social networks).

\item \textbf{Multi-Modal Learning}: Extending the LLMs to handle multi-modal inputs is the main exploration focus at present for both AIGC and AGI research. Graphs actually can serve as the modeling data representation for bridging the data in different modalities, {\eg} \textit{knowledge graph} from texts and \textit{scene graph} from images. Exploring to integrate all such multi-modal data based learning systems within one unified framework via graph equipped LLMs can also be a promising research direction.

\end{itemize}

\end{itemize}

\bibliography{reference}
\bibliographystyle{plain}
\newpage
\appendix

{
\sloppy
\setlength{\columnseprule}{0.4pt}

\section{Appendix: Graph Data Format}

\subsection{Graph Data Loading}

The datasets are stored in binary format, which can be loaded with pickle, e.g., the GPR dataset can be loaded as follows:

\begin{lstlisting}[language=Python]
import pickle
f = open('./gpr', 'rb')
dataset = pickle.load(f)
f.close()
print(dataset.keys())
\end{lstlisting}

\subsection{Graph Dataset}

For the datasets with 1 single large-scale graph/network (including, Cora, Pubmed, Citeseer; Twitter, Foursquare; Amazon, Last-FM, Movielens; WordNet, Freebase), the loaded "dataset" is organized with a python dictionary with the following format:

\begin{lstlisting}[language=Python]
dataset = {
        "data_profile": {
            'name': dataset_name,
            'order': node_number,
            'size': link_number,
            'is_directed': boolean,
            'is_weighted': boolean,
            # besides the above profile information, for some data, we will also include some other attributes, like feature vector dimensions, label space dimension, etc., in the data profile dict.
        },
        "nodes": {
          node_id: {'features': feature, 'label': label,}
        },
        "links": {
          node_pair: {'features': feature, 'label': label,}
        }
    }
\end{lstlisting}
    
\subsection{Graph Instance Dataset}

For the datasets with multiple graph instances (GPR; Proteins, Mutag, NCI1, PTC), the loaded "dataset" is organized with a python dictionary with the following format:

\begin{lstlisting}[language=Python]
dataset = {
        "data_profile": {
            'name': dataset_name,
            'graph_number': graph_number,
            'is_directed': boolean,
            'is_weighted': boolean,
            # besides the above profile information, for some data, we will also include some other attributes, like feature vector dimensions, label space dimension, etc., in the data profile dict.
        },
        'graph_set': {
            graph_id: {
                'nodes': node_set,
                'links': link_set,
                'label': graph_instance_label,
            }
        }
    }
\end{lstlisting}   
 

\section{Appendix: Prompt Dataset}

\subsection{Shared Prompt Datasets}

This directory contains the graph reasoning prompts for 15 different graph datasets. They all have their corresponding raw graph datasets (which can be downloaded from this page).

Each directory contains the train/test graph reasoning tuples for different graph datasets.

\begin{itemize}
\item \textbf{mixed}: it merges all train/test prompts from all the following 15 graph dataset (except the graph\_data\_loading), we will use this for the LLM tuning.
\item \textbf{graph\_properties}: it contains the train/test prompts for the gpr dataset created in this paper on graph property reasoning
\item \textbf{bibliographic\_networks}: it contains the train/test prompts for 3 bibliographc network reasoning datasets, cora, pubmed, citeseer
\item \textbf{molecular\_graphs}: it contains the train/test prompts for 4 molecular graph reasoning datasets, proteins, mutag, nci1, ptc
\item \textbf{recommender\_systems}: it contains the train/test prompts for 3 recommender system reasoning datasets, amazon, last-fm, movielens
\item \textbf{social\_networks}: it contains the train/test prompts for 2 social network reasoning datasets, foursquare, twitter
\item \textbf{knowledge\_graphs}: it contains the train/test prompts for 2 knowledge graph reasoning datasets, wordnet, freebase
\end{itemize}

These above prompts are all created with prompt templates augmented by ChatGPT based on the concrete graph datasets. In the prompts, we will use the concrete data instances, node ids, relations, and reasoning outputs.

In addition to these prompts corresponding to concrete graph datasets, we also include a prompt dataset purely generated by chatgpt on graph loading:

\begin{itemize}
\item \textbf{graph\_data\_loading}: it contains the pure-chatgpt generated graph data loading prompts
\end{itemize}

\subsection{Prompt Format}

Each of the above directory contains two files: 
\begin{lstlisting}[language=Python]
prompts_train
prompts_test
\end{lstlisting}   
which denote denote the prompts for training and testing, respectively. Each prompt instance in the training/testing sets has 3 entries: 
\begin{lstlisting}[language=Python]
Input, Output, Reasoning Result
\end{lstlisting}   

\begin{itemize}
\item \textbf{Input}: The input contains the potential query inputs to the Graph-toolformers.
\item \textbf{Output}: The output will be the annotated query output generated by the LLMs with added graph reasoning API calls.
\item \textbf{Reasoning Result}: We also add the reasoning result for many prompt tupes, which will be used for reasoning result evaluation only.
\end{itemize}

\section{Appendix: Pre-trained Graph Models}

6 different pre-trained graph models are included in the {\our} framework for different graph reasoning tasks, which are listed as follows:

\begin{itemize}
\item \textbf{Graph Property Reasoning Model}: toolx
\item \textbf{Bibliographic Network Reasoning Model}: Graph-Bert
\item \textbf{Molecular Graph Reasoning Model}: SEG-Bert
\item \textbf{Online Social Network Reasoning Model}: KMeans
\item \textbf{Recommender System Reasoning Model}: BPR (Bayesian Personalized Ranking)
\item \textbf{Knowledge Graph Reasoning Model}: TransE
\end{itemize}

\subsection{Toolx for Graph Property Reasoning}

The current toolx model is implemented based on networkx, and toolx will implement different functions to calculate different graph properties mentioned in the paper, which are also listed as follows:

\begin{lstlisting}[language=Python]
- order
- size
- density
- eccentricity
- radius
- diameter
- center
- shortest_path
- avg_path_length
- min_path_length
- max_path_length
- periphery
\end{lstlisting}   

\subsection{Graph-Bert for Bibliographic Network Paper Topic Reasoning}

The Graph-Bert model was proposed in paper entitled "Graph-Bert: Only Attention is Needed for Learning Graph Representations".

The Graph-Bert will be used to implement the bibliographic network paper topic inference function, which will be reduced to the node classification task:

\begin{lstlisting}[language=Python]
- node_classification in GraphBertNodeClassification.py
\end{lstlisting}  

The model has been pre-trained on the cora, pubmed, citeseer datasets already based on the identical train/test sets introduced in the paper. Both the model code and the pre-trained model parameter checkpoints are provided.

As to the original source code, readers may consider to refer to the repository (https://github.com/jwzhanggy/Graph-Bert) for more information.

\subsection{SEG-Bert for Molecualr Graph Function Reasoning}

The SEG-Bert model was proposed in the paper entitled "Segmented Graph-Bert for Graph Instance Modeling ".

The SEG-Bert will be used to implement the molecular graph function inference function, which will be reduced to the graph classification task:

\begin{lstlisting}[language=Python]
- graph_classification in SegmentedGraphBertGraphClassification.py
\end{lstlisting}   

The model has been pre-trained on the proteins, mutag, nci1 and ptc datasets already based on the identical train/test sets introduced in the paper. Both the model code and the pre-trained model parameter checkpoints are provided.

As to the original source code, readers may consider to refer to the repository (https://github.com/jwzhanggy/Graph-Bert) for more information.

\subsection{KMeans for Social Network Community Reasoning}

To detect the social network community, based on the social network structure (adjacency matrix), we calculate the nodes' pairwise common neighbor numbers to define their closeness, which will be fed to KMeans for community detection.

\begin{lstlisting}[language=Python]
- community(graph): return the community label for all user nodes in the social network
- community(graph, node): return the community label for specify input user nodes
- community_count(graph): return the number of detected communities
- community_avg_size(graph): return the average size of detected communities
- community_max_size(graph): return the max size of detected communities
- community_size(graph, node): return the size of community that the input user node belongs to
- common_community_check(graph, node1, node2): check if user1 and user2 belong to the same community or not
\end{lstlisting}   

\subsection{BPR for Recommender System Reasoning}

The BPR model was proposed in the paper entitled "BPR: Bayesian personalized ranking from implicit feedback".

The BPR model will be used to implement the social network community detection functions, which will be reduced to the graph partition/clustering tasks:

\begin{lstlisting}[language=Python]
- recommendation in BPR.py
- topk_recommendation in BPR.py
\end{lstlisting}   

\subsection{TransE for Knowledge Graph Reasoning}

The TransE model was proposed in the paper entitled "Transition-based Knowledge Graph Embedding with Relational Mapping Properties".

The TransE will be used to implement the knowledge graph entity/relation searching functions, which will be reduced to the graph searching tasks:

\begin{lstlisting}[language=Python]
- search_head_entity in TransE.py
- search_tail_entity in TransE.py
- search_relation in TransE.py
\end{lstlisting}

}


\end{document}